\documentclass[fleqn,10pt]{wlscirep}
\usepackage[utf8]{inputenc}
\usepackage[T1]{fontenc}
\usepackage{tabularx}
\usepackage{graphicx}
\usepackage{subcaption}
\usepackage{float}  
\usepackage{lineno}
\usepackage{makecell}  
\usepackage{array} 
\usepackage{booktabs}
\usepackage{multirow}
\usepackage{graphicx}   
\usepackage{lineno}
\linenumbers
\usepackage{makecell} 
\usepackage{booktabs} 
\usepackage{longtable}
\usepackage{cleveref}
\usepackage{url}
\DeclareUnicodeCharacter{2212}{\ensuremath{-}}
\usepackage{hyperref}
\usepackage{authblk}
\usepackage{caption}
\title{A Comprehensive Benchmark of Histopathology Foundation Models for Kidney Digital Pathology Images}

\author[1]{Harishwar Reddy Kasireddy}
\author[2,3]{Patricio S. La Rosa}
\author[4]{Akshita Gupta}
\author[3]{Anindya S. Paul}
\author[1]{Jamie L. Fermin}
\author[5]{William L. Clapp}
\author[6]{Meryl A. Waldman}
\author[7]{Tarek M. El-Ashkar}
\author[8]{\quad \quad \quad Sanjay Jain}
\author[9]{Lu\'{\i}s Rodrigues}
\author[10]{Kuang Yu Jen}
\author[11]{Avi Z. Rosenberg}
\author[7]{Michael T. Eadon}
\author[12 *]{Jeffrey B. Hodgin}
\author[3 *]{Pinaki Sarder}

\affil[1]{Department of Electrical and Computer Engineering, University of Florida, Gainesville, FL, USA}
\affil[2]{Seed Production Innovation, Crop Science Division, Bayer Company, St.\ Louis, MO, USA}
\affil[3]{Division of Medicine – Quantitative Health, University of Florida, Gainesville, FL, USA}
\affil[4]{Department of Health Outcomes and Biomedical Informatics, University of Florida, Gainesville, FL, USA}
\affil[5]{Department of Pathology, Immunology and Laboratory Medicine, University of Florida College of Medicine, Gainesville, FL, USA}
\affil[6]{Kidney Disease Branch, National Institute of Diabetes and Digestive and Kidney Diseases, National Institutes of Health, Bethesda, MD, USA}
\affil[7]{Indiana University School of Medicine, Indianapolis, IN, USA}
\affil[8]{Departments of Medicine, Washington University School of Medicine, St.\ Louis, MO, USA}
\affil[9]{Universidade de Coimbra, Coimbra, Coimbra District, Portugal}
\affil[10]{Department of Pathology and Laboratory Medicine, University of California at Davis School of Medicine, Sacramento, CA, USA}
\affil[11]{Department of Pathology, Johns Hopkins University School of Medicine, Baltimore, MD, USA}
\affil[12]{Department of Pathology, University of Michigan, Ann Arbor, MI, USA}
\affil[*]{Correspondence: \url{jhodgin@med.umich.edu}; \url{pinaki.sarder@ufl.edu}}

\begin{abstract}
Histopathology foundation models (HFMs), trained on large-scale histopathology cancer datasets, have advanced computational digital pathology. However, the applicability of these HFMs to non-cancerous chronic kidney conditions remains underexplored. This gap is critical to address since chronic kidney pathologies can coexist with cancer types such as renal cell carcinoma and urothelial carcinoma. In this study, we aim to fill that gap by evaluating 11 publicly available HFMs across 11 kidney-specific downstream tasks across various staining protocols (PAS, H\&E, PASM, IHC), scales (tile and slide-level), task categories (classification, regression, and copy detection), and outcome types (detection, diagnosis, and prognosis) to provide a comprehensive understanding of HFMs transferability to kidney-specific histology tasks. Tile-level tasks are evaluated using repeated stratified group cross-validation, while slide-level tasks are assessed using the repeated nested stratified cross-validation evaluation framework. Additionally, to assess statistical significance, we performed post-hoc analysis, including the Friedman test, followed by the pairwise Wilcoxon signed-rank test with Holm-Bonferroni correction, and visualization with compact letter display. To promote reproducibility, we additionally release an open-source Python package, kidney-hfm-eval, \url{https://pypi.org/project/kidney-hfm-eval/}, which reproduces slide-level and tile-level evaluation pipelines used in this study. Our findings suggest that HFMs demonstrate moderate to strong performance on tasks driven by coarse, meso-scale renal morphology, including diagnostic classification and detection of prominent structural changes. In contrast, performance consistently deteriorates for tasks requiring fine-grained microstructural discrimination, biologically complex phenotypes, or prognostic inference at the slide level, independent of stain type. Overall, our findings suggest that current HFMs encode static meso-scale structural features and may have limited ability to capture subtle renal pathology or prognosis-related signals. Our results highlight the need for kidney-specific, multi-stain, and multimodal foundation models to support clinically reliable decision-making in nephrology.

\end{abstract}

\keywords{Foundation Models, Kidney, Histopathology}

\begin{document}
\nolinenumbers
\flushbottom
\maketitle
\thispagestyle{empty}

\section{Introduction}

Foundation models (FMs) are large-scale artificial intelligence (AI) models trained with vision transformer (ViT) backbones on large and diverse datasets. This enables them to generalize well across a wide range of downstream tasks such as  classification~\cite{breen2025comprehensive}, regression, image retrieval~\cite{alfasly2025validation}, semantic segmentation~\cite{marza2025thunder}, image captioning~\cite{zhang2020evaluating}, and others. In recent years, many such FMs have been developed in computational digital pathology using histopathology whole slide images (WSIs) and are often referred to as histopathology foundation models (HFMs). Most of these HFMs have been trained using publicly available cancer datasets, such as the Pathology AI Platform (PAIP)~\cite{kim2021paip}, The Cancer Genome Atlas (TCGA)~\cite{weinstein2013cancer}, and the Clinical Proteomic Tumor Analysis Consortium (CPTAC)~\cite{ellis2013connecting}, or using large-scale proprietary datasets. Several of these vision HFMs are designed for tile-level feature extraction, including models such as CTransPath~\cite{wang2022transformer}, REMEDIS~\cite{azizi2023robust}, RudolfV~\cite{dippel2024rudolfv}, Atlas~\cite{alber2025novel}, PLUTO~\cite{juyal2024pluto},  H-optimus-0~\cite{hoptimus0}, H-optimus-1~\cite{hoptimus1}, Hibou-B~\cite{nechaev2024hibou}, Hibou-L~\cite{nechaev2024hibou}, UNI~\cite{chen2024towards}, UNI2-h~\cite{chen2024towards}, Virchow~\cite{vorontsov2024foundation}, Virchow2.~\cite{zimmermann2024virchow2} and models developed by Campanella {\em \textit{et al.}}~\cite{campanella2024computational} (SP22M, SP85M, Campanella-L). Additionally, slide-level feature extractors such as GigaSSL~\cite{lazard2023giga}, MADELEINE~\cite{jaume2024multistain}, CHIEF~\cite{wang2024pathology}, COBRA~\cite{lenz2025unsupervised}, PRISM~\cite{shaikovski2024prism}, Prov-GigaPath~\cite{xu2024whole}, and TITAN~\cite{ding2025multimodal} have also been developed for learning global WSI representations.

Even though HFMs pretrained on H\&E and IHC-stained cancer datasets have demonstrated impressive performance on various downstream tasks based on cancer datasets,~\cite{breen2025comprehensive, neidlinger2025benchmarking, campanella2025clinical, ma2025pathbench}, their utility in kidney-specific, multi-stain settings remains underexplored. In this study, we evaluate the predictive performance of the embeddings extracted using HFMs from WSIs containing kidney-related pathologies. This work is biologically and clinically motivated by the coexistence of chronic kidney diseases (CKDs) with renal cell carcinoma (RCC) and urothelial cancer (UC)~\cite{canter2011prevalence, brooks2024chronic, kompotiatis2019association, tendulkar2022risk, lowrance2014ckd, kitchlu2022cancer, dey2017chronic, mashni2015new}(carcinomas of the bladder, ureters, and renal pelvis). For example, Canter \textit{et al.}~\cite{canter2011prevalence} found that  22\% of patients in the Fox Chase Kidney Cancer Database presented with CKD. From a representation learning standpoint, this work is further motivated by the fact that many HFMs are trained using self-DIstillation with NO labels (DINO)~\cite{caron2021emerging}/DINOv2,~\cite{oquab2024dinov}. These self-supervised learning (SSL) frameworks enforce consistent representations across various augmentations (e.g., color jitter, cropping, and blurring).~\cite{caron2021emerging, oquab2024dinov} Such augmentations have been shown to improve stain invariance and domain robustness, helping HFMs learn semantically meaningful representations rather than overfitting to color distributions.~\cite{tellez2019quantifying, nerrienet2023standardized, ciga2022self}

Given that kidney histopathology requires the use of multiple stains to highlight different structures of the nephron, it is important to evaluate the predictive performance of HFMs across various staining protocols. Accordingly, we benchmark HFMs pretrained on hematoxylin and eosin (H\&E) and immunohistochemistry (IHC) stained WSIs in both within-stain (H\&E, IHC) and cross-stain (Periodic acid–Schiff (PAS), Periodic acid-Schiff-Methenamine silver (PASM)) settings, which is crucial for assessing HFMs generalizability in kidney histopathology. This evaluation helps us determine whether HFM representations capture fundamental tissue morphology rather than superficial color distributions, and whether they transfer effectively to CKD pathology. Thus, evaluating these HFMs across various stains serves as a litmus test for domain invariance of embeddings obtained from HFMs. While prior studies in computational pathology have explored benchmarking HFMs primarily on cancer-centric datasets, kidney-specific evaluations remain limited.~\cite{breen2025comprehensive, neidlinger2025benchmarking, campanella2025clinical, ma2025pathbench}. Recently, a study conducted by Kurata \textit{et al.}~\cite{kurata2025multiple} performed an evaluation of HFMs in kidney-specific pathology, such as acute interstitial fibrosis and diabetic kidney disease, and showed that HFMs can transfer on H\&E-stained kidney data for diagnostic tasks. However, to our knowledge, no prior study has systematically evaluated both the cross-stain transferability and within-stain performance of HFMs in non-cancer settings, particularly within the domain of kidney histopathology.

In this study, we addressed the question of whether the CKD co-existing in RCC and UC datasets used for HFM training is sufficient for their application on CKD WSIs. We also assess the cross-stain transferability of HFMs on kidney WSIs consisting of various kidney-related pathological changes. We evaluated publicly available HFMs pre-trained on more than 100,000 WSIs to maximize exposure to a broad spectrum of histological features, especially RCC and UC. We then benchmarked their performance on kidney-specific tasks at both the tile and slide levels. At the tile level, we examined classification tasks including globally vs. non-globally sclerotic glomeruli on PAS-stained images, glomerular basement membrane (GBM) spike classification on PASM-stained images, inflamed vs. non-inflamed areas on H\&E-stained images, reference vs. abnormal tubule classification in H\&E-stained images, artery stenosis classification in PAS-stained images, a regression task including cell type estimation from H\&E-stained images, and a copy detection task on IHC-stained (CD34 and CD45) images. At the slide level, we evaluated classification tasks such as predicting treatment response (responder vs. non-responder) in membranous nephropathy (MN), diagnosing diabetic nephropathy (DN) cases from reference cases, and predicting estimated glomerular filtration rate (eGFR) decline from one year to three years post-renal transplant, as well as predicting eGFR at one year post-renal transplant, all using PAS-stained WSIs. The main contribution of this work is as follows:

\begin{itemize}
    \item We present a comprehensive assessment of how well current HFMs, originally designed for cancer, generalize to kidney pathologies in within- and cross-stain histopathology WSIs, providing insights into their strengths, limitations, and potential directions.

    \item  We bridge the existing gaps in benchmarking analysis by developing a robust benchmarking pipeline comprising a repeated nested stratified $K$-fold cross-validation framework followed by a rigorous post-hoc analysis, which includes statistical evaluation and comparison of HFMs. The post-hoc analysis includes the Friedman test~\cite{friedman1937use} to detect overall performance differences across HFMs, followed by the pairwise Wilcoxon signed-rank test~\cite{woolson2007wilcoxon} with Holm-Bonferroni correction~\cite{holm1979simple} to control for multiple comparisons, and visualize the results with a compact letter display~\cite{gramm2007algorithms}.

    \item To promote transparency and ease of adoption, we release an open-source Python package that enables researchers and clinicians to benchmark HFMs on their own tasks following our standardized framework. The package is publicly available at \url{https://pypi.org/project/kidney-hfm-eval/} and archived on Zenodo with a DOI~\cite{harishwar_reddy_kasireddy_2026_18882125}.
\end{itemize}

We believe that these findings, together with the open-source package, will inspire and facilitate future research efforts towards the development of HFMs, particularly in the domain of chronic kidney disease. 

\section{Motivation}
Two main hypotheses motivated us to evaluate cancer pre-trained HFMs on kidney pathologies. We first hypothesize whether HFMs have learned any morphological features that are present in CKD. The hypothesis is grounded by the fact that there is co-existence of CKD with cancer (in particular RCC and UC~\cite{canter2011prevalence, brooks2024chronic, kompotiatis2019association, tendulkar2022risk, lowrance2014ckd, kitchlu2022cancer, dey2017chronic, mashni2015new}), therefore, we expect the HFMs pre-trained on cancer WSIs to learn features that also appear in CKD. In our second hypothesis, we evaluate whether HFMs have learned stain-invariant, tissue-relevant morphological features for CKD. Since these HFMs are pretrained using the DINO/DINOv2 SSL framework, their embeddings are expected to exhibit robustness to color shifts and domain variability~\cite{tellez2019quantifying, nerrienet2023standardized, ciga2022self}.

To substantiate the hypothesis of co-existence between cancer and CKD, we discuss several epidemiological studies that have reported co-occurrence and overlapping risk factors between renal malignancies and CKD. For instance, Canter \textit{et al.}~\cite{canter2011prevalence} examined 1114 patients from the Fox Chase Kidney Cancer Database and found that 22\% of cancer patients present with CKD. Consistent with this observation, Brooks \textit{et al.}~\cite{brooks2024chronic} reported an adjusted hazard ratio (aHR) of 1.89 for RCC and 1.35 for UC, indicating 89\% and 35\% higher risks for CKD, respectively. In line with these findings, Kompotiatis \textit{et al.}~\cite{kompotiatis2019association} 
combined 19 studies including 1,931,073 ESKD patients and found estimated incidences of RCC and UC in ESKD patients of 0.3\% (95\%CI: 0.2-0.5\%) and 0.5\% (95\%CI: 0.3-0.8). Similarly, Tendulkar \textit{et al.}~\cite{tendulkar2022risk} showed that among 13,750 patients, 2,758 (20.1\%) developed malignancy with a median time of 8.5 years from CKD diagnosis. Further supporting this association, Lowrance \textit{et al.}~\cite{lowrance2014ckd} showed in a dataset of 1,190,538 patients that decreasing eGFR, a sign of CKD, increased RCC risk (aHR, 1.39; 95\% confidence interval, 1.22 to 1.58 for eGFR=45-59, HR, 1.81; 95\% CI, 1.51 to 2.17 for eGFR=30-44; HR, 2.28; 95\% CI, 1.78 to 2.92 for eGFR<30). Expanding on these population-based findings, Kitchlu  \textit{et al.}~\cite{kitchlu2022cancer} analyzed 5,882,388 individuals with eGFR data, identifying 325,895 cancer diagnoses over 29,993,847 person-years of follow-up corresponding to a cumulative incidence of 10.8\% and 15.3\% among those with kidney disease. Similarly, Dey \textit{et al.}~\cite{dey2017chronic} reported that among 1569 patients undergoing renal cortical tumor surgery, CKD status was low risk in 860 (55\%), moderately increased in 381 (24\%), high in 194 (12\%), and very high in 134 (9\%) patients.  In another surgical cohort, Mashni \textit{et al.}~\cite{mashni2015new} analyzed the kidney cancer surgery cohort 2110, revealing that a substantial proportion had pre-existing CKD before nephrectomy. So, there is high co-existence between RCC and CKD due to shared risk factors (hypertension, diabetes, and nephrotoxic drugs). In fact, one could argue that knowing the prevalence of co-existence in the datasets used to train the HFM models studied here might hint at the degree of generalization of HFM when applied as feature embeddings for models targeted to predict CKD. However, there is no way to assess the prevalence of CKD co-existence in training datasets used for training HFMs. Therefore, evaluating HFMs on downstream tasks specific to CKD will be a way for us to understand and see if the HFMs are trained on enough CKD data or not.

Beyond epidemiologic co-occurrence, our second hypothesis aligns with the representational properties of embeddings obtained from HFMs. Specifically, whether the self-supervised embeddings obtained from HFMs truly capture the domain-invariant morphological features rather than stain-specific color-based features. SSL helps in universal representation learning, where a single pre-trained encoder is adaptable to various tasks (classification, regression, retrieval)~\cite{balestriero2023cookbook} across diverse tissue types. HFMs trained using DINO~\cite{caron2021emerging}/DINOV2~\cite{oquab2024dinov} SSL frameworks 
are meant to learn embeddings invariant to stain appearances because they are trained using various augmentations such as color jitter, gaussian blur, noise, random cropping, rotation, and solarization~\cite{caron2021emerging, oquab2024dinov}. These augmentations try to mimic cross-domain visual shifts~\cite{tellez2018whole}. Pathology workflows are inherently multi-stain and multi-center~\cite{madusanka2023impact}, so variability can occur in staining protocols, scanners, and institutions. Hence, the model that generalizes across these differences is effectively domain agnostic, i.e., it captures biologically invariant features that persist across various stains. If an H\&E-trained HFM performs well on PAS or PASM-stained images with just probing, it implies the model’s latent space captures morphological primitives (e.g., nuclei, tubules, glomeruli) rather than superficial stain color cues. Hence, evaluating HFMs on cross-stain  (PAS, PASM) and within-stain (H\&E, IHC) histology provides a direct, quantitative test of whether self-supervised invariances learned under color and spatial augmentations actually generalize to staining variations not seen during training.

Several studies have evaluated the performance of HFMs on various datasets. However, current HFM benchmarks~\cite{breen2025comprehensive, neidlinger2025benchmarking, campanella2025clinical, ma2025pathbench}  are more cancer-centric, often limited to H\&E WSIs. By carrying out the evaluation of our two hypotheses in kidney histopathology, we test the latent robustness of embeddings and limitations of current HFMs, guiding future models that can explicitly incorporate multi-stain invariance or stain‐aware adaptation modules. Building on these insights, our work extends the evaluation to multi-stain, multi-task kidney pathology through data obtained from multiple institutions.

\section{Histopathology Vision Foundation Models}
In this study, we mainly focus on HFMs that have been pretrained on more than 100,000 WSIs. This ensures that models have been exposed to a sufficiently diverse and representative set of histological patterns during pretraining, especially RCC and UC. This is crucial for evaluating their generalizability to new tasks and domains, particularly in kidney pathology. Based on this criterion, we include the following HFMs for tile and slide level tasks: UNI, UNI2-h, Virchow, Virchow2, H-optimus-0, H-optimus-1, Hibou-B, Hibou-L, SP22M, SP85M, and Prov-Gigapath. These models span a diverse range of architectures, training datasets, and SSL strategies, all optimized to extract meaningful representations from high-resolution WSIs. In the following, we provide a description of each HFM, also described in Table~\ref{tab:FM_description}.

\begin{table}[h!]
\centering
\small
\begin{tabular}{|>{\centering\arraybackslash}p{1.8cm}|>{\centering\arraybackslash}p{1.25cm}|>{\centering\arraybackslash}p{1.6cm}|>{\centering\arraybackslash}p{2.5cm}|>{\centering\arraybackslash}p{2cm}|>{\centering\arraybackslash}p{1.85cm}|>{\centering\arraybackslash}p{1.5cm}|>{\centering\arraybackslash}p{1.2cm}|>{\centering\arraybackslash}p{1.2cm}|}
\hline
\textbf{Model Name} & \textbf{Encoder Type} & \textbf{Dataset Size} & \textbf{Pretraining Image Type} & \textbf{SSL Strategy} & \textbf{Image Magnification}  & \textbf{Number of Parameters} & \textbf{Feature Dimension} \\
\hline
H-optimus-0~\cite{hoptimus0} & ViT-G & 500k $\dagger$  & H\&E & DINOv2~\cite{oquab2024dinov} & 20x  & 1.10B & 1536* \\
\hline
H-optimus-1~\cite{hoptimus1} & ViT-G & 1M $\dagger$  & H\&E & DINOv2 & 20x  & 1.10B & 1536* \\
\hline
Hibou-B~\cite{nechaev2024hibou} & ViT-B & 512M$\dagger\dagger$ & H\&E, non-H\&E, and cytology & DINOv2 & Not reported & 86M & 768* \\
\hline
Hibou-L~\cite{nechaev2024hibou} & ViT-L & 1.2B$\dagger\dagger$ & H\&E, non-H\&E, and cytology & DINOv2 & Not reported  & 303M & 1024* \\
\hline
Prov-Gigapath~\cite{xu2024whole} & ViT-G + LongNet~\cite{ding2023longnet} & 171k$\dagger$  & H\&E and IHC & DINOv2 + MAE~\cite{he2022masked} & 20x  & 1.10B & 1536*, 768** \\
\hline
SP22M~\cite{campanella2024computational} & ViT-S & 423k$\dagger$ & H\&E & DINO~\cite{caron2021emerging} & 20x  & 22M & 384** \\
\hline
SP85M~\cite{campanella2024computational} & ViT-B & 423k$\dagger$ & H\&E  & DINO & 20x  & 86M & 768* \\
\hline
UNI~\cite{chen2024towards} & ViT-L & 100k$\dagger$ & H\&E & DINOv2  & 20x & 303M & 1024* \\
\hline
UNI2-h~\cite{chen2024towards} & ViT-H & 350k$\dagger$ & H\&E and IHC & DINOv2 & Not reported & 681M & 1536* \\
\hline
Virchow~\cite{vorontsov2024foundation} & ViT-H & 1.5M$\dagger$ & H\&E & DINOv2 & 20x  & 632M & 2560* \\
\hline
Virchow2~\cite{zimmermann2024virchow2} & ViT-H & 3.1M$\dagger$ & H\&E and IHC & DINOv2 & 5x, 10x, 20x, 40x  & 632M & 2560* \\
\hline
\end{tabular}
\caption{\label{tab:FM_description}\textbf{Summary of vision histopathology foundation models (HFMs) evaluated in this study.} Each model varies in architecture type, pretraining dataset scale (no. of tiles), pretraining image modality (i.e., type of stain), SSL strategy, magnification levels of images and their resolution (i.e., number of pixels), number of HFM parameters, and output feature dimensionality.\\
\textit{Note:} * indicates tile-level embedding dimension; ** indicates slide-level embedding dimension; $\dagger$ indicates number of WSIs;$\dagger\dagger$ indicates number of tiles.}
\end{table}

\begin{itemize}
    \item H-optimus-0~\cite{hoptimus0} HFM is trained with a ViT-G/14 architecture comprising 1.10 billion parameters, trained on a proprietary dataset comprising over 500,000 H\&E-stained WSIs from various body regions, from more than 200,000 patients. This model is trained on tiles at 20× magnification, using the DINOv2 SSL strategy, which is based on a student-teacher distillation framework. The model combines two main loss functions: a self-distillation loss, masked image modeling loss, and KoLeo regularization~\cite{sablayrolles2018spreading}.

    \item H-optimus-1~\cite{hoptimus1} HFM is also trained with a ViT-G/14 architecture comprising more than one million WSIs from more than 800,000 patients obtained from three different scanner types from more than 50 organ tissues. This model is also trained on more than two billion tiles at 20× magnification, using the DINOv2 SSL strategy.

    \item Hibou-B~\cite{nechaev2024hibou} HFM is trained with ViT-B/14 architecture comprising 86 million parameters, trained on a proprietary dataset consisting of 936,441 H\&E slides and 202,464 non-H\&E slides, all derived from 306,400 unique cases, along with 2,676 cytology slides. This model utilizes  a total of 512 million tiles of size 224×224 using the customized DINOv2 framework.

    \item Hibou-L~\cite{nechaev2024hibou} is an extended version of Hibou-B, trained on the same dataset but scaled to 1.2 billion tiles using a ViT-L/14 architecture, having 307 million parameters, again employing customized DINOv2 for SSL.

    \item Prov-Gigapath~\cite{xu2024whole} is a multi-scale HFM designed to generate embeddings at both tile and slide levels. The model introduces Gigapath, a novel ViT architecture built upon LongNet~\cite{ding2023longnet}, which aggregates tile-level embeddings to construct a holistic slide-level representation. It was trained on the Prov-Path dataset, a proprietary collection of 171,189 WSIs from Providence, a large U.S.\ healthcare network with 28 cancer centers. In total, the dataset includes 1.38 billion tiles at 20× magnification, obtained from H\&E and IHC-stained slides. These slides originate from over 30,000 patients, spanning 31 major tissue types. The tile encoder is pretrained using DINOv2, while the slide encoder is trained using a masked autoencoder (MAE)~\cite{he2022masked} strategy applied within the LongNet framework. However, in this paper, we only use the tile encoder to maintain consistency with other HFMs.

    \item SP22M~\cite{campanella2024computational} HFM is trained with a ViT-S backbone, comprising 22 million parameters, while SP85M HFM adopts a ViT-B architecture. SP22M and SP85M were both pretrained on an identical collection of 423,000 H\&E WSIs (over 3.2 billion tiles) sourced from 88,035 cases (76,794 patients) at the Mount Sinai Health System. This dataset spans 70 organ types and a wide range of pathologies. Both the models are trained with the DINO SSL framework~\cite{caron2021emerging}.

    \item UNI~\cite{chen2024towards} HFM is trained with a ViT-L/16 backbone comprising 300 million parameters and trained on the Mass-100k dataset, which includes approximately 100 million tiles extracted from 100,426 H\&E-stained WSIs spanning 20 tissue types. The training data was sourced from the Genotype–Tissue Expression (GTEx) consortium \cite{gtex2015genotype} as well as internal datasets from Massachusetts General Hospital and Brigham and Women’s Hospital. UNI employs the DINOv2 \cite{oquab2024dinov} SSL training strategy.

    \item UNI2-h~\cite{chen2024towards} HFM is an extended version of UNI, using a larger ViT-H/14 architecture and trained on over 200 million tiles, including both H\&E and IHC modalities. These images were sampled from more than 350,000 diverse WSIs collected from the Mass General Brigham health system. Similar to UNI, UNI2-h also utilizes the DINOv2 SSL strategy.

    \item Virchow~\cite{vorontsov2024foundation} HFM is trained with a ViT-H/14 architecture with 632M parameters and trained on a proprietary dataset consisting of 1,488,550 H\&E slides from 1,207,837 formalin-fixed paraffin-embedded (FFPE) samples, which in turn originated from 392,268 tissue samples across 208,815 patients. These were drawn from a cohort of 119,629 patients, spanning 17 cancer types. The model is trained with DINOv2 framework.

    \item Virchow2~\cite{zimmermann2024virchow2} HFM builds upon Virchow, using the same ViT-H/14 architecture, but with a significantly larger training set. It was trained on 3,134,922 H\&E and IHC slides derived from 2,078,093 FFPE samples, covering 871,025 tissue samples and 493,332 patient cases from a total of 225,401 patients. Virchow2 follows the DINOv2 multi-view strategy. Unlike Virchow, which is trained only on 20× magnification, Virchow2 incorporates multi-scale inputs from 5×, 10×, 20×, and 40× magnifications, enabling better generalization across variable resolutions.

\end{itemize}

Table~\ref{tab:FM_description} provides the compact representation of various details pertaining to each FM. Together, these models represent some of the well-known HFMs available in digital pathology and are evaluated in our study for their ability to generalize to kidney-specific downstream tasks.

\section{Significance of Evaluation Tasks}
In this section, we describe the datasets used in this study and highlight their clinical importance and relevance for benchmarking the performance of HFMs in renal histopathology. The evaluation tasks are designed to systematically assess the extent to which HFMs can capture clinically meaningful and morphologically relevant features across multiple spatial resolutions. We employ both tile-level and slide-level tasks to comprehensively evaluate the discriminative and generalization capabilities of HFMs. The tile-level tasks focus on localized histological patterns such as glomerulosclerosis, GBM changes, tubular abnormalities,  arterial stenosis, and inflammatory activity, providing insights into the model's ability to recognize disease-defining microstructures across varying levels of morphological complexity. In contrast, the slide-level tasks assess global contextual understanding by aggregating features across tiles to predict clinically or biologically relevant outcomes. Together, these complementary evaluation tasks offer a robust framework for benchmarking HFMs in terms of their pathology-aware representation quality and potential for clinical translation in kidney histopathology. In the following, we provide a detailed explanation of each dataset used in this study for benchmarking HFMs, also concisely described in Table~\ref{tab:evaluation_tasks}.
  
\subsection{\textit{\textbf{Tile-level tasks}}}

To systematically assess the effectiveness of HFMs in capturing clinically relevant features, we conducted a total of seven tile-level classification tasks. The classification tasks in this study are designed to evaluate the ability of HFMs to capture clinically meaningful features across a diverse range of morphological changes in kidney pathologies. We focus on four binary classification tasks: globally vs. non-globally sclerotic glomeruli, GBM spike vs. no-GBM spike, reference vs. abnormal tubule, 
inflammation vs. non-inflammation classification, a three-class classification for arteriolar stenosis classification, a cell type estimation regression task, and a copy detection problem. A concise summary of these evaluation tasks, including the number of slides, number of tiles, data sources, number and distribution of classes, and the staining protocols used, is provided in Table~\ref{tab:evaluation_tasks}. For tile-level classification tasks, we employed both linear and $k$-nearest neighbor ($k$NN) probing to evaluate the discriminative capacity of HFM embeddings. The cell-type estimation task was formulated as a regression problem and addressed using ridge regression. For the copy detection task, representational robustness was assessed by quantifying the cosine similarity between original and augmented embeddings. These tasks aim to capture a wide range of morphological variations, which can help in precise patient stratification by informing about various chronic kidney disease changes.

\begin{table}[h!]
\centering
\small
\begin{tabular}{|>{\centering\arraybackslash}p{3.6cm}|>{\centering\arraybackslash}p{0.8cm}|>{\centering\arraybackslash}p{1.6cm}|>{\centering\arraybackslash}p{1cm}|>{\centering\arraybackslash}p{1cm}|>{\centering\arraybackslash}p{2cm}|>{\centering\arraybackslash}p{0.8cm}|>{\centering\arraybackslash}p{1.6cm}|>{\centering\arraybackslash}p{1cm}|}
\hline
\textbf{Evaluation Task} & Level & \textbf{Task type} & \textbf{Number of Slides} & \textbf{Number of tiles} & \textbf{Data source} & \textbf{Number of classes}  & \textbf{Class ratio} & \textbf{Staining Protocol} \\
\hline
Globally vs. non-globally sclerotic glomerulus classification & Tile & Classification & 215 & 5761 & WUSTL, Vanderbilt University  & 2 & 1:4.46 & PAS \\
\hline
GBM vs. no GBM spike classification & Tile & Classification & 63 & 695 & NIH/ NIDDK & 2 & 1:1.93 & PASM \\
\hline
Reference vs. abnormal tubule classification & Tile & Classification& 12 & 1200  & Indiana University & 2 & 1:1 & H\&E \\
\hline
Inflammation vs. no inflammation classification & Tile & Classification & 14 & 8245 & UC Davis & 2  & 1.26:1 & H\&E \\
\hline
Arterial stenosis classification (mild, moderate, severe) & Tile & Classification & 155 & 242 & University of Coimbra, UC Davis & 3 & 5.32:2.08:1  & PAS \\
\hline
Cell type proportion estimation & Tile & Regression & 12 & 17,478 & Indiana University & NA  & NA & H\&E \\
\hline
Copy detection (deform, color, geo, noise ) & Tile &  Classification & 5 & 2405 & JHU & 4  & 1:1:1:1 & CD34, CD45 \\
\hline
Reference vs. DN & Slide & Classification & 137 & NA & WUSTL  & 2 & 1:1.79 & PAS\\
\hline
MN treatment response prediction (response vs. no response)& Slide & Classification & 85 & NA &                    NIH/ NIDDK  & 2 & 5.07:1 & PAS\\
\hline
Renal transplant eGFR decline prediction (no decline vs. decline) & Slide & Classification &  132 & NA & JHU, UC Davis  & 2 & 1.28:1 & PAS\\
\hline
Renal transplant eGFR prediction at one year & Slide & Regression & 204 & NA & JHU, UC Davis  & NA & NA & PAS\\
\hline
\end{tabular}
\caption{\label{tab:evaluation_tasks}\textbf{Summary of datasets being used for evaluating histopathology foundation models.} \textbf{Abbreviations:} WUSTL =  Washington University in St. Louis, NIH = National Institutes of Health, NIDDK = National Institute of Diabetes and Digestive and Kidney Diseases, UC Davis = University of California Davis, JHU = Johns Hopkins University, NA = Not Applicable, GBM = Glomerular Basement Membrane, eGFR = estimated Glomerular Filtration Rate, MN = Membranous Nephropathy, DN = Diabetic Nephropathy.}
\end{table}
\begin{figure}[H]
\centering
\begin{subfigure}[t]{0.25\linewidth}
    \centering
    \includegraphics[width=\linewidth]{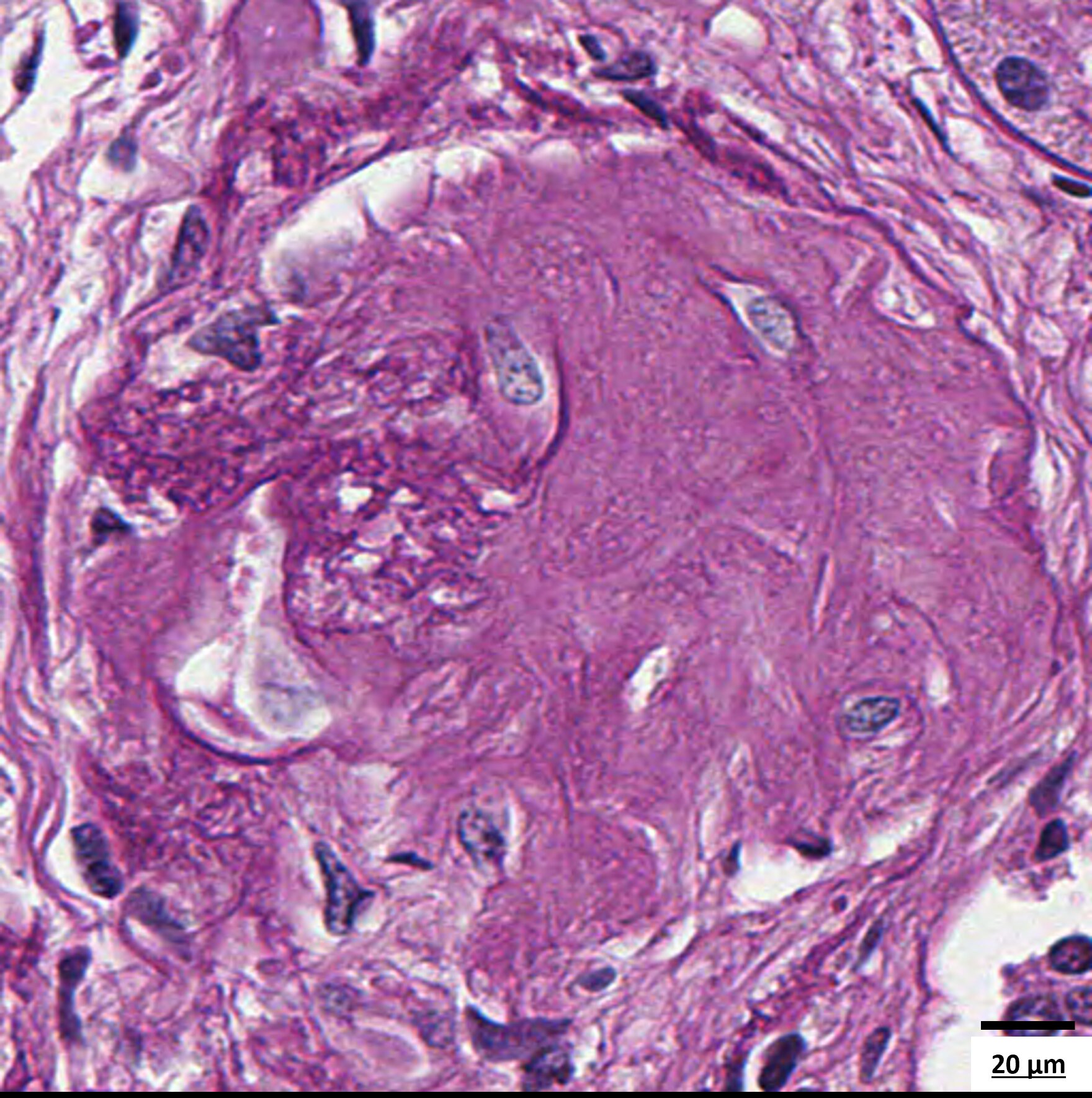}
    \caption{Globally sclerotic glomerulus.}
    \label{fig:globally_sclerotic}
\end{subfigure}
\hspace{0.04\linewidth}
\begin{subfigure}[t]{0.25\linewidth}
    \centering
    \includegraphics[width=\linewidth]{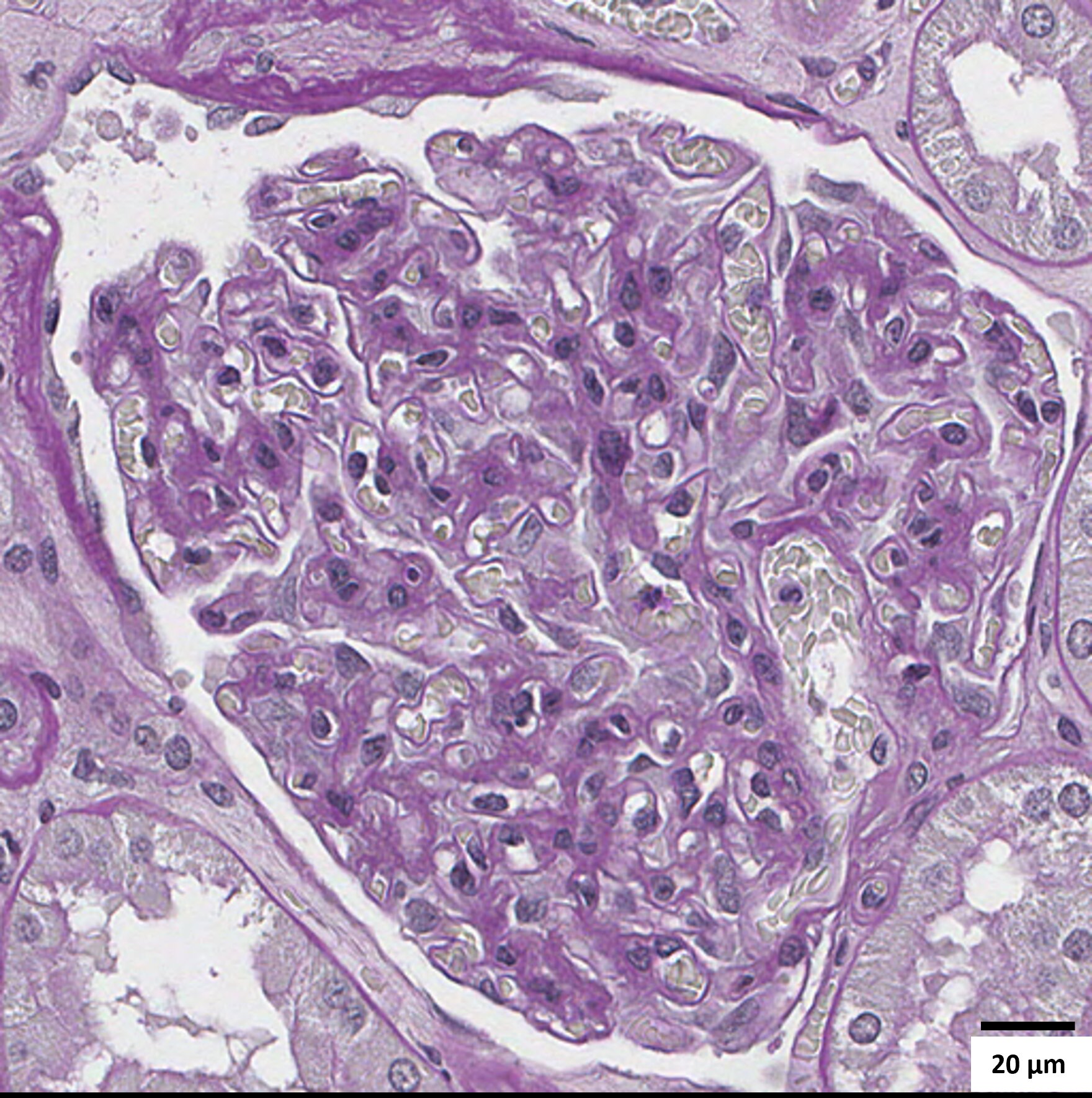}
    \caption{Non-globally sclerotic glomerulus.}
    \label{fig:non_globally_sclerotic}
\end{subfigure}
\caption{Representative histopathology examples of glomerulosclerosis.
}
\label{fig:glomerular_sclerosis}
\end{figure}
\bigskip
\textit{\textbf{Classification tasks}}

\textit{\textbf{Globally vs. non-globally sclerotic glomerulus classification:}}  
This binary classification task involves classifying a glomerulus as either globally sclerotic (Figure~\ref{fig:globally_sclerotic}) or non-globally sclerotic (Figure~\ref{fig:non_globally_sclerotic}). To generate ground truth, we utilize a multi-compartment segmentation (MCS) algorithm, as detailed in our previous work~\cite{mimar2024compreps}, to segment and extract both globally sclerotic and non-globally sclerotic glomeruli from the WSIs using the Detectron2 Deeplabv3~\cite{chen2017rethinking} model. We utilized 215 PAS-stained WSIs acquired at 0.25 micrometer per pixel (mpp) from Washington University in St. Louis (WUSTL) and Vanderbilt University, from which we extracted 5,761 glomeruli, of which 4,704 are non-globally sclerotic and 1,054 are globally sclerotic. Globally sclerotic glomeruli show complete loss of normal morphology, with extensive fibrosis and capillary tuft collapse, along with PAS-positive basement membrane material~\cite{haas2020consensus}, indicating advanced glomerular injury. In contrast, non-globally sclerotic glomeruli preserve their capillary loops and structural integrity, reflecting healthier tissue. Distinguishing between these two categories is essential for evaluating chronic kidney damage, as the proportion of globally sclerotic glomeruli abnormal for age correlates with renal dysfunction and poor renal outcomes~\cite{nasri2012association, chung2020age, niu2019predominant}.

\begin{figure}[H]
\centering
\begin{subfigure}[t]{0.25\linewidth}
    \centering
    \includegraphics[width=\linewidth]{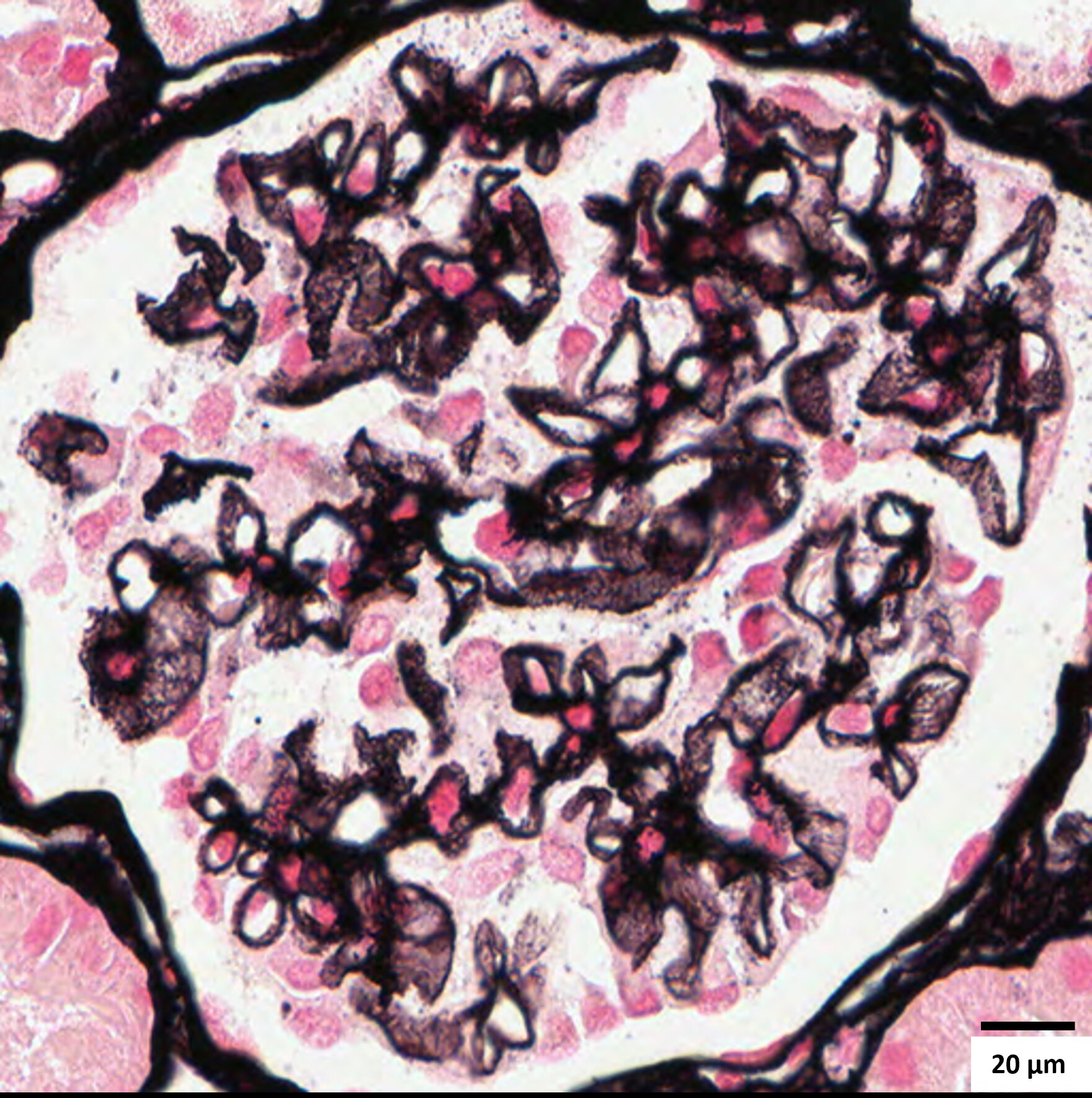}
    \caption{Glomerulus with GBM spike.}
    \label{fig:gbm_spike}
\end{subfigure}
\hspace{0.04\linewidth}
\begin{subfigure}[t]{0.25\linewidth}
    \centering
    \includegraphics[width=\linewidth]{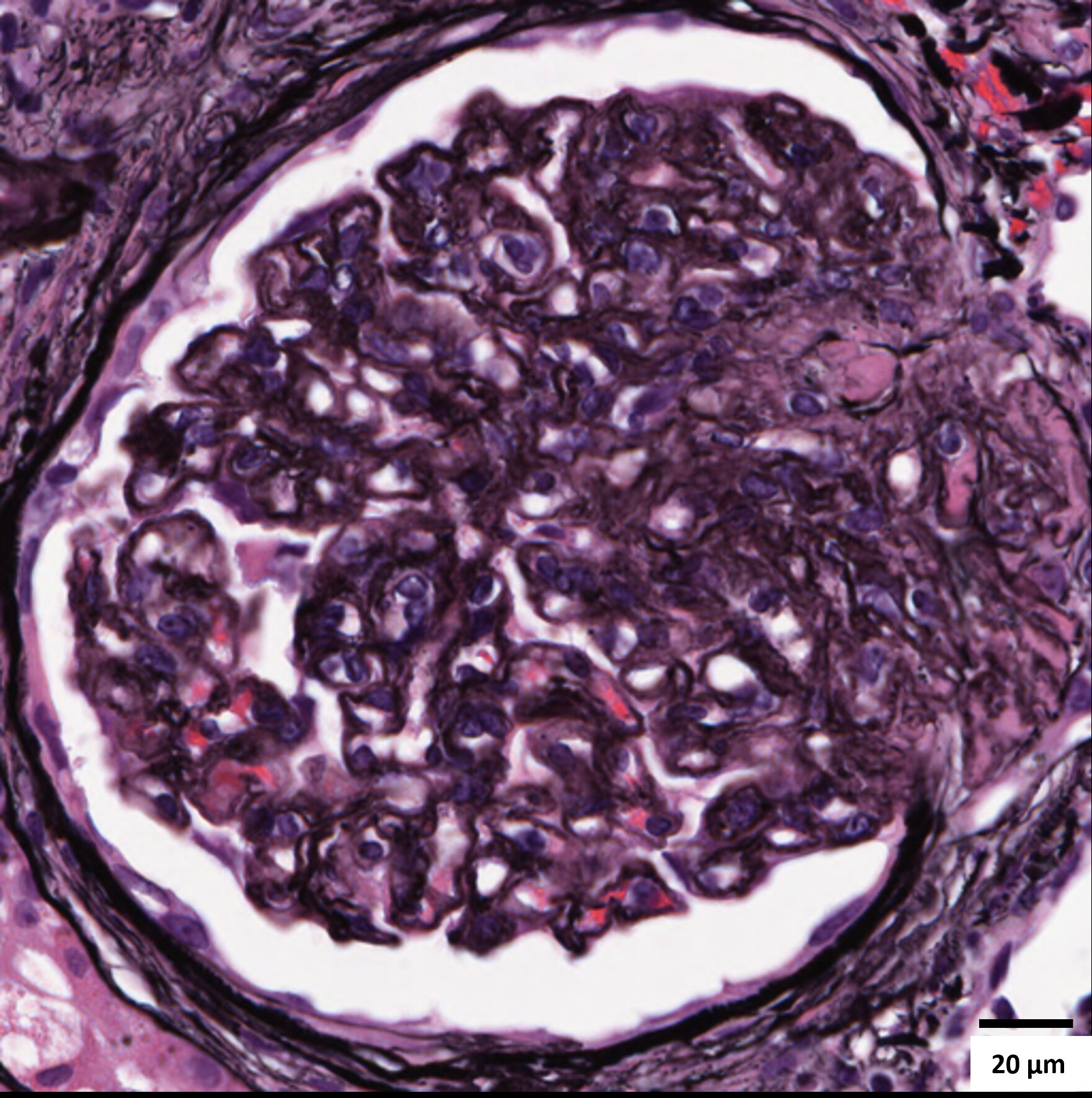}
    \caption{Glomerulus without GBM spike.}
    \label{fig:no_gbm_spike}
\end{subfigure}
\caption{Representative histopathology examples of GBM spike detection.}
\label{fig:glomerular_basement_membrane_spike}
\end{figure}

\textit{\textbf{GBM spike classification:}} This binary classification task involves classifying glomeruli with a GBM spike (Figure~\ref{fig:gbm_spike}) vs. without a GBM spike (Figure~\ref{fig:no_gbm_spike}). As in the previous task, to generate ground truth, we use the MCS model to segment and extract glomeruli from the PASM-stained WSIs. The extracted glomeruli are manually annotated by a renal pathologist based on the presence or absence of GBM spikes. We utilized 63 PASM-stained WSIs acquired at 0.25 mpp from the National Institutes of Health, National Institute of Diabetes and Digestive and Kidney Diseases (NIH/NIDDK), from which we extracted 695 glomeruli, of which 237 have GBM spikes, and the remaining 458 are without GBM spikes. Accurately classifying GBM spikes is essential since these projections along the basement membrane define MN~\cite{chung2022membranous, li2019clinical}, correlate with immune complex deposition, directly impact diagnosis, provide reliable disease staging~\cite{xiaofan2019new}, and provide prognostic assessment~\cite{troyanov2006renal}.

\begin{figure}[H]
\centering
\begin{subfigure}[t]{0.25\linewidth}
    \centering
    \includegraphics[width=\linewidth]{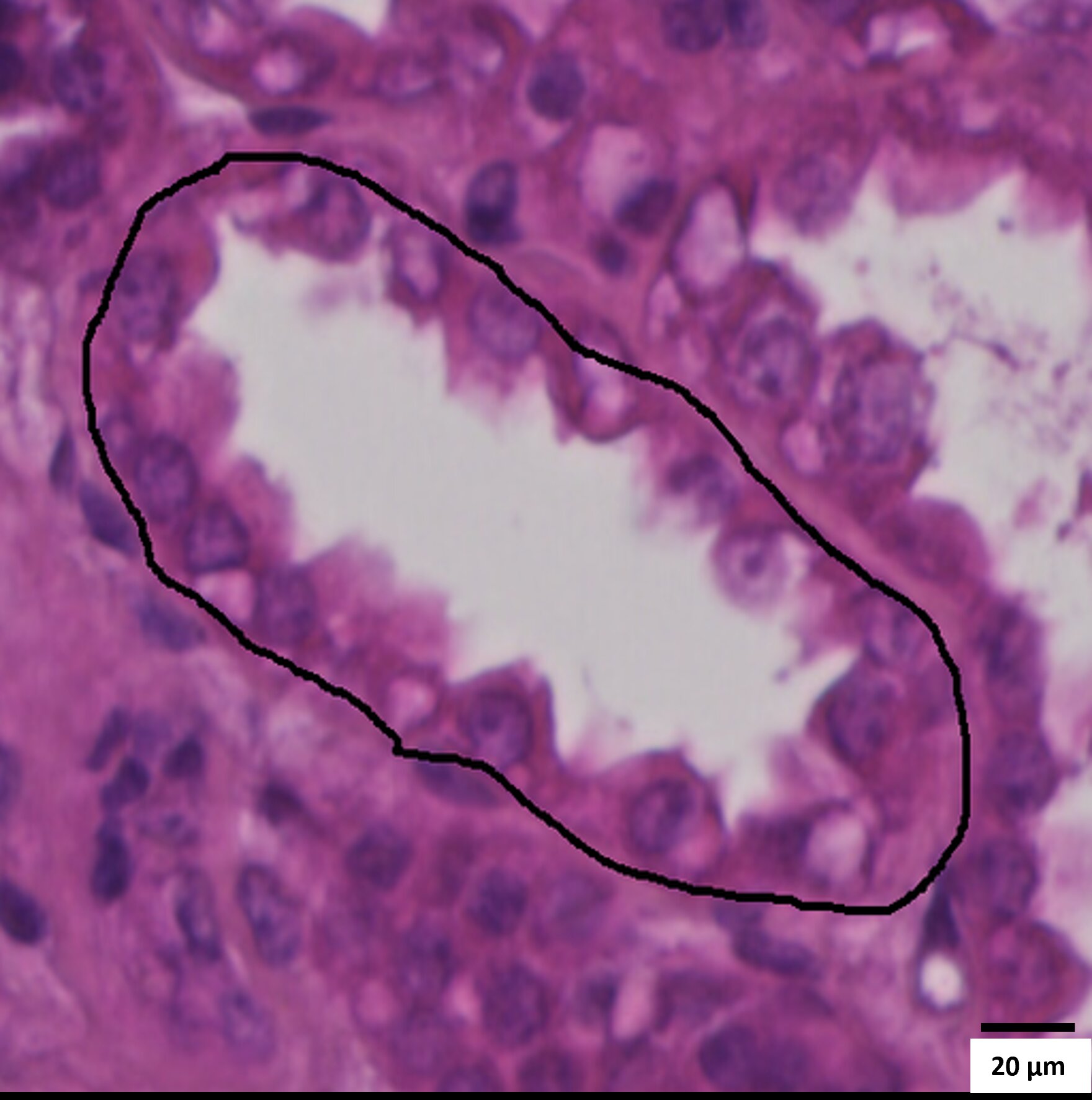}
    \caption{Reference tubule.}
    \label{fig:Reference_tubule}
\end{subfigure}
\hspace{0.04\linewidth}
\begin{subfigure}[t]{0.25\linewidth}
    \centering
    \includegraphics[width=\linewidth]{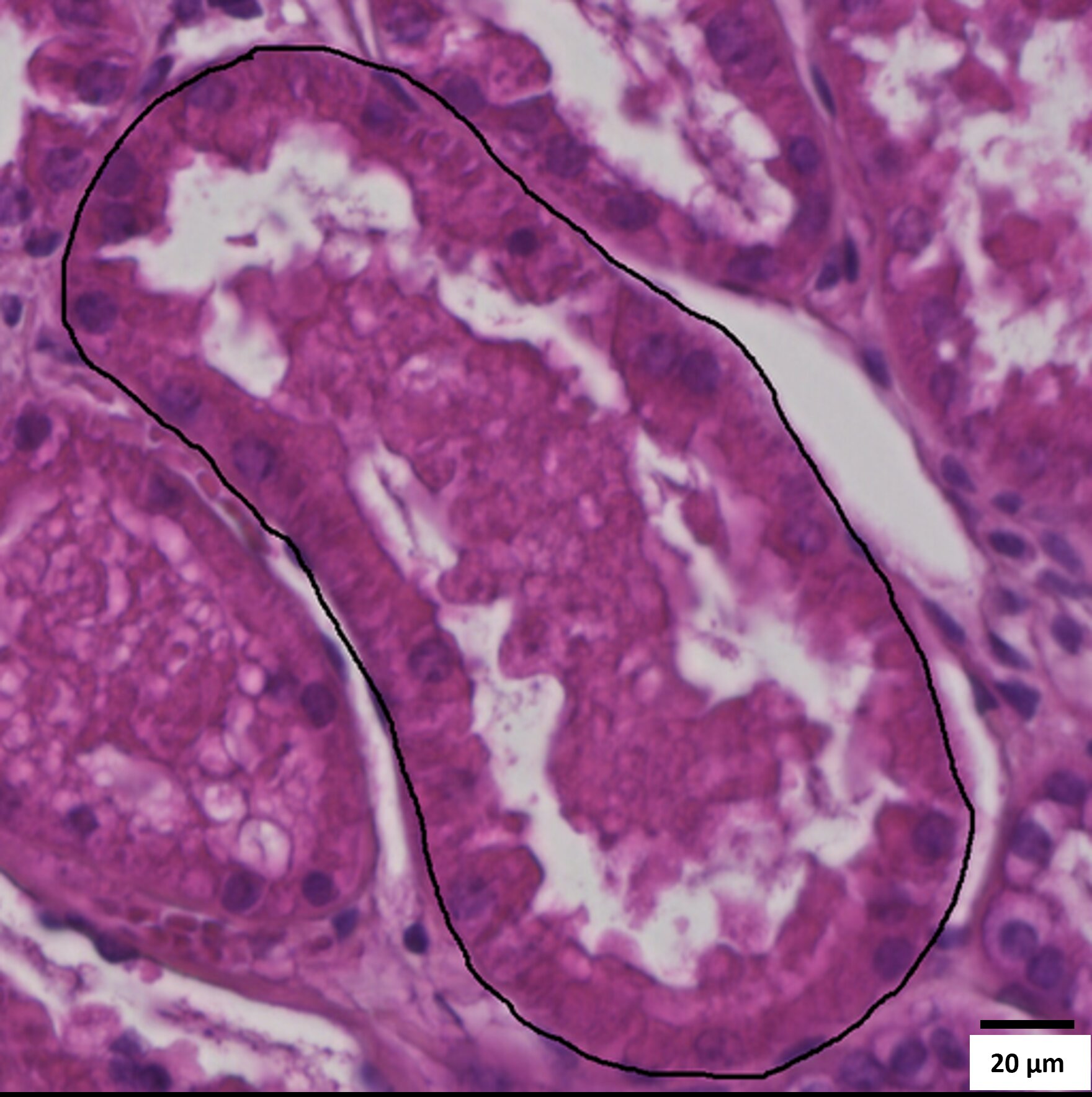}
    \caption{Abnormal tubule.}
    \label{fig:Abnormal_tubule}
\end{subfigure}
\caption{Representative histopathology examples reference and abnormal tubules.}
\label{fig:tubule_comparison}
\end{figure}

\textit{\textbf{Reference vs. abnormal tubule classification:}}  
This binary classification task involves classifying a tubule as either a reference tubule (Figure ~\ref{fig:Reference_tubule}) or an abnormal tubule (Figure ~\ref{fig:Abnormal_tubule}). We segmented the tubules using the same MCS model~\cite{mimar2024compreps} as used in the globally sclerotic glomerulus classification task. A total of 12 FFPE H\&E WSIs were used, of which four are DN cases, and eight are reference cases from Indiana University. Each of these 12 samples comes with the same section \emph{10X Visium} spatial transcriptomics (ST) data. We randomly selected 1200 tubules, 100 from each WSI, of these 1200 tubules, 600 are reference tubules, and the remaining 600 are abnormal tubules. We derived the labels using Visium ST data in conjunction with histology-derived segmented renal tubules. Tubules were labeled "reference" if the majority of epithelial cells were in the reference state; otherwise, they were treated as "abnormal". More details on ground-truth generation are available in \textit{Supplementary Section~\ref{sup:sec:tubule_gt}}. Abnormal tubules typically exhibit features such as brush border loss, tubular atrophy, necrosis, proteinaceous casts, and basement membrane thickening~\cite{floyd2024acute}, all of which signify tubular injury. Accurately classifying these subtle morphologic alterations is critical, as the degree of tubular injury strongly correlates with renal function decline, fibrosis severity, and patient prognosis~\cite{liu2018renal}. Consequently, this task evaluates the ability of HFMs to capture molecular information derived from ST data, thereby serving as a sensitive indicator of model performance in clinically relevant feature learning.

\begin{figure}[H]
\centering
\begin{subfigure}[t]{0.25\linewidth}
    \centering
    \includegraphics[width=\linewidth]{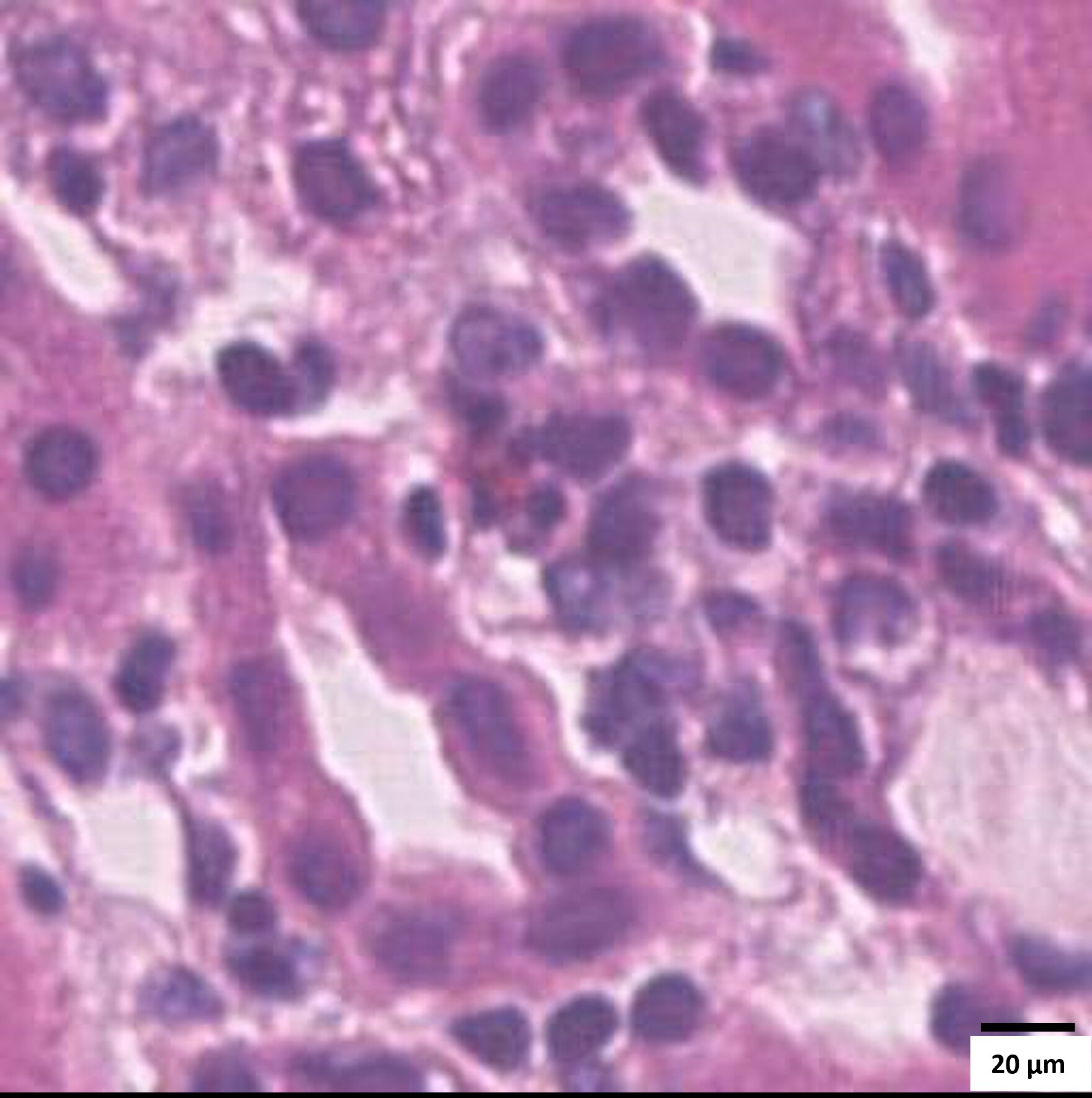}
    \caption{Inflamed region}
    \label{fig:inflamed}
\end{subfigure}
\hspace{0.04\linewidth}
\begin{subfigure}[t]{0.25\linewidth}
    \centering
    \includegraphics[width=\linewidth]{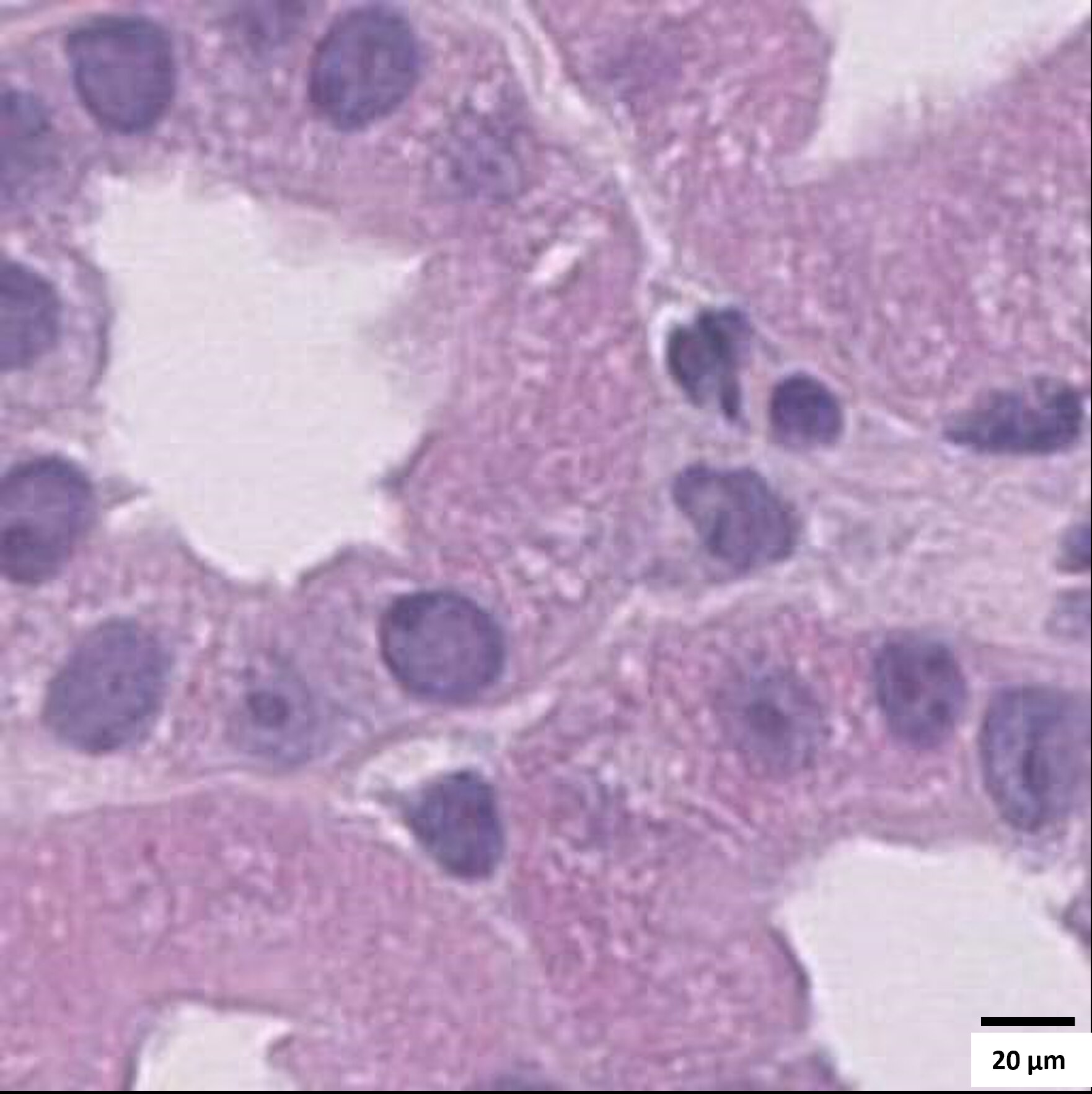}
    \caption{Non-inflamed region}
    \label{fig:not_inflamed}
\end{subfigure}
\caption{Representative histopathology examples with and without inflammation.}
\label{fig:inflammation_comparison}
\end{figure}

\textit{\textbf{Inflammation classification:}}  
This binary classification task involves classifying the images into inflamed (Figure~\ref{fig:inflamed}) vs. non-inflamed (Figure~\ref{fig:not_inflamed}). We used data obtained from the University of California Davis (UC Davis) Medical Center, including 14 patients with 42 WSIs (three slides for each patient), acquired at 0.25 mpp. Three adjacent sections are sliced from the block and stained with H\&E, CD45-IHC, and PAS in the respective order. We treated the CD45-IHC-stained WSI as ground truth and manually mapped it onto the H\&E-stained WSI by applying the affine transformation matrix obtained from the registration of the two WSIs. This step requires manual tuning to ensure that the CD45-IHC-stained WSI is aligned with the landmarks on the H\&E-stained WSI. We performed analysis on these 14 WSIs, from which we extracted around 8,245 tiles of size 224 $\times$ 224 px at 0.5 mpp, comprising 4,603 inflamed and 3,642 non-inflamed tiles. Interstitial inflammation in renal tissue is an indicator of disease activity in conditions such as lupus nephritis~\cite{hsieh2011tubulointerstitial} and reduced survival in renal transplant~\cite{sellares2011inflammation}. The inflamed region shows interstitial infiltration by leukocytes, signifying an ongoing immune-mediated injury~\cite{tang2018lupus}. In comparison, the non-inflamed region appears cell-sparse, with minimal to no immune cell presence, indicating tissue that is either reference or inactive. Accurate recognition of inflammation is crucial for assessing disease severity, guiding immunosuppressive therapy, and monitoring patient response during follow-up~\cite{parodis2020prediction}.

\begin{figure}[H]
\centering
\begin{subfigure}[t]{0.25\linewidth}
    \centering
    \includegraphics[width=\linewidth]{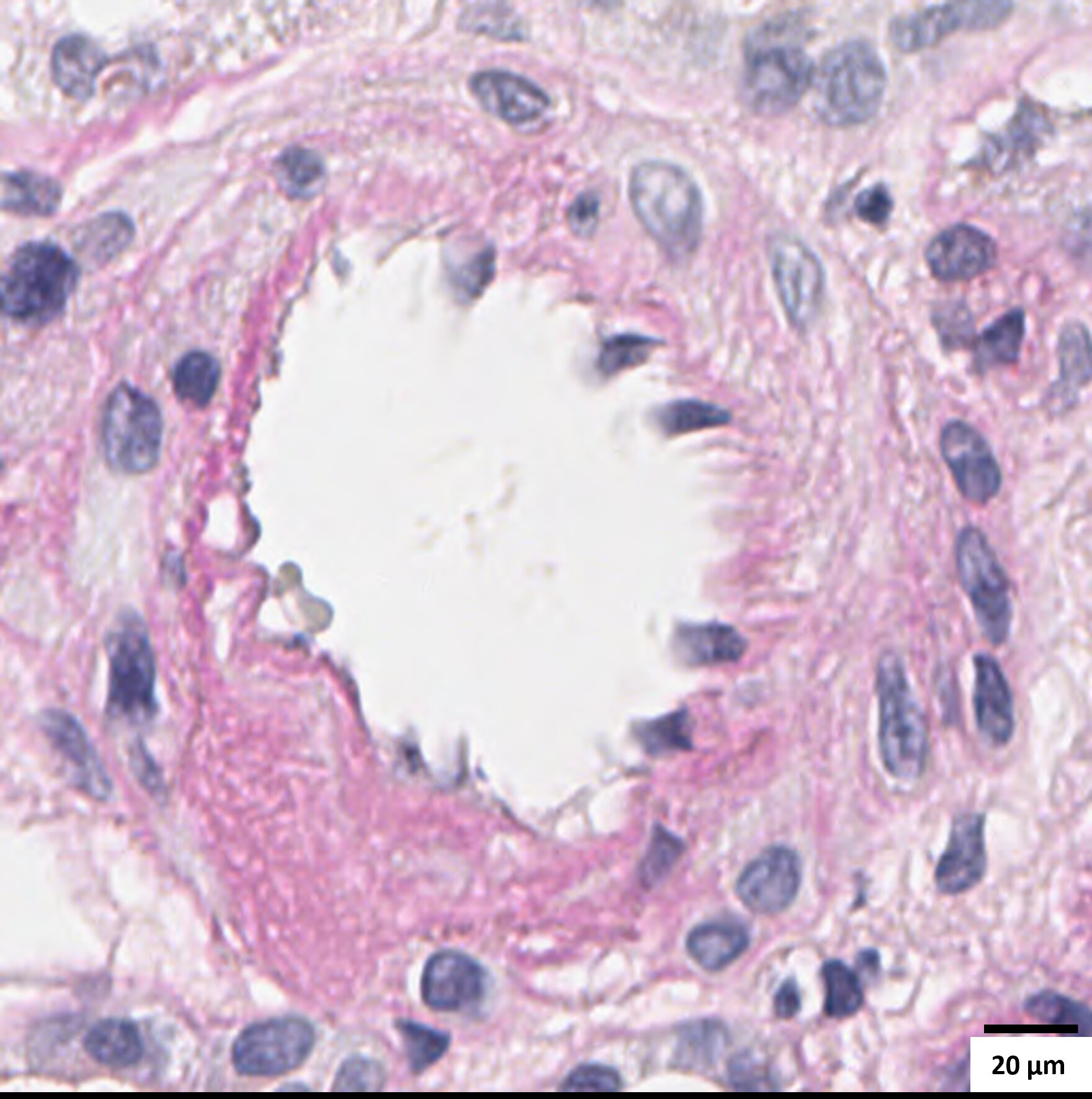}
    \caption{Mild stenosis artery}
    \label{fig:mild_stenosis}
\end{subfigure}
\hspace{0.04\linewidth}
\begin{subfigure}[t]{0.25\linewidth}
    \centering
    \includegraphics[width=\linewidth]{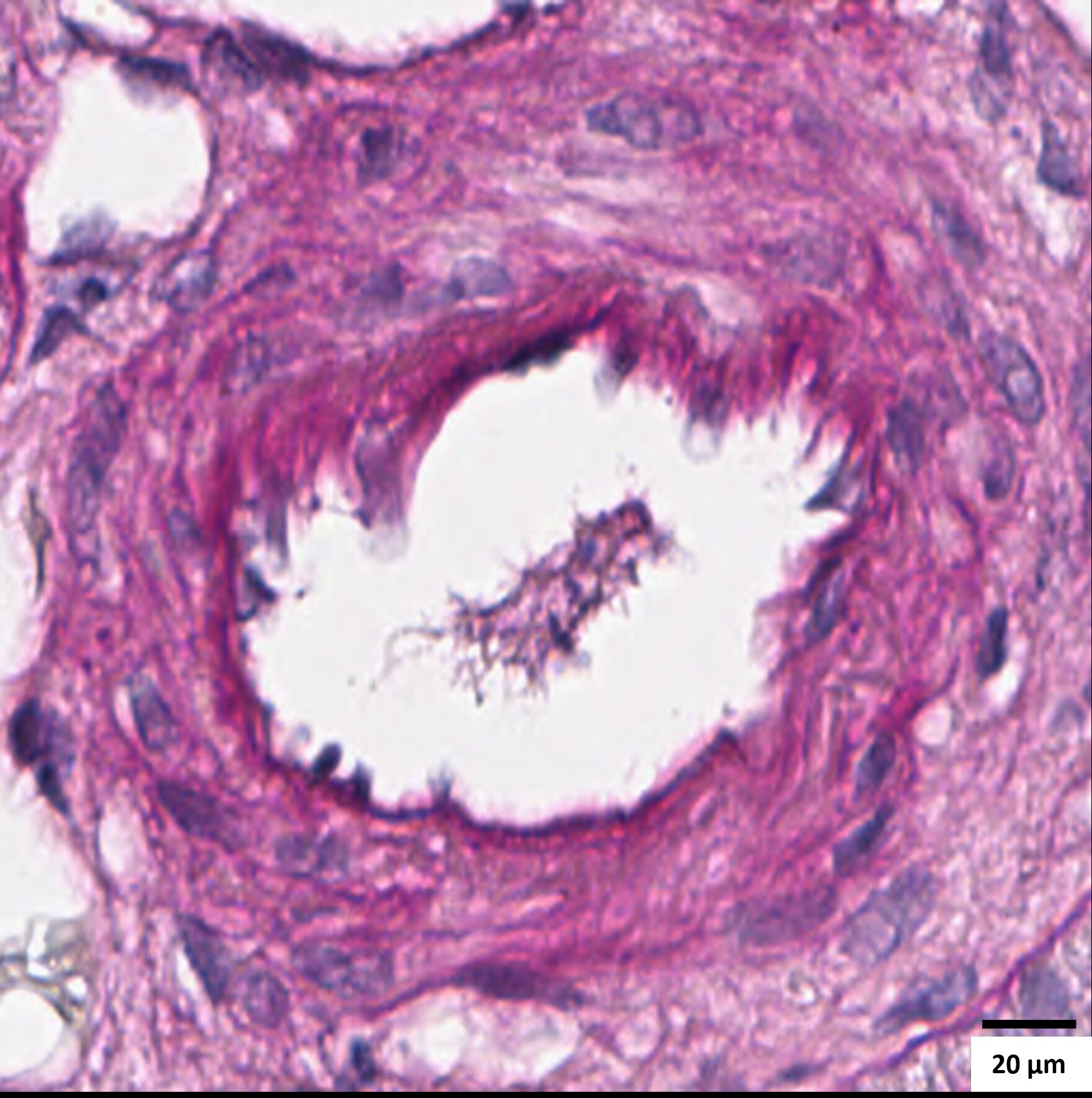}
    \caption{Moderate stenosis artery}
    \label{fig:moderate_stenosis}
\end{subfigure}
\hspace{0.04\linewidth}
\begin{subfigure}[t]{0.25\linewidth}
    \centering
    \includegraphics[width=\linewidth]{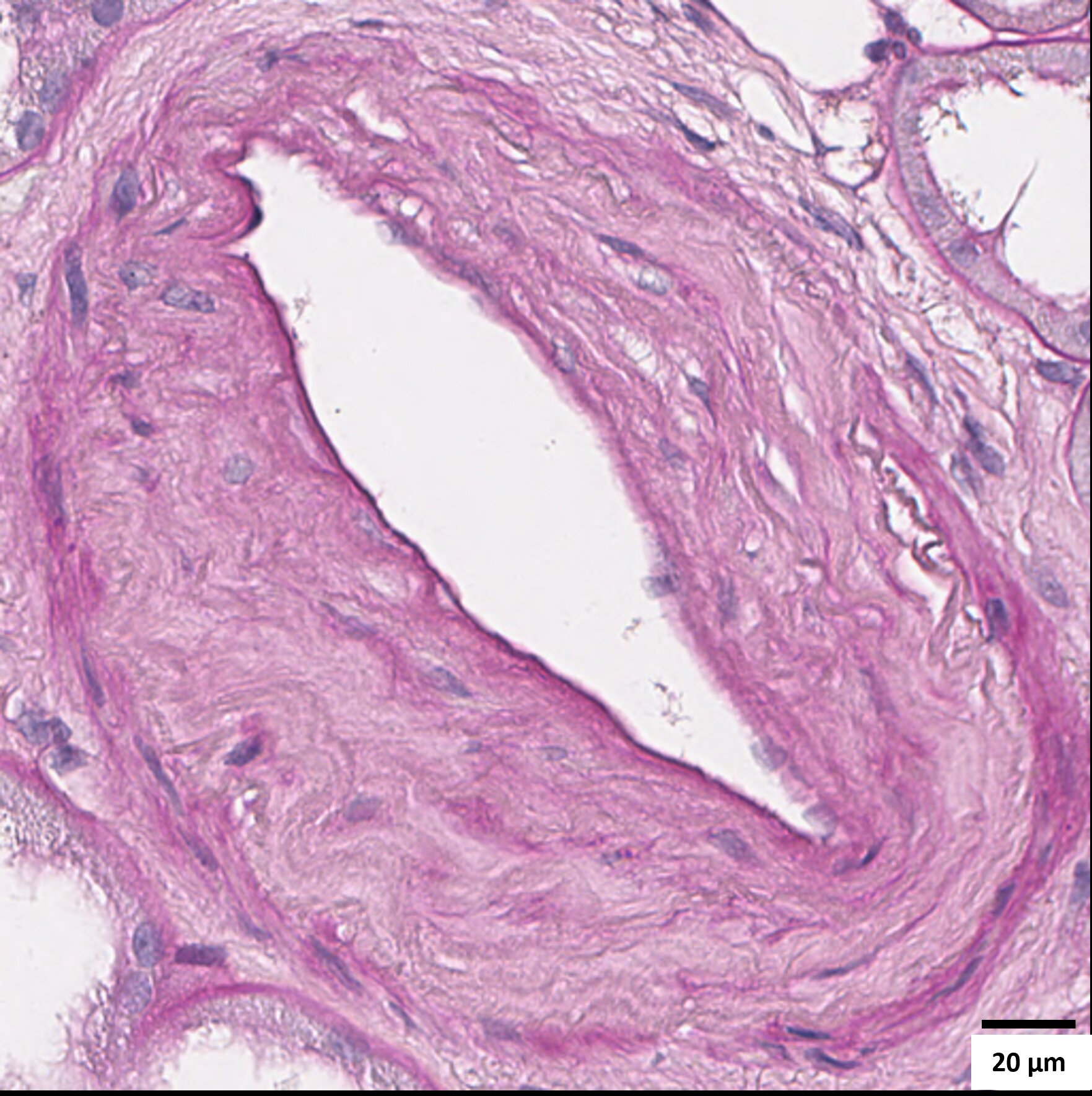}
    \caption{Severe stenosis artery}
    \label{fig:severe_stenosis}
\end{subfigure}
\caption{Representative histopathology examples of mild, moderate, and severe arteriolar stenosis.}
\label{fig:Arteriolar_Stenois_comparison}
\end{figure}

\textit{\textbf{Arterial stenosis classification:}} This multi-class classification task involves classifying the extent of stenosis in the vessels (arteries/arterioles) using established angiographic thresholds\cite{arab2022review, kliewer1993renal, desberg1990renal}. Our dataset comprised 210 arteries drawn from 73 PAS-stained WSIs acquired at 0.25 mpp, with 133 arteries in Class 0 (0–49\%, Figure~\ref{fig:mild_stenosis}), 52 in Class 1 (50–69\%, Figure~\ref{fig:moderate_stenosis}), and 25 in Class 2 (70–100\%, Figure~\ref{fig:severe_stenosis}). Pathologists selected and annotated the most severely affected arteries in each renal allograft biopsy because Banff~\cite{roufosse20182018} and donor‐biopsy scoring systems~\cite{munivenkatappa2008maryland, de2013predictive} are designed to assess chronic vascular injury in arteries with a clearly defined intima and media. Assessing arteriolosclerosis in the kidney correlates with interstitial fibrosis, tubular atrophy, urine protein, and baseline eGFR, thus impacting disease severity and prognosis~\cite{zhang2021modified, sugiura2021severity}. Precise classification supports prognostic evaluation and helps personalize antihypertensive or cardiovascular therapies to preserve renal function~\cite{oshiro2022age, shimizu2019association}. 

\bigskip
\textit{\textbf{Regression task}} 

\begin{figure}[H]
\centering
\begin{subfigure}[t]{0.65\linewidth}
    \centering
    \includegraphics[width=\linewidth]{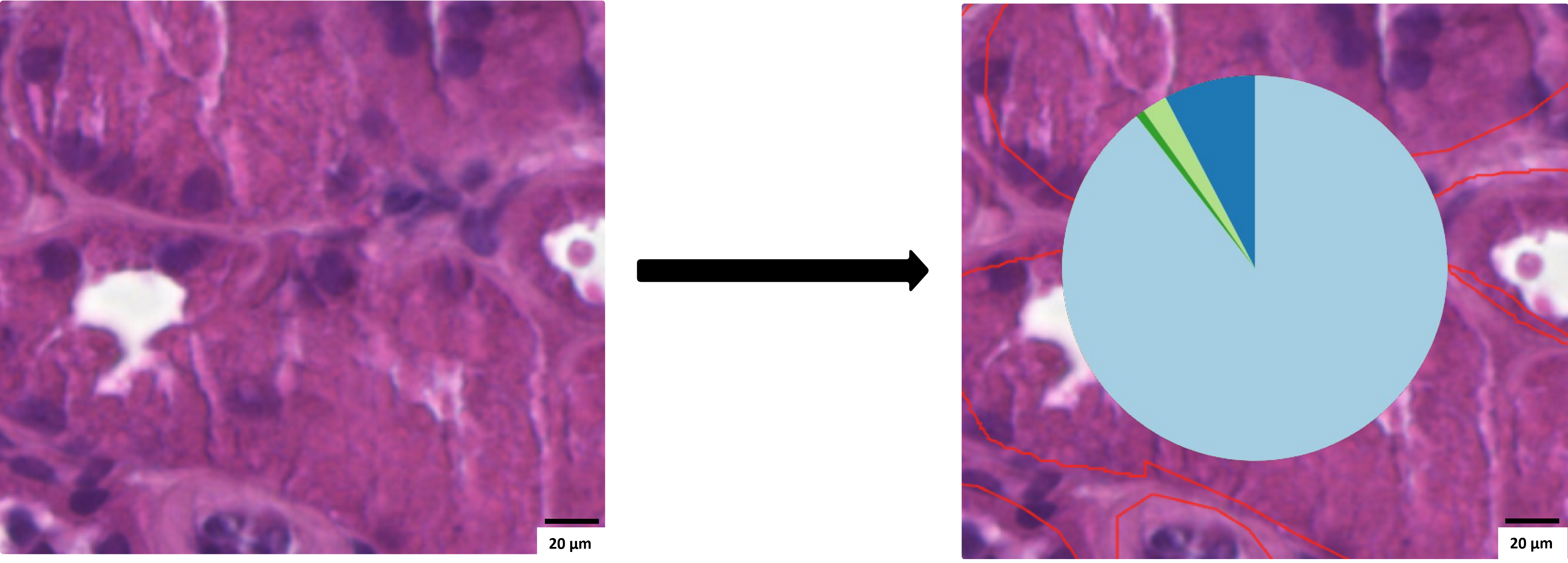}
    \caption{Cell Type Estimation}
    \label{fig:cell_type_estimation}
\end{subfigure}
\caption{Sample histology image patch from which the cell type is estimated.}
\label{fig:cell_type_regression}
\end{figure}

\textit{\textbf{Cell type proportion estimation:}} This ridge regression task involves estimating the relative proportions of 16 main renal cell types (ascending thin limb [ATL], connecting tubule [CNT], distal convoluted tubule [DCT], descending thin limb [DTL], endothelial cell [EC], fibroblast [FIB], Intercalated cell [IC], immune cell [IMM], neural cell [NEU], principal cell [PC], parietal epithelial cell [PEC], podocyte [POD], proximal tubule [PT], papillary tip epithelial cell [PapE], thick ascending limb [TAL], and vascular smooth muscle cell or pericyte [VSM/P]) directly from histology image patches (Figure~\ref{fig:cell_type_regression}). We used the same dataset as the tubule classification task, from which we extracted histology image tiles of size 224 $\times$ 224 px at their native resolution, spatially aligned with Visium FFPE ST data for 17,478 spots. A detailed description of all the cell-type definitions is provided in our previous work~\cite{border2025fusion}. Accurate prediction of cellular proportions is crucial, as the spatial organization and abundance of specific cell populations such as podocytes, endothelial cells, tubular epithelial cells, fibroblasts, and infiltrating immune cells play a crucial role in defining the niches of renal injury~\cite{lake2023atlas, ferreira2021integration}. Changes in these proportions are associated with key pathological processes such as fibrosis and inflammation, which are predictive of disease progression and treatment response~\cite{abedini2024single, ferreira2021integration,polonsky2024spatial}. This regression task thus evaluates the HFMs capacity to encode high-dimensional, biologically meaningful features that reflect both the cellular heterogeneity and functional state of the kidney tissue, bridging molecular and morphologic pathology.

\begin{figure}[H]
\centering
\begin{subfigure}[t]{0.19\linewidth}
    \centering
    \includegraphics[width=\linewidth]{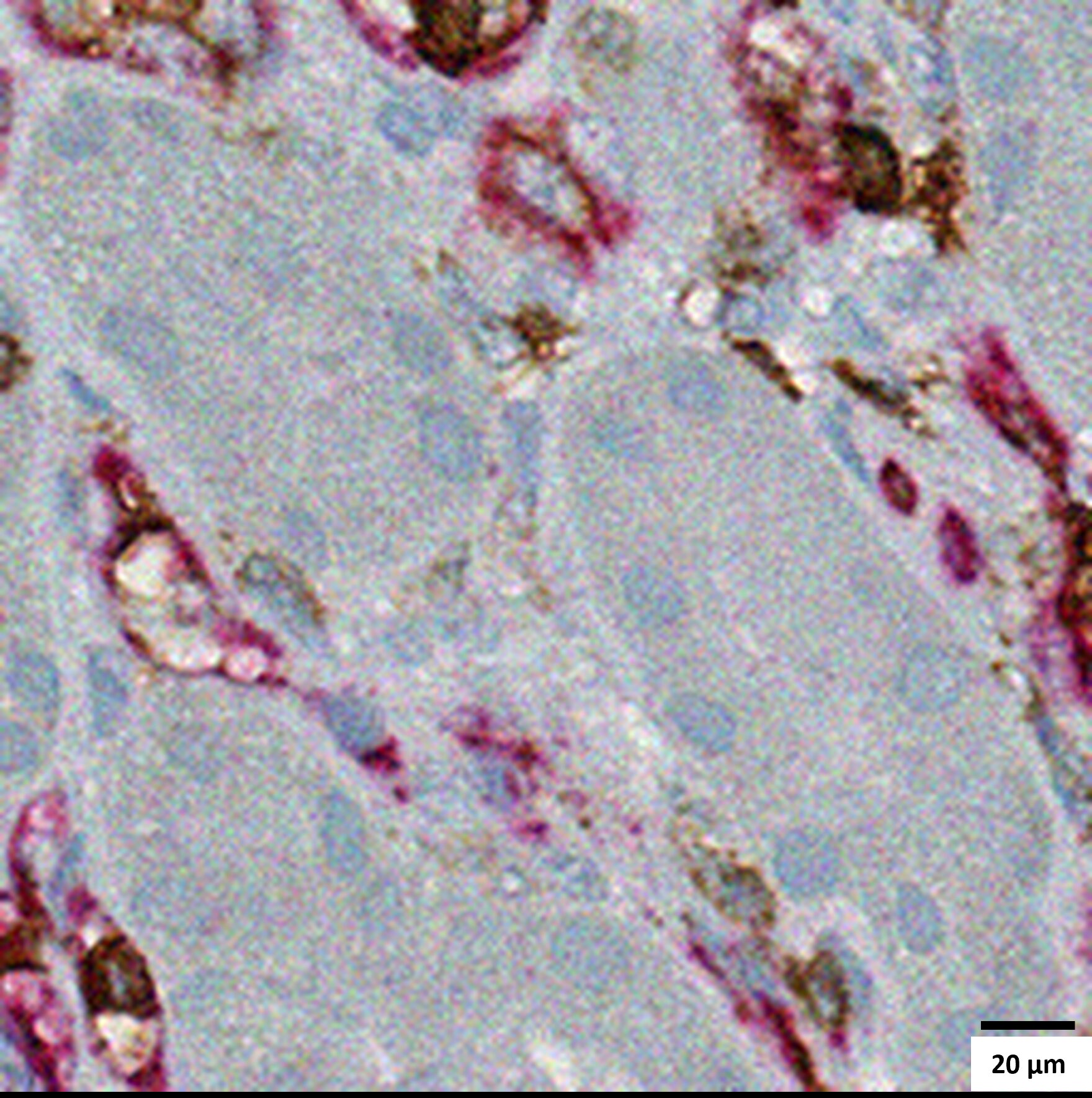}
    \caption{Original patch}
    \label{fig:Original image}
\end{subfigure}
\begin{subfigure}[t]{0.19\linewidth}
    \centering
    \includegraphics[width=\linewidth]{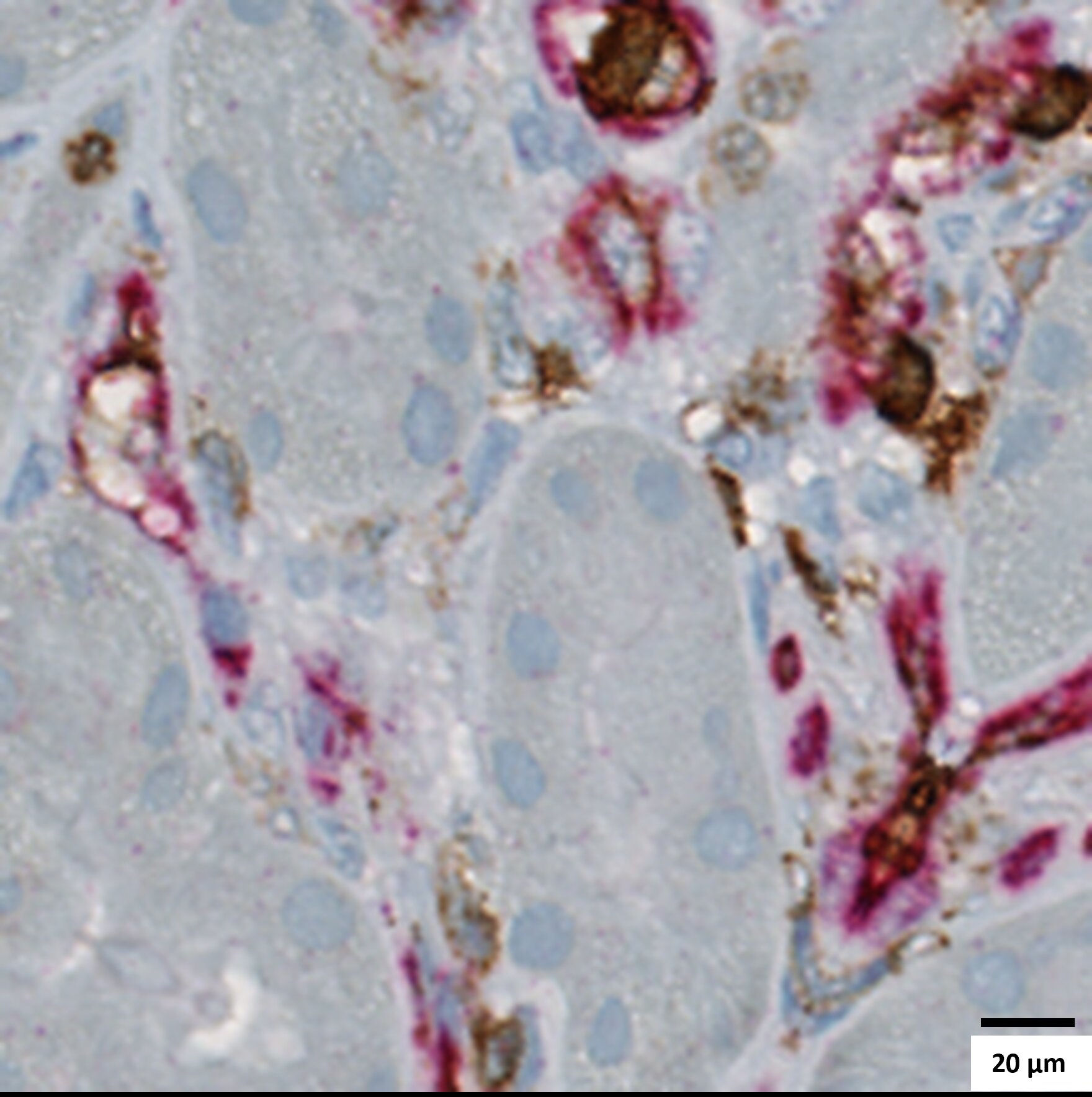}
    \caption{Geometric transformation}
    \label{fig:Geometric augmentation }
\end{subfigure}
\begin{subfigure}[t]{0.19\linewidth}
    \centering
    \includegraphics[width=\linewidth]{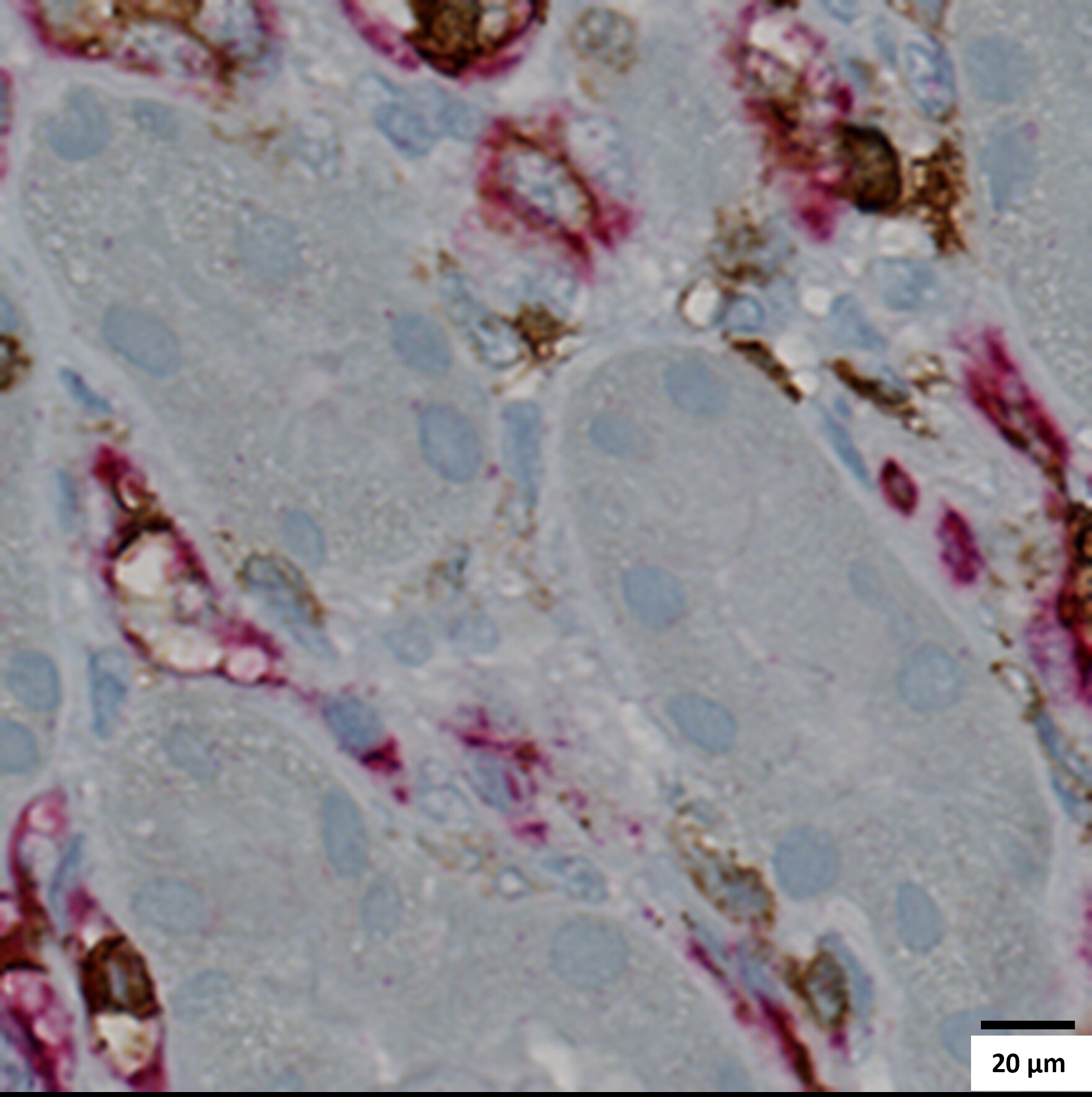}
    \caption{Color adjustment}
    \label{fig:Color augmentation}
\end{subfigure}
\begin{subfigure}[t]{0.19\linewidth}
    \centering
    \includegraphics[width=\linewidth]{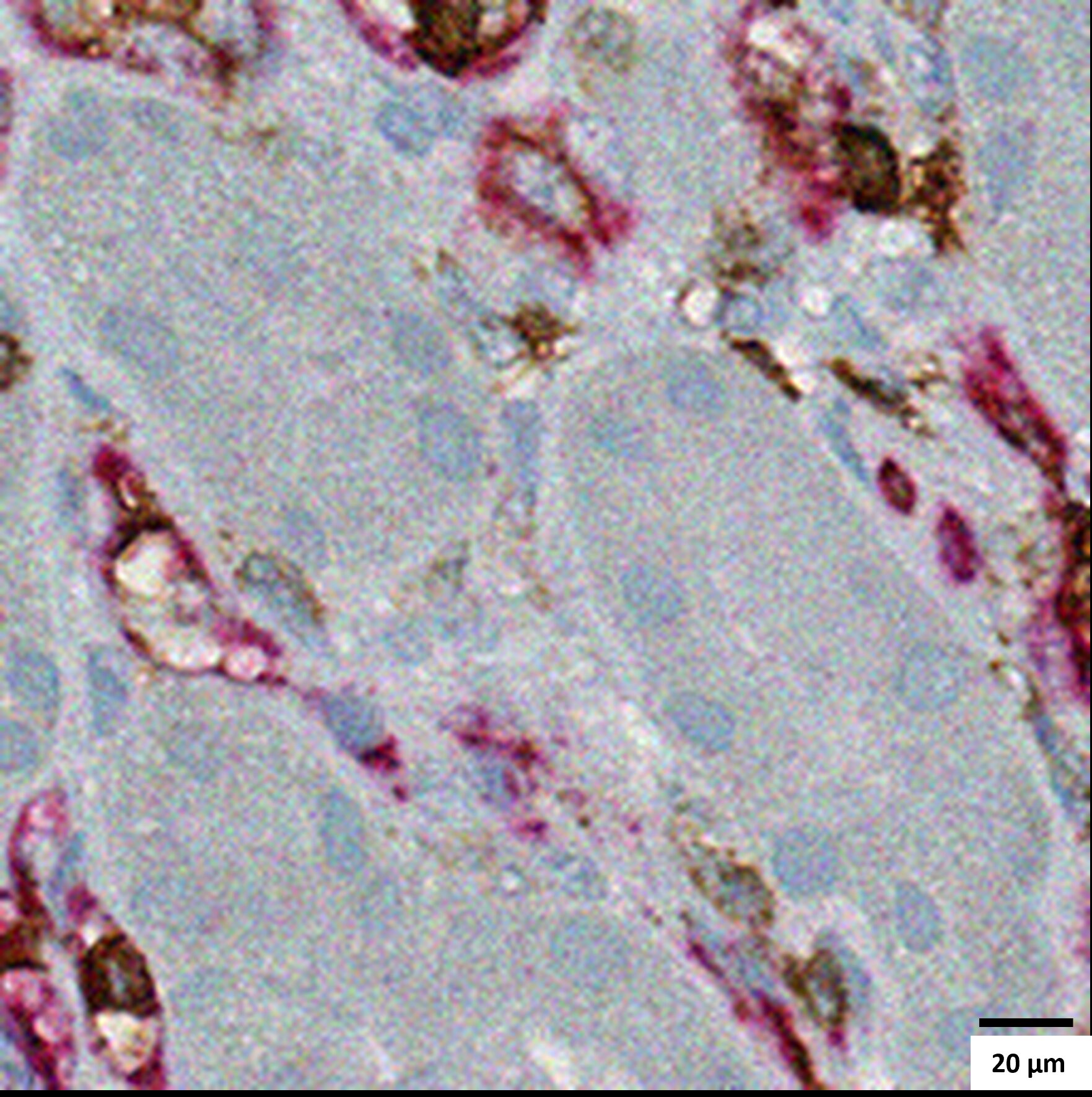}
    \caption{Noise/JPEG artifacts}
    \label{fig:Noise augmentation}
\end{subfigure}
\begin{subfigure}[t]{0.19\linewidth}
    \centering
    \includegraphics[width=\linewidth]{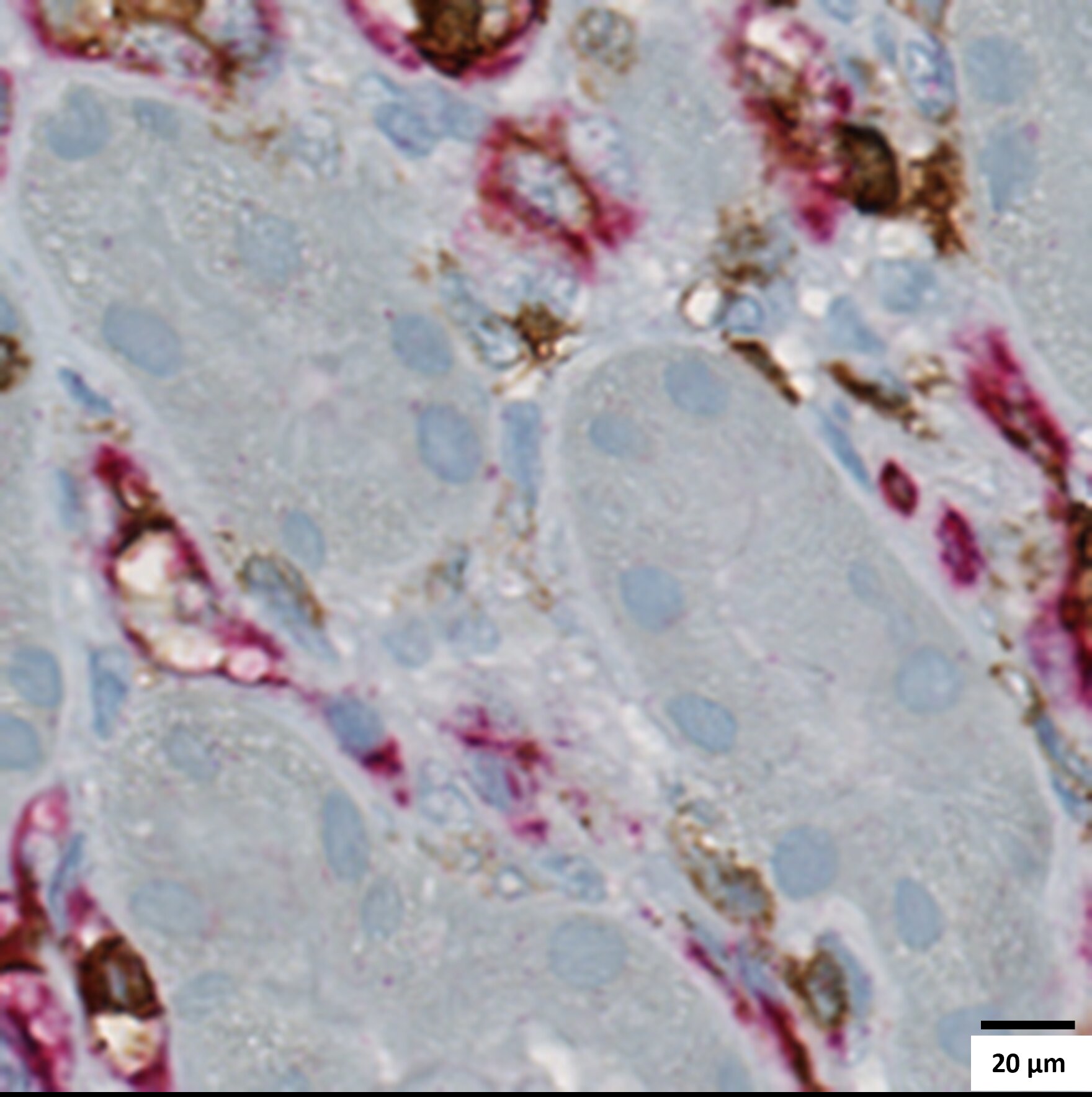}
    \caption{Elastic deformation}
    \label{fig:Elastic deformation augmentation}
\end{subfigure}
\caption{Representative histopathology examples of the augmentations applied to a single microvascular inflammation tissue patch for the copy detection task.}
\label{fig:copy_detection}
\end{figure}

 \textit{\textbf{Copy detection problem}}: This copy detection task involves assessing HFM embeddings robustness by calculating the cosine similarity between an original image’s embedding and the embeddings of its augmented variants (Figure~\ref{fig:copy_detection}), following the approach of Pizzi \textit{et al.}~\cite{pizzi2022self, ciga2022self}. We extracted 2,405 tiles of size 224 $\times$ 224 px at 0.5 mpp resolution from four microvascular inflammation section IHC (CD34 and CD45) WSIs obtained from Johns Hopkins University (JHU), on which we applied a set of augmentations to mirror variations encountered in practice. Specifically, we used geometric adjustments (rotations, shifts, scaling), color and illumination changes (brightness, contrast, gamma), scanner-related distortions (Gaussian noise), and mild elastic deformations mimicking tissue handling. More details on the Python functions used to simulate these augmentations are available in \textit{Supplementary Section~\ref{sup:sec:aug_copy_detection_task}}.

Together, these evaluation tasks establish a comprehensive and clinically meaningful benchmark for assessing HFMs in renal histopathology at the tile-level. The tile-level classification tasks evaluate each model’s ability to detect morphological changes, such as global sclerosis, GBM spikes in glomeruli, abnormal tubules, and subtle inflammatory activity, distinguishing inflamed from non-inflamed regions, as well as fine-grained vascular changes through multi-class arteriolar stenosis classification. The cell type estimation task further tests the HFM's capacity to infer complex cell-type composition by predicting the proportions of 16 distinct cell types directly from histologic features. Finally, the copy detection experiment involves evaluating the robustness of embeddings subjected to geometric, color, noise, and elastic augmentations, which mimic realistic slide-to-slide and scanner-to-scanner variability. Collectively, these experiments span a broad pathophysiological spectrum of chronic kidney injury changes, thus providing a rigorous framework for benchmarking HFM representations in kidney pathology.

\subsection{\textit{\textbf{Slide-level tasks}}}
To evaluate the utility of HFM embeddings at the WSI level, we designed four clinically relevant downstream tasks. These include three binary classification tasks, namely predicting treatment response in MN patients, predicting eGFR decline from one year to three years post-renal transplant, and differentiating between reference vs. DN samples. In addition, we have also incorporated a regression task for estimating eGFR one year post-renal transplant to assess the representation quality of HFM embeddings. These tasks are essential for diagnosis, prognosis, and treatment monitoring in patients with kidney disease, and they reflect a broad spectrum of clinical needs, from therapeutic decision-making to long-term renal function tracking.

\bigskip
\textit{\textbf{Classification tasks}} 

\textit{\textbf{Reference vs. diabetic nephropathy}}: 
This binary classification task aims to distinguish DN cases from reference renal tissue using WSIs. The dataset includes 137 PAS-stained WSIs acquired at 0.25 mpp resolution, comprising 49 reference and 88 DN samples sourced from WUSTL. Accurate discrimination between DN and normal kidney morphology is essential for early diagnosis, disease staging, and guiding treatment strategies~\cite{qi2017classification, pourghasem2015histological}. DN is characterized by various histopathologic changes, including GBM thickening, increased mesangial matrix expansion, nodular (Kimmelstiel–Wilson) sclerosis, and varying degrees of tubulointerstitial injury~\cite{najafian2015ajkd}. These structural changes collectively signify injury and loss of renal function~\cite{liu2022early}. This task evaluates the discriminative capacity of HFMs to recognize and embed these disease-defining morphologic signatures, enabling differentiation of DN from reference renal tissue across patient cohorts. 

\textit{\textbf{Response vs. no response prediction to drug treatment in MN:}}  
This binary classification task aims at predicting treatment response after 52 weeks in patients diagnosed with MN. The dataset comprised 85 PAS-stained WSIs acquired at 0.25 mpp resolution obtained from the NIH/NIDDK, including 14 non-responders and 71 responders. Early identification of patients unlikely to respond to treatment is clinically important, as it enables personalization of drug treatment and risk stratification~\cite{rojas2025identification, liu2023prognostic}. This task assesses the capacity of HFMs to extract and leverage histomorphologic features that correlate with treatment outcomes in MN, thereby providing insight into their potential utility for precision prognostication in glomerular diseases.

\textit{\textbf{eGFR decline from one year to three years post-renal transplant:}}  
This binary classification task aims to predict whether a renal transplant recipient will experience a decline in eGFR between one and three years post-renal transplant, a critical indicator of graft function and overall renal health. The dataset comprised 132 donor biopsy PAS-stained WSIs acquired at 0.25 mpp resolution, collected from JHU and UC Davis, of which 74 cases had stable eGFR (i.e., no decline) and 58 cases exhibited eGFR decline. Early identification of patients at risk for eGFR decline enables timely clinical intervention, closer monitoring, and potentially preemptive treatment strategies to preserve graft longevity~\cite{schold2022clinical, irish2021change, hong2022personalized, suzuki2025deep}. 

\bigskip
\textit{\textbf{Regression task}} 

\textit{\textbf{eGFR prediction one year post-renal transplant:}}  
This regression task involves assessing the ability of HFM embeddings to predict renal function, quantified by the eGFR one year post-renal transplant. We use a total of 204 PAS-stained WSIs acquired at 0.25 mpp resolution from renal transplant donor biopsies collected across two institutions, JHU and UC Davis, providing a diverse dataset for evaluating model generalization. By linking tissue morphology to functional outcomes, this task provides a critical bridge between image-derived features and measurable clinical biomarkers of kidney health. This task evaluates whether histological representations learned by HFMs encode morphological features that correlate with renal functional decline. Clinically, this enables estimation of kidney function directly from biopsy WSIs, potentially reducing reliance on frequent biochemical testing and facilitating better prognosis~\cite{suzuki2025deep}.

Together, these slide-level classification and regression tasks encompass key clinical objectives in renal pathology, including the prediction of treatment response in MN, forecasting post-renal transplant kidney function, and distinguishing DN from reference renal tissue. Collectively, they span the spectrum of diagnostic, prognostic, and therapeutic monitoring needs in kidney disease. By linking histologic morphology to clinically meaningful outcomes such as diagnosis and prognosis, these tasks provide a comprehensive and biologically relevant testbed for evaluating the robustness, clinical relevance, and generalizability of HFM representations at the WSI level.

\section{Methodology}
\label{sec:methodology}
\subsection{Data Preprocessing} 
All WSIs for slide-level tasks were tiled into non-overlapping $224{\times}224$~px patches at 0.5 mpp resolution. Tile-level filtering was performed to remove excessive background and artifacts such as pen markings and black borders commonly encountered in WSIs. Using the PyVIPS library~\cite{Cupitt_The_libvips_image_2025}, each tile was loaded and decomposed into its red, green, and blue channels. Three complementary quality-control filters were applied:
(1)~the background fraction, computed via Otsu’s~\cite{otsu1975threshold} method on grayscale histograms, which excluded tiles with more than 90\% background pixels;
(2)~the blackish fraction, which quantified low-intensity RGB regions and discarded tiles with more than 5\% dark area; and
(3)~the pen-mark fraction, which identified pen annotations based on empirically defined RGB thresholds for red, green, and blue inks, removing tiles where any color channel exceeded 5\% coverage.

Following quality-controlled tiling, feature representations were extracted for each tile without normalization from each HFM, following a unified and reproducible inference pipeline. Each retained tissue tile was resized to the model’s native input dimension (typically $224\times224$~px) and processed using that model’s prescribed preprocessing pipeline. We implemented feature extraction using PyTorch, leveraging pre-trained encoders from publicly available repositories (e.g., UNI, UNI-2h, Virchow, Virchow2, Hibou-B/L, SP22M, SP85M, H-optimus-0/1, and Prov-Gigapath).

\subsection{Generalized Predictive Performance Evaluation Strategy}

\begin{figure}[!tbp]
\centering
    \centering
    \includegraphics[width=\linewidth]{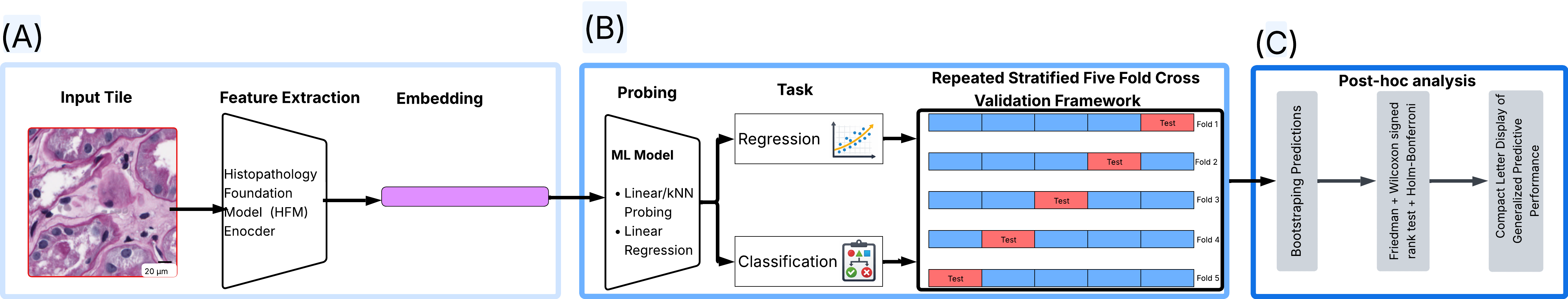}
\caption{
\textbf{Overview of the tile-level evaluation workflow for benchmarking histopathology foundation models (HFMs).} 
\textbf{(A)} An input histology tile is passed through one of 11 HFM encoders 
(H-optimus-0, H-optimus-1, Hibou-B, Hibou-L, Prov-Gigapath, SP22M, SP85M, UNI, UNI2-h, Virchow, Virchow2) 
to obtain a fixed-dimensional feature embedding. 
\textbf{(B)} The extracted embedding is used to train lightweight machine learning models either linear or $k$NN probing for classification tasks, or ridge regression for continuous prediction tasks. 
\textbf{(C)} Model performance is evaluated using a repeated group five-fold cross-validation framework, followed by bootstrapping of predictions, and a statistical framework integrating Friedman test, Wilcoxon signed-rank tests, and Compact Letter Display (CLD)~\cite{gramm2007algorithms} analysis to quantify variance and compute generalized predictive performance across all tile-level tasks.
}
    \label{fig:tile_level_benchmark}
\label{fig:Methodology_benchmarking}
\end{figure}

We summarize the statistical framework used to ensure fair, robust, and reproducible comparison across HFMs for patch-based tasks. Figure~\ref{fig:tile_level_benchmark} shows the strategy for tile-level evaluations, conducted using a repeated stratified group K-fold scheme to account for patient-level grouping and prevent tile-level data leakage. Figure~\ref{fig:slide_level_benchmark} shows the strategy for slide-level evaluations, conducted using a repeated nested stratified K-fold scheme. Five folds and three random seeds were used for both tile and slide-level evaluation, yielding 15 independent train-test evaluations, each seed producing one test prediction per tile. For each seed, model training and validation were performed across all folds while preserving class balance and ensuring that tiles originating from the same patient never appeared in both training and test partitions. Across the three seeds, this yielded up to three test-time predictions per tile, which were aggregated to derive stable performance estimates.

To quantify uncertainty, we applied bootstrap resampling (B = 1000 replicates) to the aggregated predictions from each HFM. For evaluation, we report the Matthews correlation coefficient (MCC) for tile and slide-level classification tasks due to its robustness to class imbalance~\cite{chicco2020advantages, chicco2021matthews}, Pearson's correlation coefficient (PCC) for the cell type estimation task~\cite{li2024recide, jaume2024hest}, $R^2$ for the slide-level regression task, and accuracy for the copy-detection task. For all metrics, we report the mean, median, minimum, maximum, and first and third quartiles of the bootstrap distribution. This unified statistical protocol provides consistent and rigorous uncertainty bounds for all evaluation tasks that follow. 

For post-hoc comparison of HFMs, we used the paired bootstrap observations across HFMs and performed non-parametric repeated measures testing. Specifically, we first used the Friedman test~\cite{friedman1937use} to evaluate whether performance differences among models were statistically significant. When the Friedman test was significant, we conducted pairwise comparisons using the Wilcoxon signed-rank test~\cite{woolson2007wilcoxon} with Holm-Bonferroni correction~\cite{holm1979simple} to control for the family-wise error rate across multiple comparisons. Adjusted \textit{p}-value matrices were used to enable clear interpretation of statistically meaningful differences among HFMs. Finally, compact letter display (CLD)~\cite{gramm2007algorithms} was used for visualization by overlaying the bootstrap metric distributions using violin plots. This post-hoc procedure, combined with bootstrap uncertainty quantification, provides a robust and reproducible statistical foundation for comparing HFMs across all tasks.

\subsubsection*{Tile‐level classification} 
For tile-level classification tasks, we follow previous works from the SSL community~\cite{chen2024towards, balestriero2023cookbook} that use logistic regression (linear probing) and \textit{k}NN~\cite{bentley1975multidimensional} probing to evaluate the discriminative capability and transfer performance of HFMs embeddings on various downstream tasks. Repeated stratified group K-fold cross-validation is employed, with five outer folds repeated across three independent random seeds for robust performance estimation in both linear and kNN probing (see Figure~\ref{fig:tile_level_benchmark}). Following established protocols in SSL and feature evaluation, we trained a logistic regression classifier using scikit-learn implementation on frozen FM embeddings to assess how effectively the learned representations separate clinical outcome classes. Following the study conducted by Chen \textit{et al.}~\cite{chen2024towards}, $\ell_2$ regularization strength $\lambda$ was fixed to $\frac{100}{M \times C}$,
where $M$ denotes the embedding dimension and $C$ the number of output classes, with the L-BFGS solver~\cite{zhu1997algorithm} (as implemented in scikit-learn~\cite{scikit-learn}) with a maximum of 1,000 iterations~\cite{kolesnikov2019revisiting}. This formulation ensures scale-invariant regularization across embeddings of varying dimensionality and task complexity.

In parallel to linear probing, we employed $k$-nearest neighbor ($k$NN) probing using scikit-learn implementation, which is another technique to evaluate the representational quality of HFM embeddings~\cite{caron2021emerging, sariyildiz2023no} in a non-parametric manner. Unlike logistic regression, which learns a linear decision boundary, NN directly measures feature separability by assigning class labels based on the majority vote among the $k$ closest embeddings in the latent space. Following common practice in representation learning~\cite{caron2021emerging, chen2024towards}, we set $k = 20$ and used Euclidean distance as the similarity metric since it is found to be stable for this evaluation setup.

\subsubsection*{Tile‐level regression} 

To assess the representational capacity of HFM embeddings for quantitative prediction tasks, we employed ridge regression~\cite{hoerl1970ridge} probing. This approach evaluates the ability of HFM-derived representations to encode continuous biological or clinical variables, such as cell type proportions or renal function metrics, without further fine-tuning of the encoder. Ridge regression introduces a $\ell_2$ regularization term that stabilizes coefficient estimates and mitigates overfitting in high-dimensional embedding spaces.  Repeated group K-fold cross-validation is employed, with five outer folds repeated across three independent random seeds for robust performance estimation in both linear and kNN probing. Following HEST-Benchmark conventions~\cite{jaume2024hest}, each regression model was trained using the scikit-learn implementation with the regularization parameter $\alpha$ set to $ \frac{100}{M \times C}$, where $M$ is the embedding dimensionality and $C$ the number of regression targets. We employed the singular value decomposition (SVD)-based closed-form ridge regression solver (solver="svd" in scikit-learn), which provides numerically stable solutions for high-dimensional and potentially ill-conditioned design matrices~\cite{scikit-learn}.

\subsubsection*{Copy detection}
In the copy detection~\cite{pizzi2022self} task, we evaluate embedding robustness by testing the invariance of HFM representations under various image augmentations (geometric, color, noise, and deformation). Following established protocols for representation similarity and copy‐detection benchmarks~\cite{caron2021emerging}, we computed top‐1 retrieval accuracy between original and augmented embeddings. Specifically, for each HFM, embeddings were extracted from the original tile and its augmented version, compared using cosine similarity. The top-$k$ metric quantifies the proportion of augmented embeddings whose most similar counterpart (by cosine similarity) corresponds to the same original tile within the top $k$ matches. This evaluation provides a quantitative measure of embedding consistency and perturbation resilience across models.

\subsubsection*{Slide‐level classification} 
\begin{figure}[H]
\centering
    \centering
    \includegraphics[width=\linewidth]{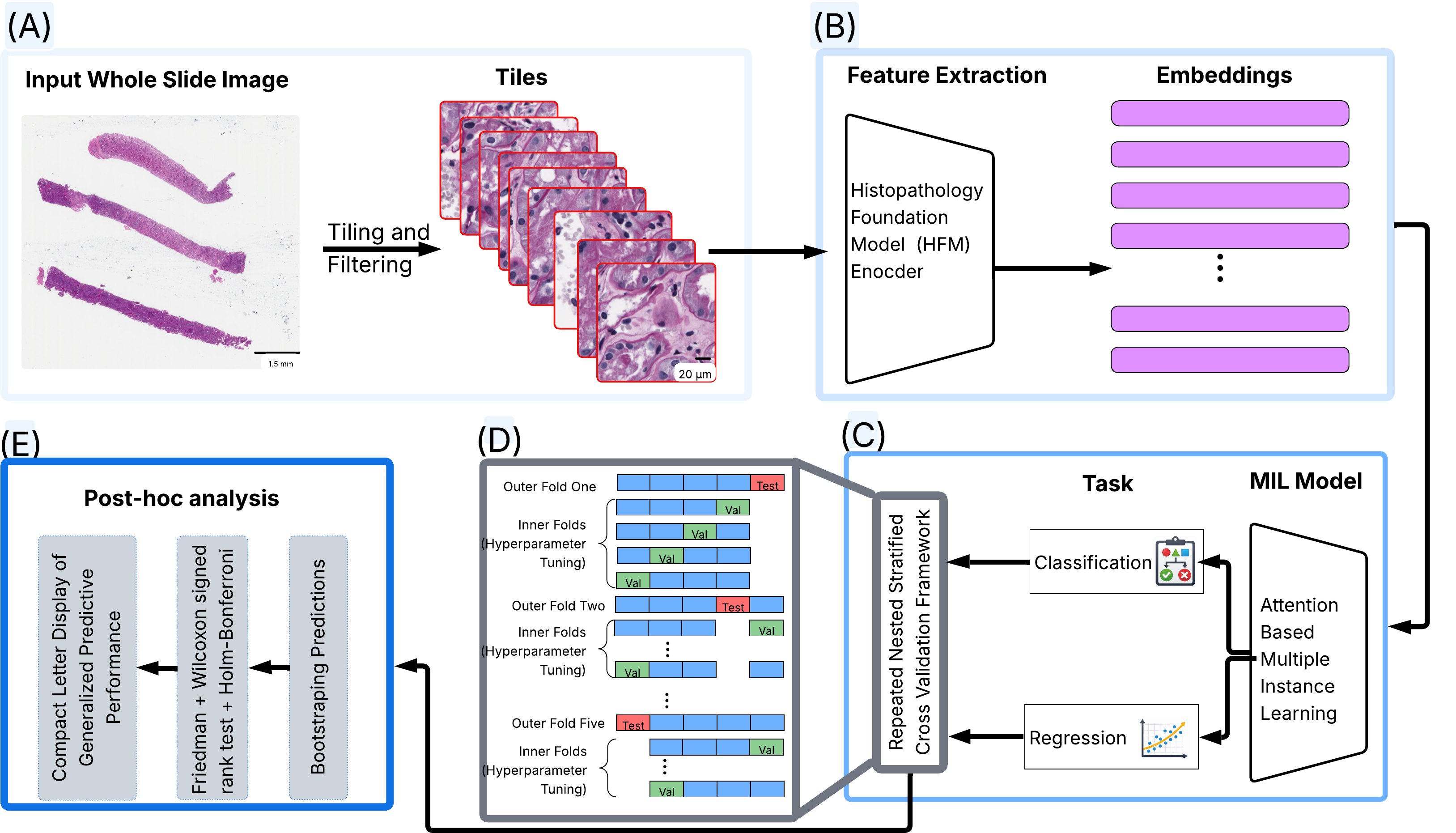}
\caption{
\textbf{Overview of the slide-level evaluation framework for benchmarking histopathology foundation models (HFMs).} 
\textbf{(A)} Whole-slide images (WSIs) are tiled and filtered to generate high-quality image patches. 
\textbf{(B)} Each tile is passed through one of 11 HFM encoders 
(H-optimus-0, H-optimus-1, Hibou-B, Hibou-L, Prov-Gigapath, SP22M, SP85M, UNI, UNI2-h, Virchow, Virchow2) 
to obtain fixed-dimensional feature embeddings. 
\textbf{(C)} Tile embeddings belonging to the same slide are aggregated using an attention-based multiple instance learning (ABMIL) model, which is trained for either regression or classification tasks. 
\textbf{(D)} ABMIL hyperparameter tuning is conducted using a repeated nested stratified five-fold cross-validation framework, where inner folds are used for hyperparameter tuning and outer folds provide unbiased test estimates. 
\textbf{(E)} Predictions from all folds are bootstrapped, followed by a statistical framework integrating the Friedman test, Wilcoxon signed-rank tests, and Compact Letter Display (CLD) analysis to quantify variance and compute generalized predictive performance across slide-level tasks.
}
    \label{fig:slide_level_benchmark}
\label{fig:Methodology_benchmarking}
\end{figure}

Recent benchmarking studies of HFMs~\cite{neidlinger2025benchmarking, campanella2025clinical, ma2025pathbench, kurata2025multiple, meseguer2025benchmarking, mallya2025benchmarking} have primarily relied on multiple instance learning (MIL) frameworks for evaluating slide-level downstream tasks. These works provide limited discussion of standardized hyperparameter choices or the rationale behind their selection, offering little guidance on consistently adopted configurations. Addressing this gap is critical for fair comparison across HFMs and for identifying the true representational capacity of pretrained encoders.
In this study, we used an evaluation framework that integrates hyperparameter optimization using a repeated nested stratified K-fold cross-validation protocol (see Figure~\ref{fig:slide_level_benchmark}). Slide-level classification was performed using an attention-based multiple instance learning (ABMIL) network~\cite{ilse2018attention} applied to tile embeddings extracted from each HFM. We restrict the hyperparameter search space to the most influential ABMIL components, namely the attention layer dimensionality, fully connected layer dimensionality, and learning rate (LR)~\cite{breen2025comprehensive}. Repeated nested stratified K-fold cross-validation is employed, with five outer folds repeated across three independent random seeds for robust performance estimation. Within each outer fold, a four-fold inner CV loop searches over a compact grid spanning learning rate $\{5e-5, 1e-4, 2e-4\}$, attention layer dimension $M \in \{256, 512, 1024\}$, and fully connected layer dimension $L \in \{32, 128, 256\}$. The candidate hyperparameter values for LR, $M$, and $L$ are obtained from the extensive hyperparameter study of Breen \textit{et al.}~\cite{breen2025comprehensive} on a diverse range of HFMs. We selected the three most frequently reported optimal values (modes) across HFMs, yielding a compact externally informed search space. The remaining training hyperparameters, including the  AdamW~\cite{loshchilov2018decoupled} first and second moment decay coefficients (0.95 and 0.99, respectively) and the stability parameter $1e-4$, are fixed to the modal values reported and treated as externally informed defaults. Models are trained for a maximum of 50 epochs with early stopping based on validation MCC (patience = 10). The optimal hyperparameter configuration in each outer fold is subsequently retrained on the combined training and validation bags and evaluated on the held-out test bags. More details on the hyperparameters used are available in \textit{Supplementary Table}~\ref{sup:tab:mil_hyperparams}. By focusing the hyperparameter search on $M$, $L$, and LR, we target the parameters with the strongest impact on ABMIL generalization while maintaining computational feasibility across all HFMs. These hyperparameters directly affect the model’s complexity (via $M$ and $L$) and optimization dynamics (via LR). In contrast, adding broader architectural factors such as number of dropout ratios, optimizer moment decays, or alternative weight-decay settings leads to a combinatorial increase of the search space and substantially increases computational cost. Therefore, we limited the hyperparameter search space to learning rate, attention layer dimensionality, and fully connected layer dimensionality to ensure a principled yet tractable model selection procedure

across all HFMs.
\subsubsection*{Slide-level regression} 
We formulated a slide-level regression task using the same exact set of hyperparameter values as the slide-level classification experiment (see Figure~\ref{fig:slide_level_benchmark}). To adapt the network for regression, the final classification layer is replaced with a single linear output unit that directly predicts the continuous target variable with mean squared error (MSE) as the loss function. Repeated nested K-fold cross-validation is employed, with five outer
folds repeated across three independent random seeds for robust performance estimation.

\section{Results}
\begin{figure}[!htbp]
\centering

\begin{subfigure}[t]{0.33\linewidth}
    \centering
    \includegraphics[width=\linewidth]{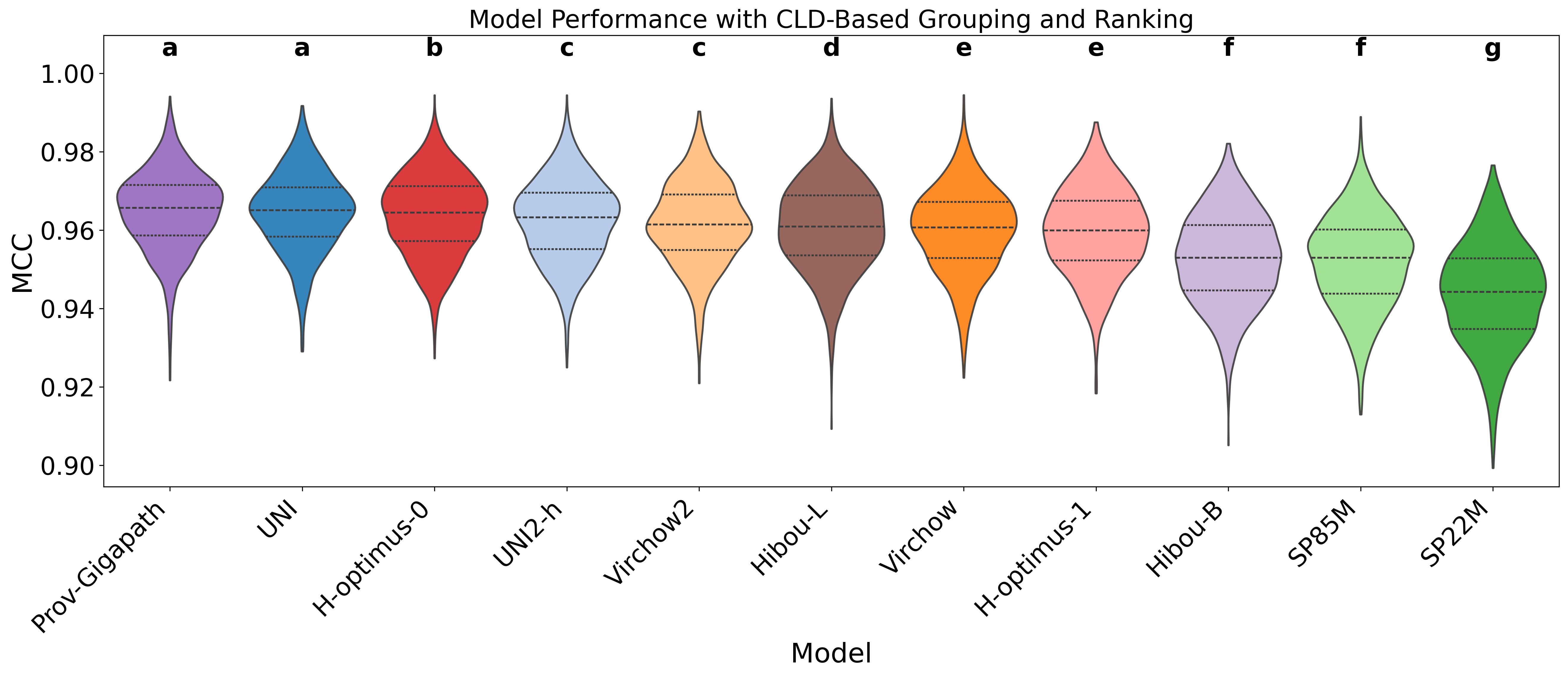}
    \caption{}
    \label{fig:lr_sclerotic}
\end{subfigure}
\begin{subfigure}[t]{0.33\linewidth}
    \centering
    \includegraphics[width=\linewidth]{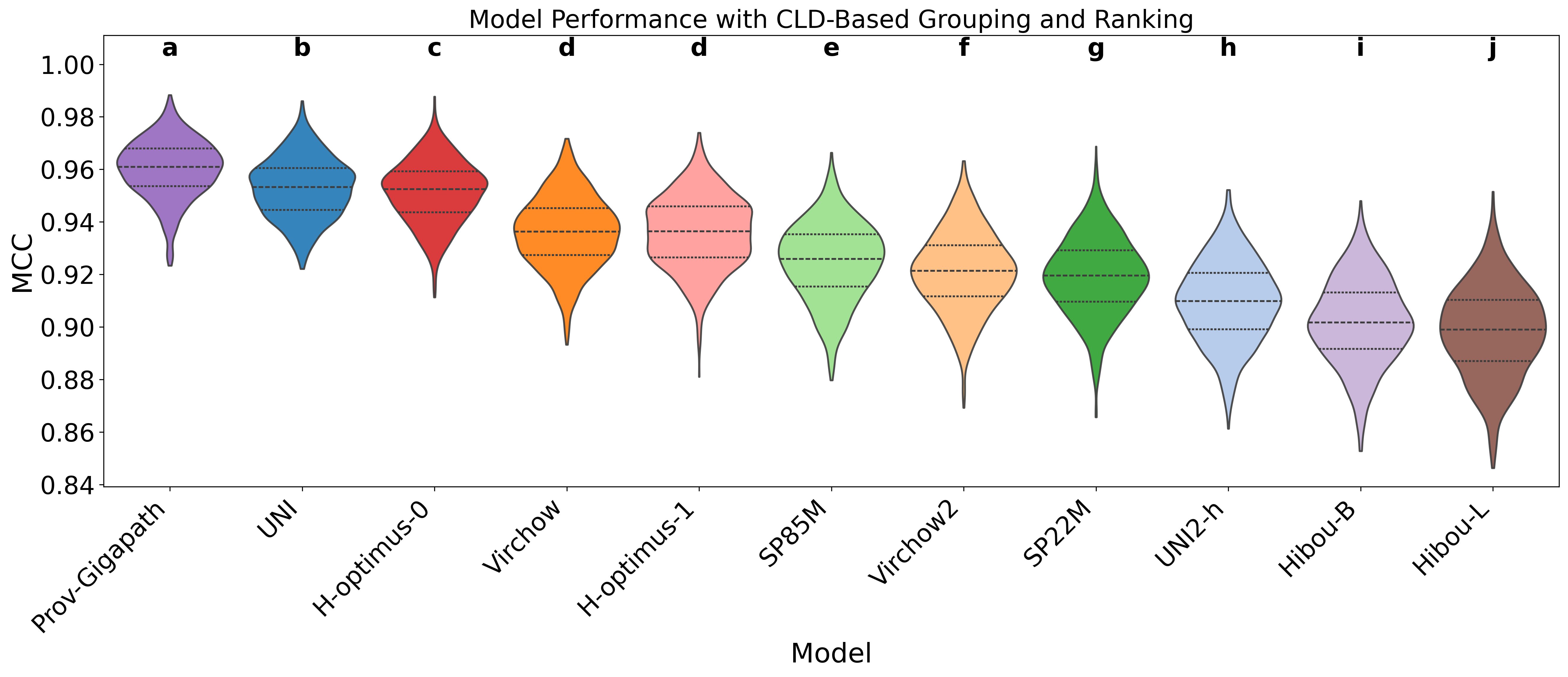}
    \caption{}
    \label{fig:kNN_sclerotic}
\end{subfigure}
\begin{subfigure}[t]{0.33\linewidth}
    \centering
    \includegraphics[width=\linewidth]{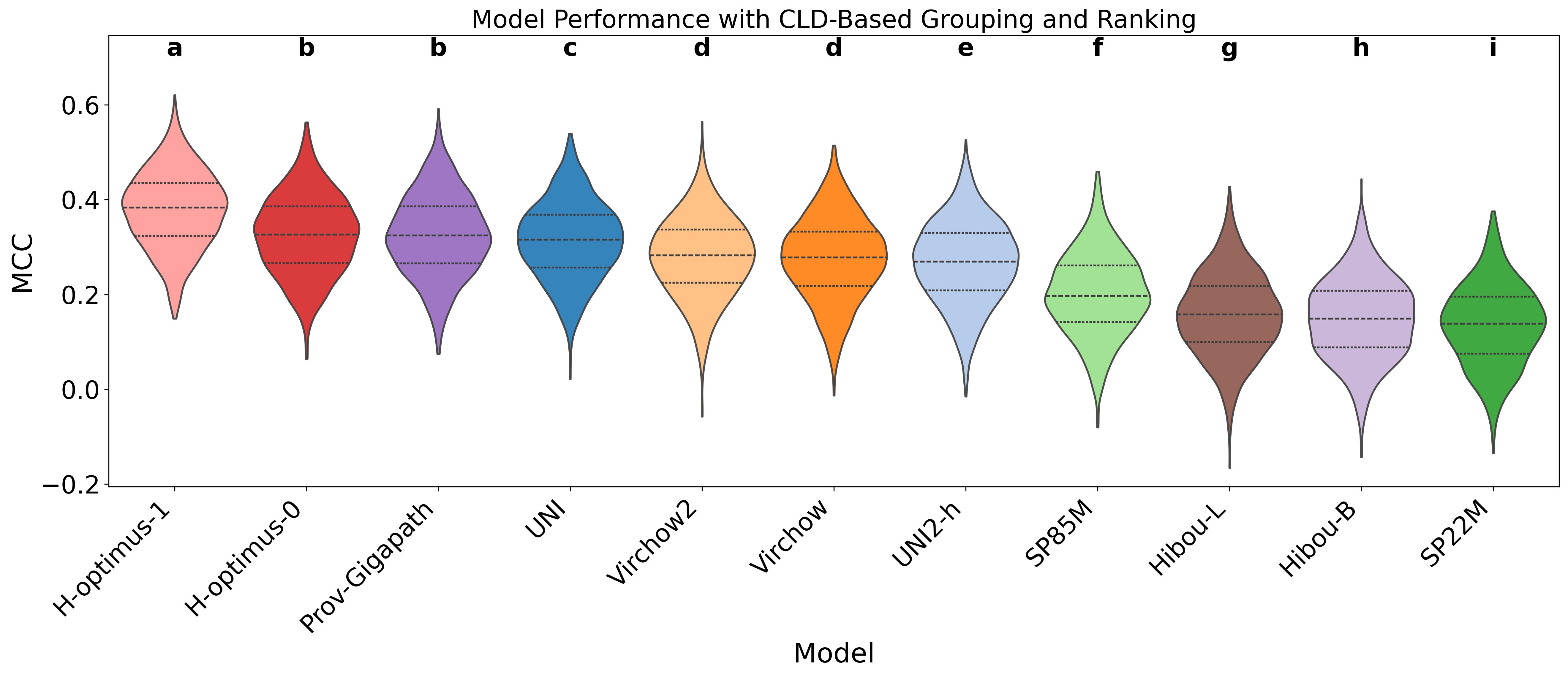}
    \caption{}
    \label{fig:lr_spike}
\end{subfigure}

\par\medskip

\begin{subfigure}[t]{0.33\linewidth}
    \centering
    \includegraphics[width=\linewidth]{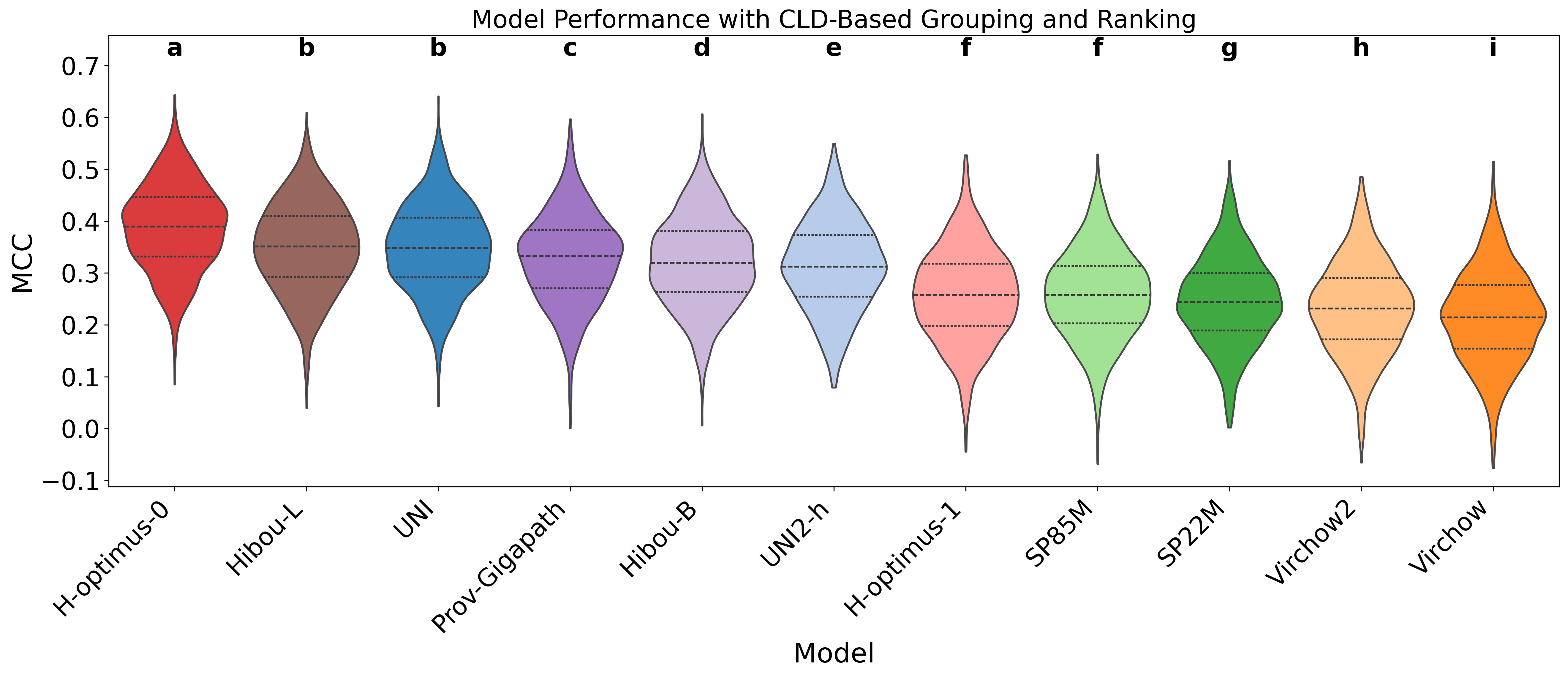}
    \caption{}
    \label{fig:kNN_spike}
\end{subfigure}
\begin{subfigure}[t]{0.33\linewidth}
    \centering
    \includegraphics[width=\linewidth]{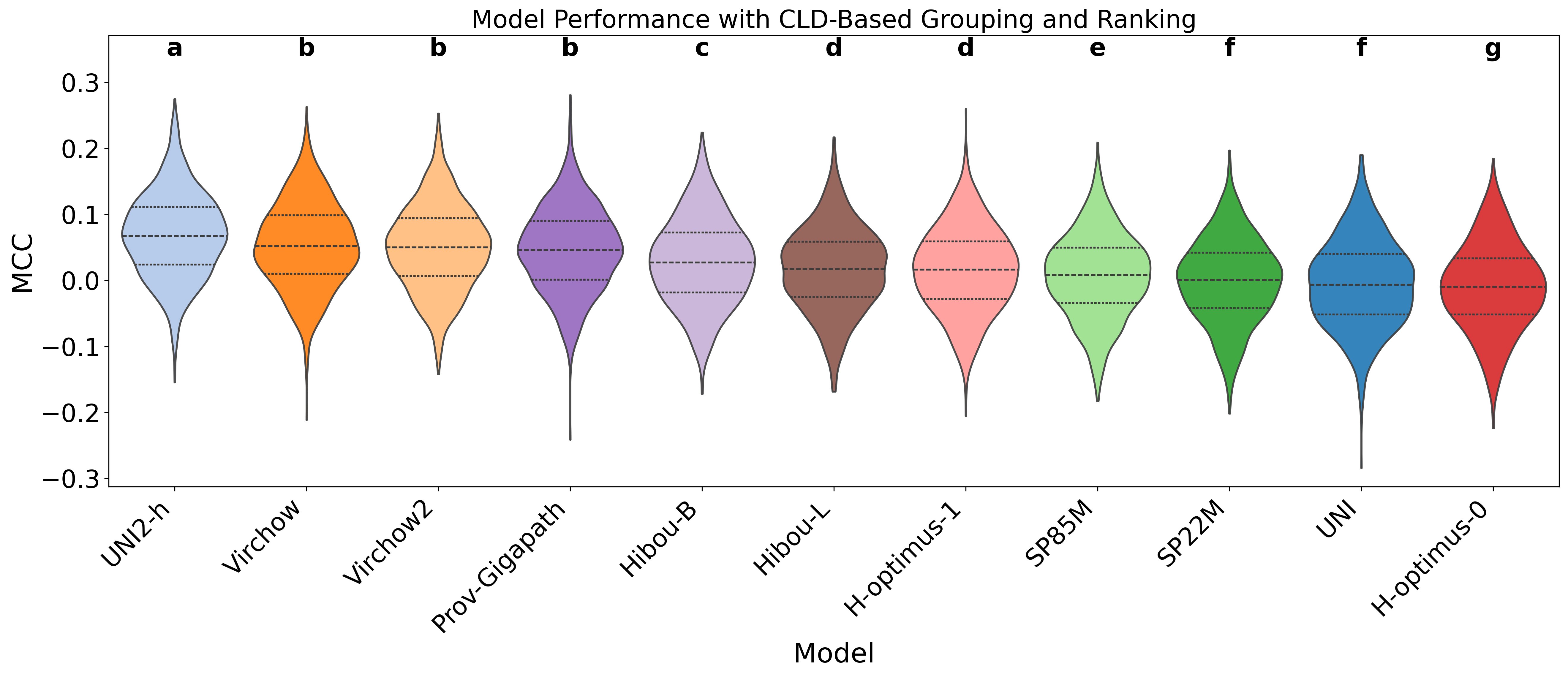}
    \caption{}
    \label{fig:lr_tubule}
\end{subfigure}
\begin{subfigure}[t]{0.33\linewidth}
    \centering
    \includegraphics[width=\linewidth]{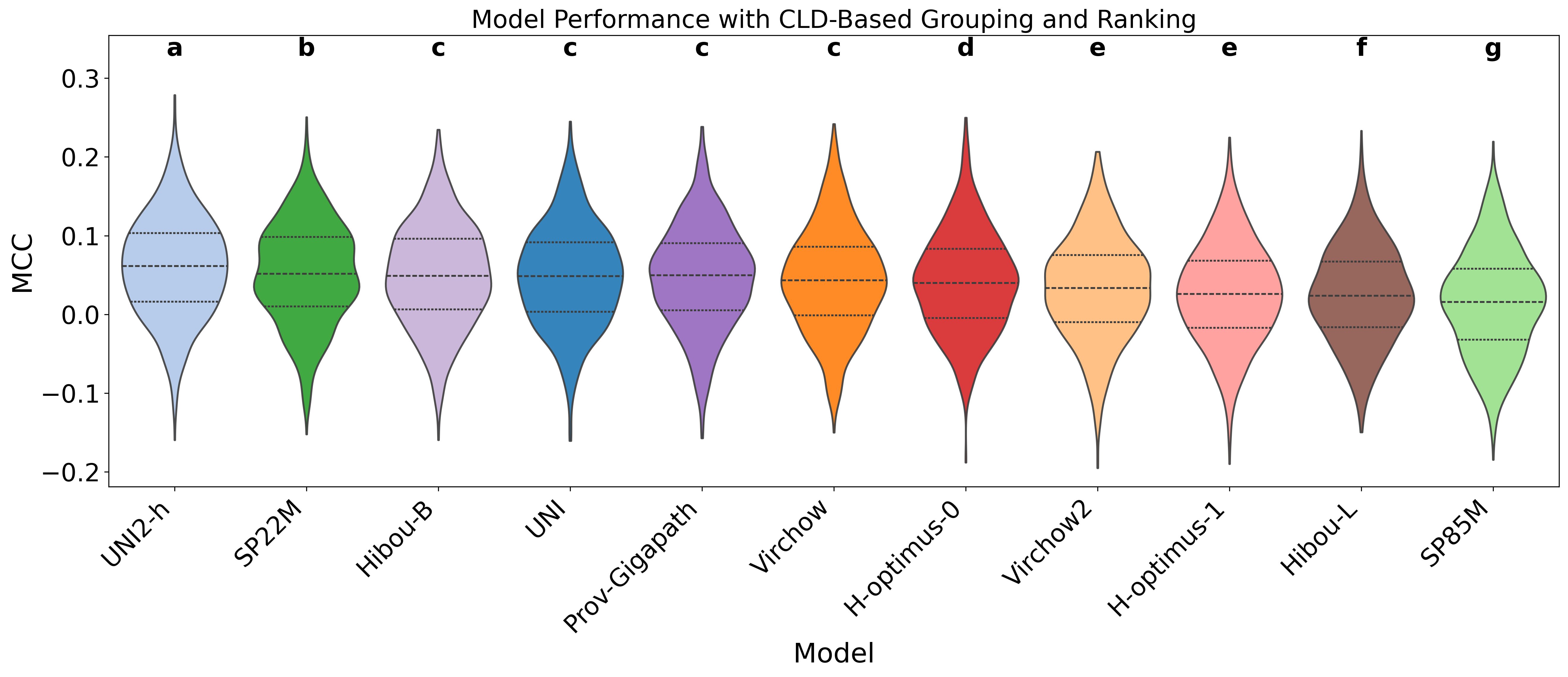}
    \caption{}
    \label{fig:kNN_tubule}
\end{subfigure}

\par\medskip
\begin{subfigure}[t]{0.33\linewidth}
    \centering
    \includegraphics[width=\linewidth]{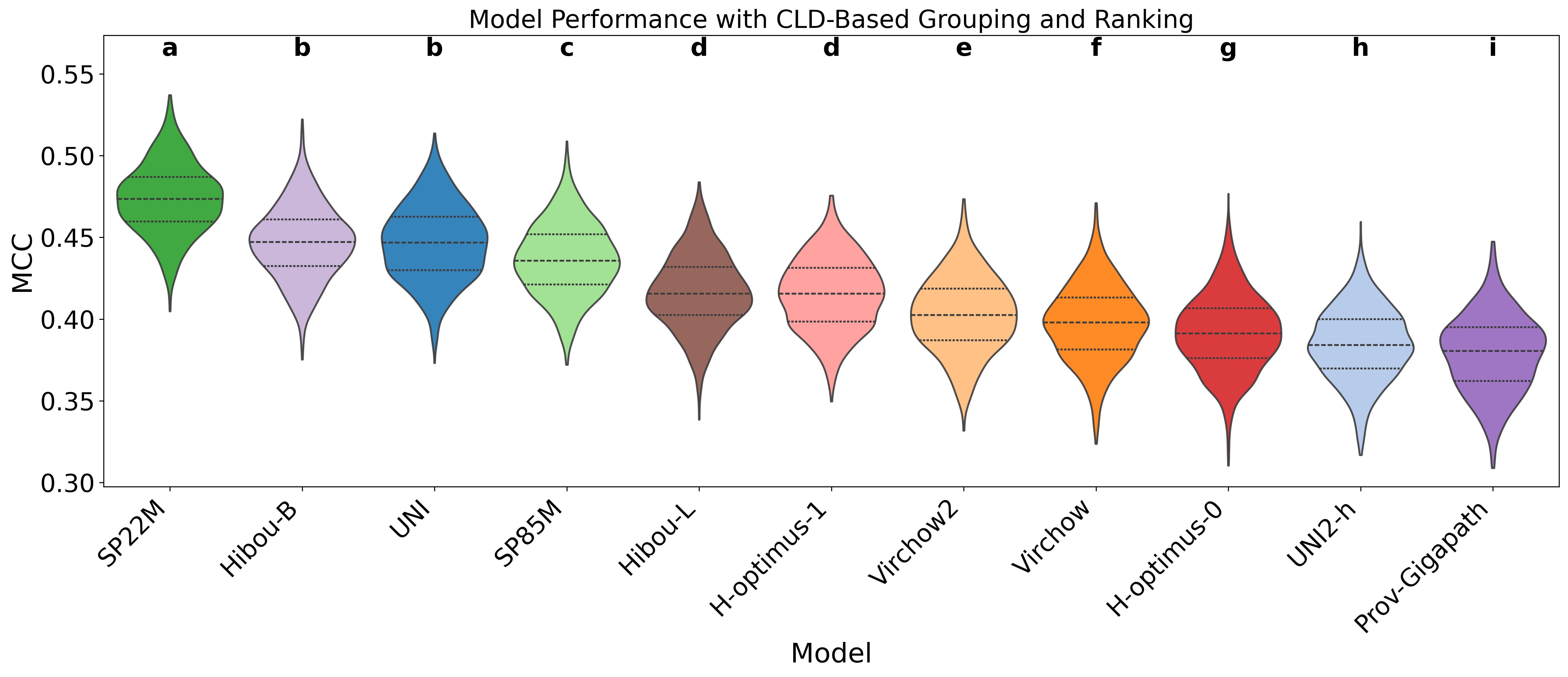}
    \caption{}
    \label{fig:lr_inflammation}
\end{subfigure}
\begin{subfigure}[t]{0.33\linewidth}
    \centering
    \includegraphics[width=\linewidth]{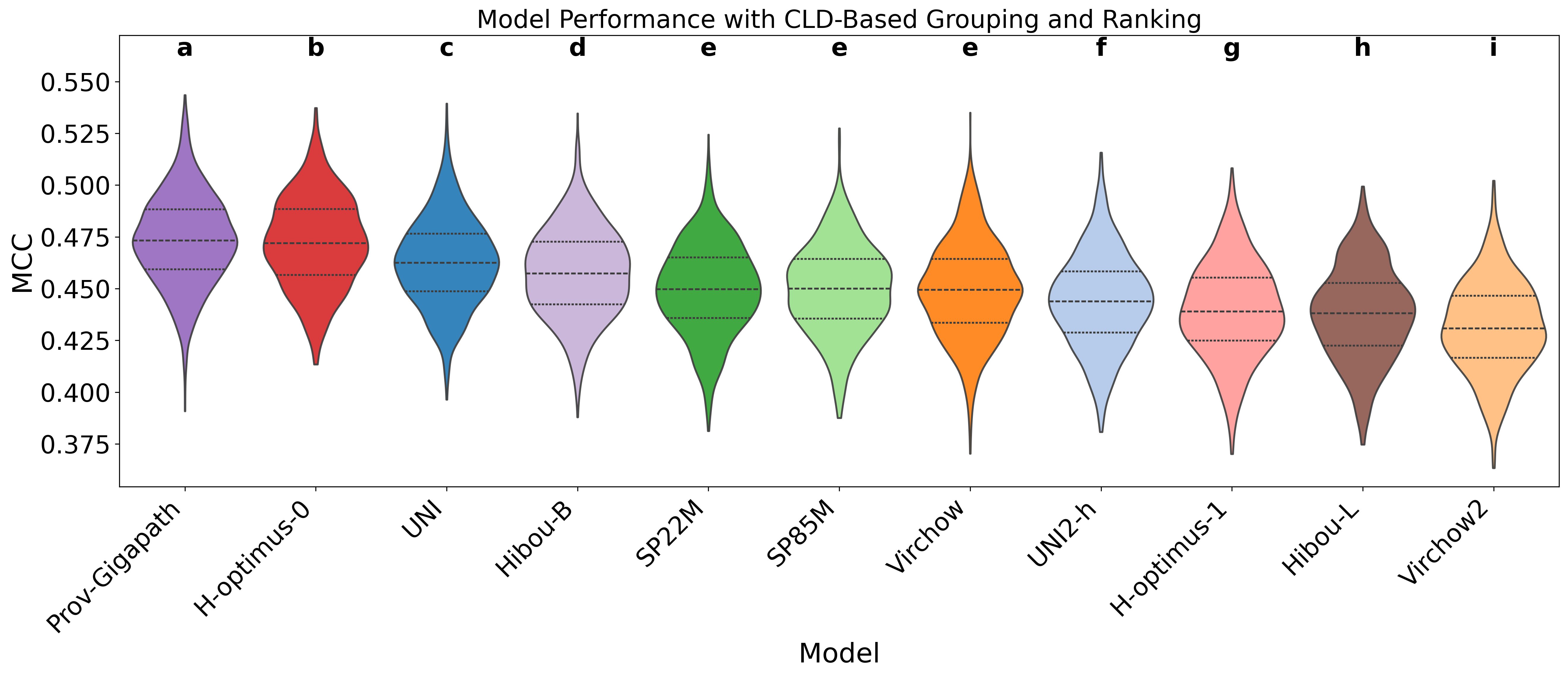}
    \caption{}
    \label{fig:kNN_inflammation}
\end{subfigure}
\begin{subfigure}[t]{0.33\linewidth}
    \centering
    \includegraphics[width=\linewidth]{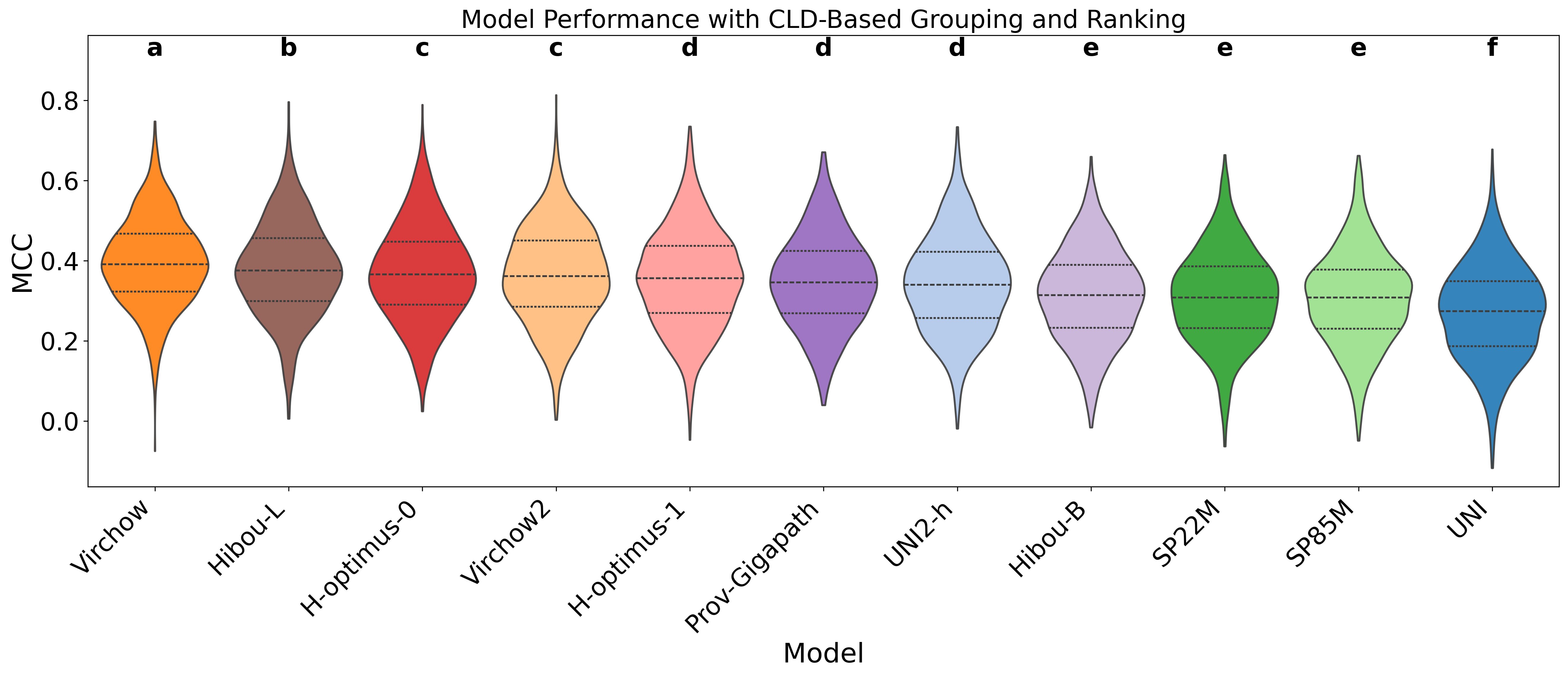}
    \caption{}
    \label{fig:lr_artery}
\end{subfigure}

\par\medskip
\begin{subfigure}[t]{0.33\linewidth}
    \centering
    \includegraphics[width=\linewidth]{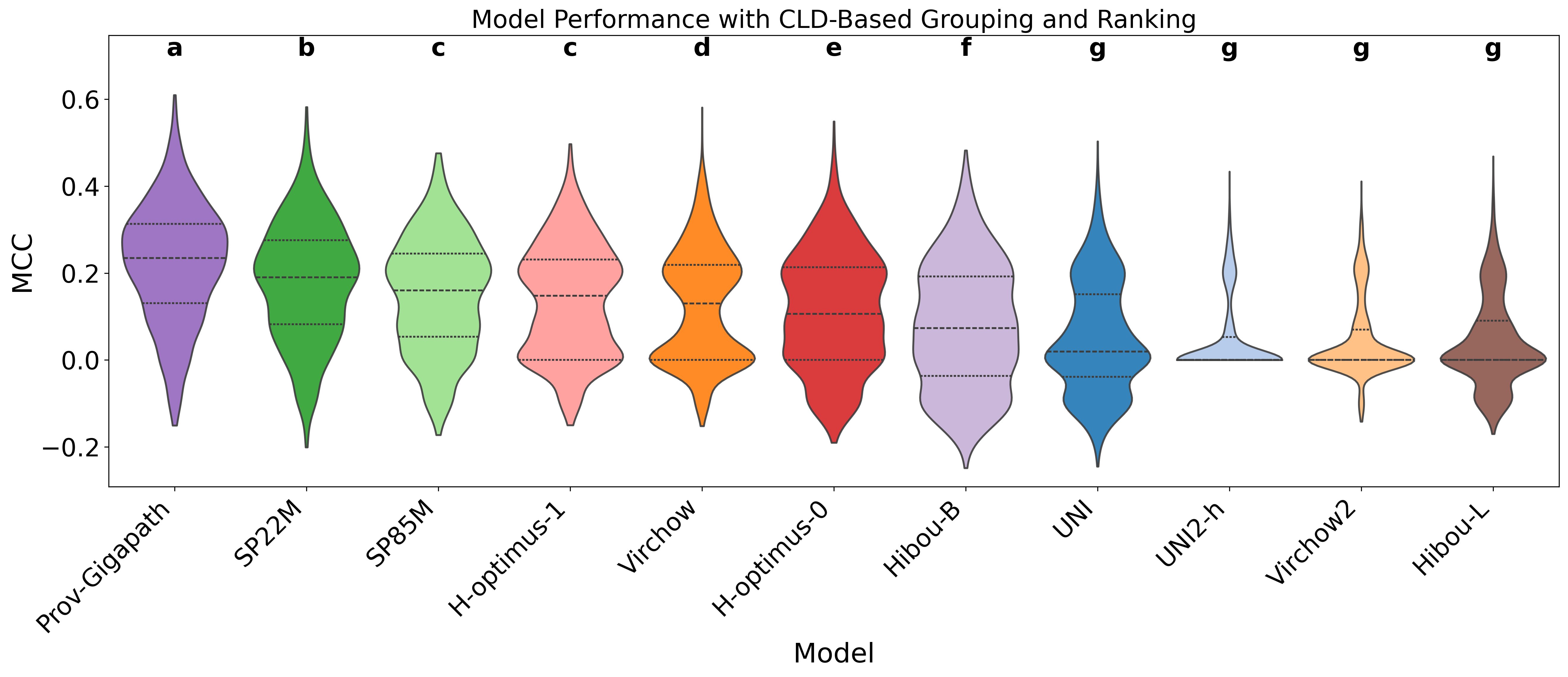}
    \caption{}
    \label{fig:kNN_artery}
\end{subfigure}
\begin{subfigure}[t]{0.33\linewidth}
    \centering
    \includegraphics[width=\linewidth]{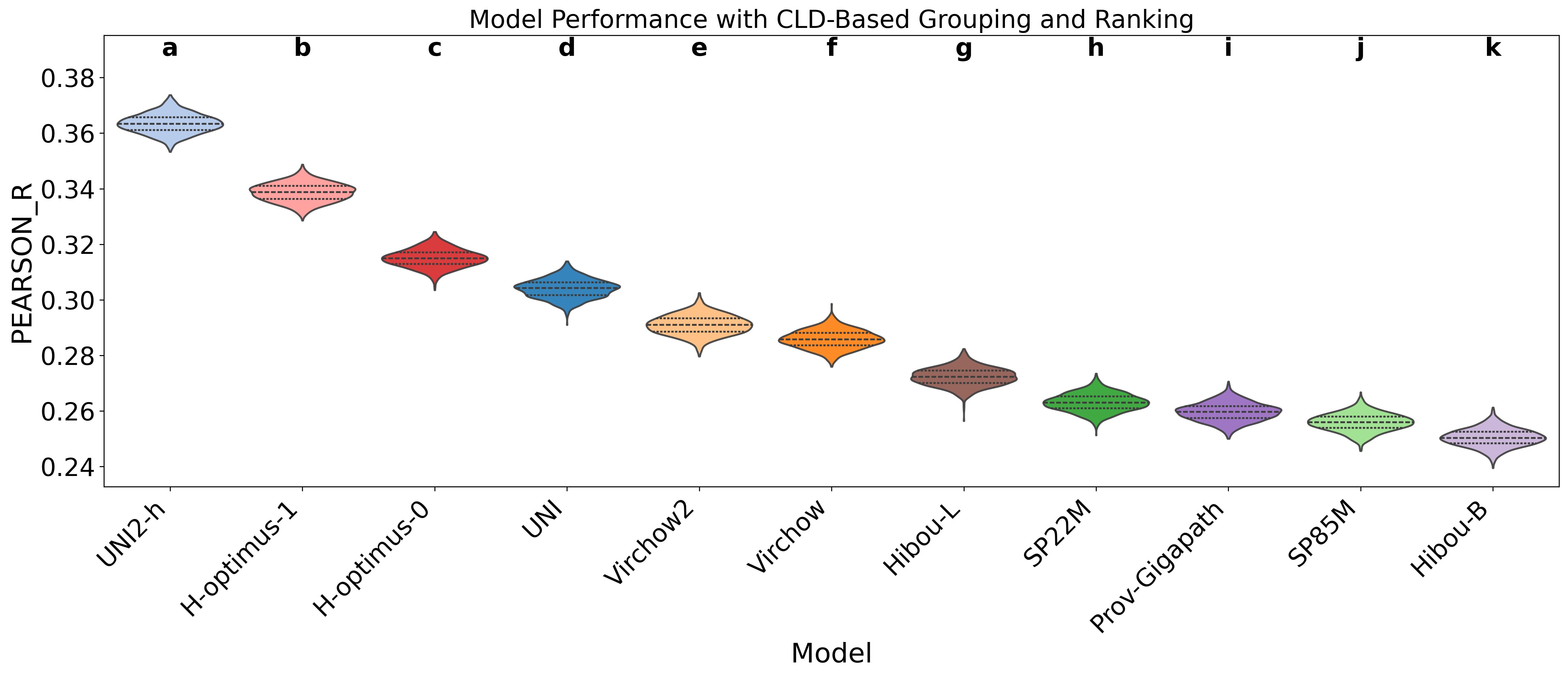}
    \caption{}
    \label{fig:cell_type_pearson}
\end{subfigure}
\begin{subfigure}[t]{0.33\linewidth}
    \centering
    \includegraphics[width=\linewidth]{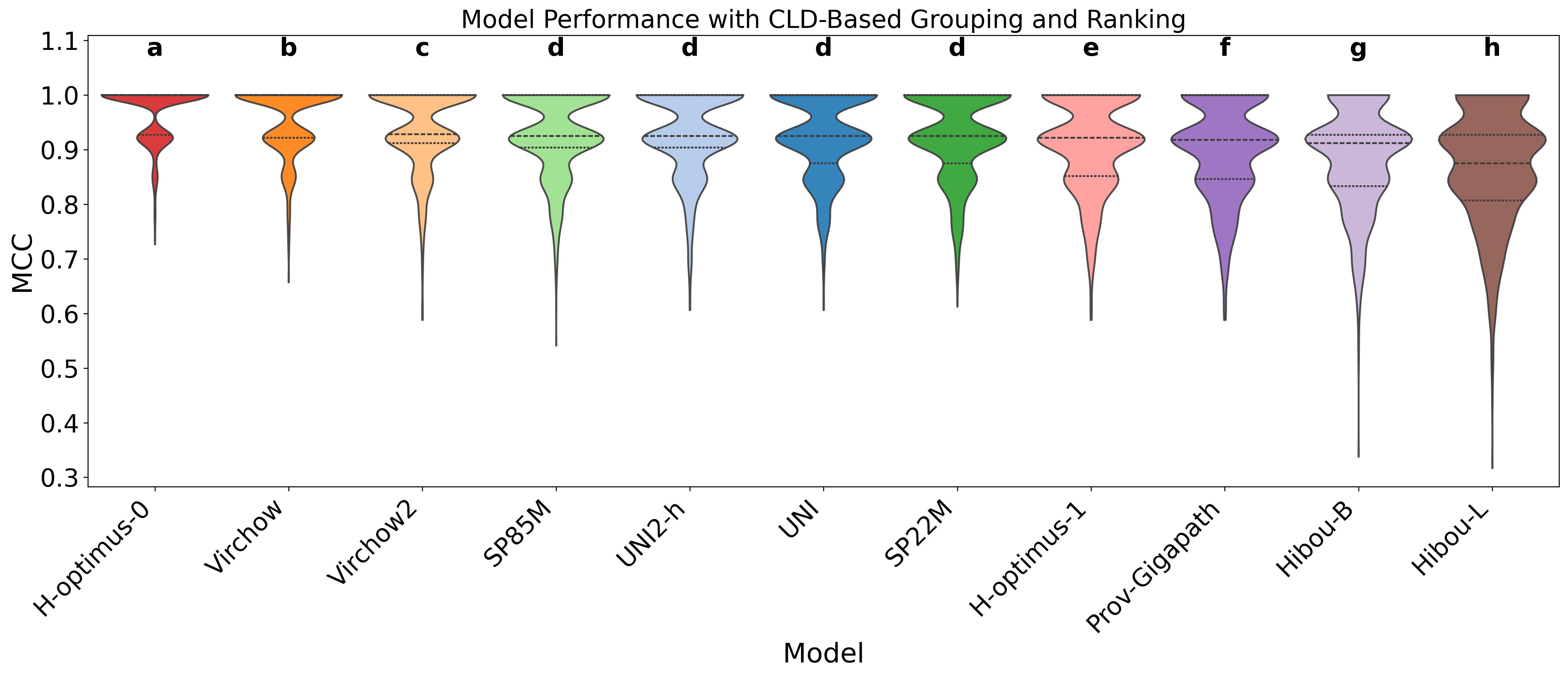}
    \caption{}
    \label{fig:MIL_DN}
\end{subfigure}

\par\medskip
\begin{subfigure}[t]{0.33\linewidth}
    \centering
    \includegraphics[width=\linewidth]{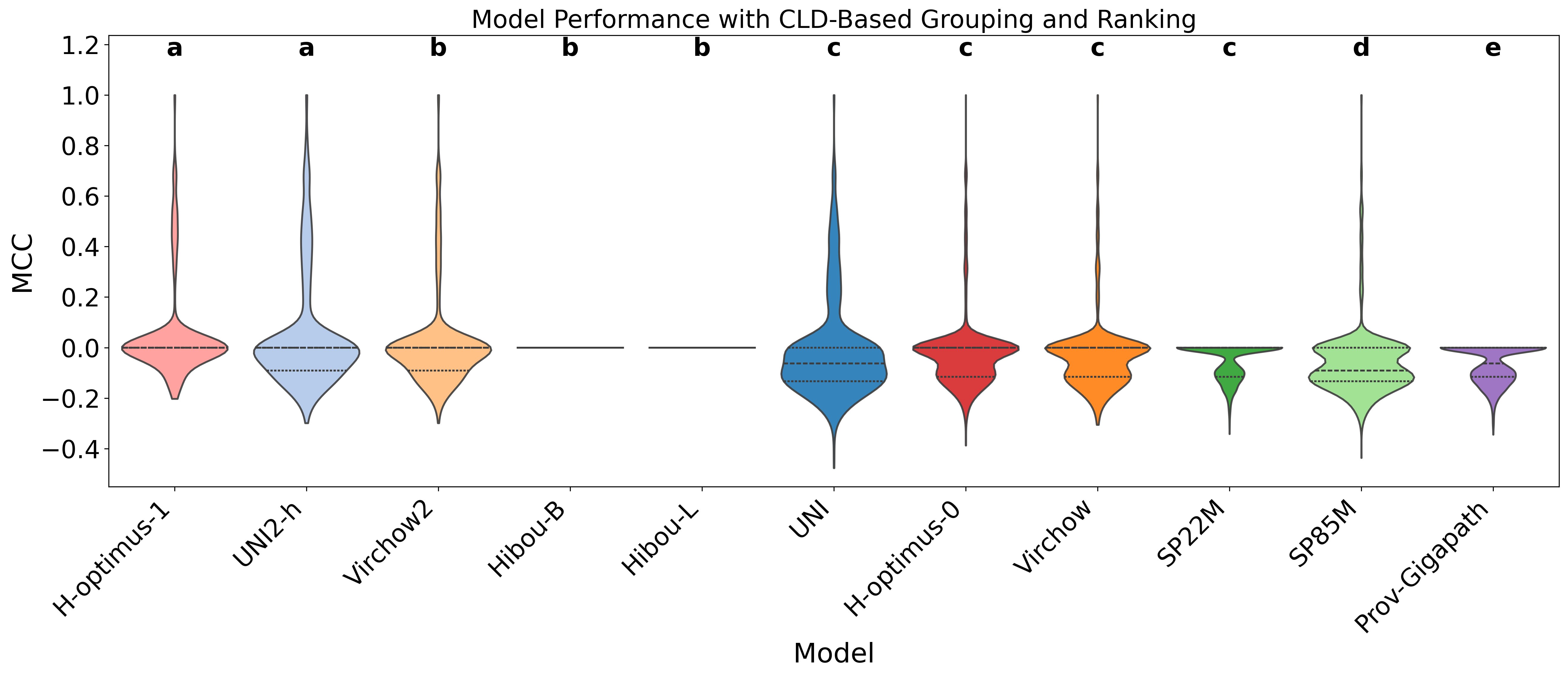}
    \caption{}
    \label{fig:MIL_MN}
\end{subfigure}
\begin{subfigure}[t]{0.33\linewidth}
    \centering
    \includegraphics[width=\linewidth]{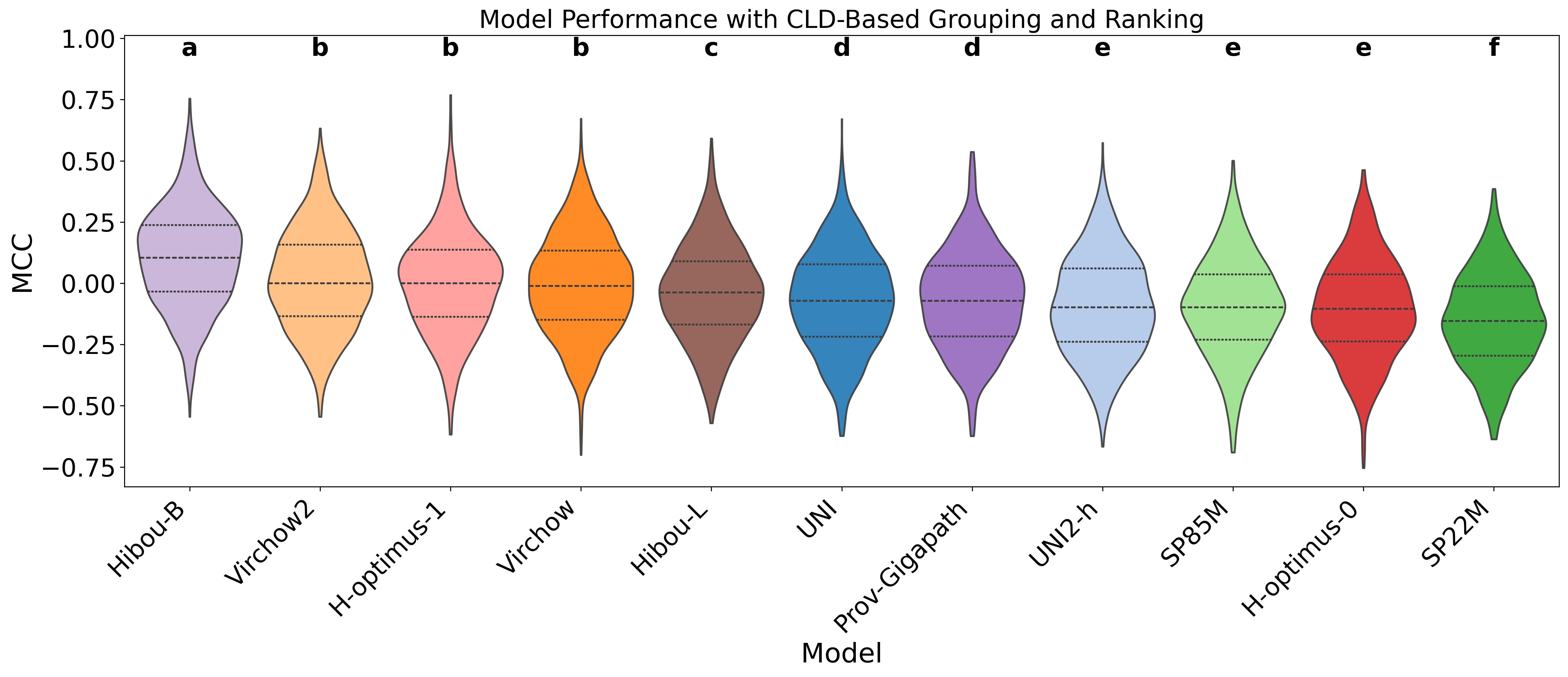}
    \caption{}
    \label{fig:MIL_Tx_classification}
\end{subfigure}
\begin{subfigure}[t]{0.33\linewidth}
    \centering
    \includegraphics[width=\linewidth]{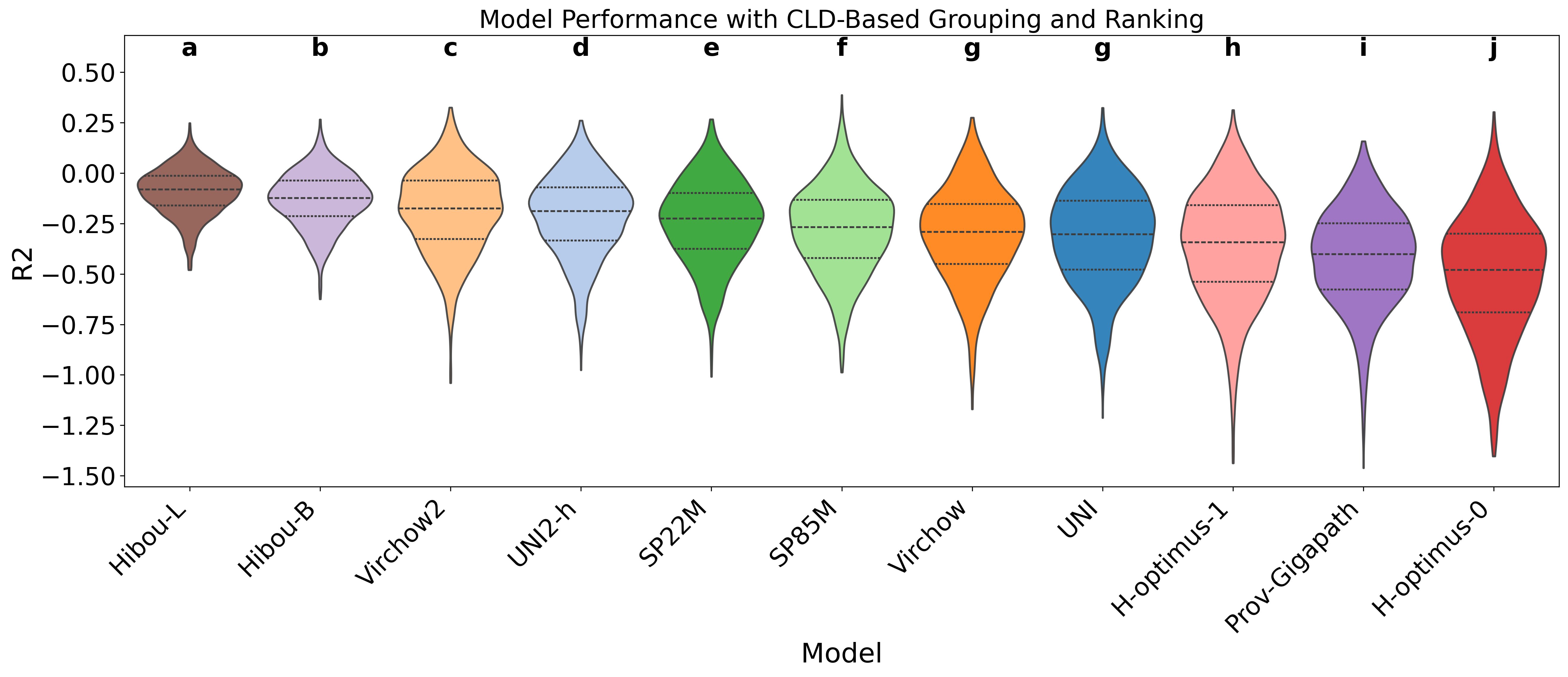}
    \caption{}
    \label{fig:MIL_Tx_regression}
\end{subfigure}

\caption{Violin plots showing the distribution of Matthews correlation coefficient (MCC) values across HFMs for multiple downstream tasks. Each subplot represents a specific task and probing method: 
(a)–(b) globally vs.\ non-globally sclerotic glomeruli classification using logistic regression (LR) and $k$NN, 
(c)–(d) GBM spike vs.\ no-GBM spike classification (LR, $k$NN), 
(e)–(f) normal vs.\ abnormal tubule classification (LR, $k$NN), 
(g)–(h) inflammation vs.\ non-inflammation classification (LR, $k$NN), 
(i)–(j) arteriolar stenosis multi-class classification (LR, $k$NN), 
(k) cell type estimation (ridge regression),
(l) diabetic nephropathy vs.\ control classification using MIL. 
(m) membranous nephropathy treatment-response prediction using multiple-instance learning (MIL), 
(n) renal transplant eGFR decline classification using MIL,
(o) renal transplant eGFR prediction using MIL.
Wider violins indicate greater performance variability, and higher median lines indicate more consistent model performance across cross-validation replicates.}
\label{fig:violin_plots}
\end{figure}

This section presents results on the comprehensive evaluation of 11 HFMs across diverse renal pathology tasks, including both tile-level and slide-level analyses. 
Our benchmarking includes (i) seven tile-level classification tasks (five classification tasks, one regression task, and a copy detection task)  focused on morphological discrimination,  molecular information estimation, and assessing the quality of embeddings. 
(ii) four slide-level evaluation tasks (three classification tasks and a regression task) focused on diagnosis and prognosis. 
Each experiment was conducted using a standardized probing framework, as explained in section~\ref{sec:methodology}, that ensures comparable assessment across HFMs and modalities. All reported metrics are statistics of 1000 bootstraps of the test-time predictions obtained using repeated group k-fold for tile-level evaluation tasks and repeated nested stratified five-fold cross-validation for slide-level evaluation tasks. Figure~\ref{fig:violin_plots} visualizes the distribution of 1000 bootstrap scores for each model and evaluation task as a violin plot. \textit{Supplementary Figure} \ref{sup:fig:heatmaps_HB_corrected}.
 shows adjusted p-value matrices from pairwise Wilcoxon signed-rank tests with Holm-Bonferroni correction, facilitating the interpretation of statistically meaningful performance differences among HFMs.
\subsection*{Tile-level tasks}

\subsubsection*{Classification tasks}

We benchmarked a total of 11 vision HFMs across seven various tasks, of which five are classification, one is regression, and one copy detection task. Tile-level classification performance was assessed using linear probing and $k$-nearest neighbor ($k$NN) classifiers.

\begin{table}[H]
\centering
\begin{subtable}{0.49\linewidth}
\centering
\caption{Logistic Regression (LR)}
\resizebox{\linewidth}{!}{%
\begin{tabular}{l l r r r r r r}
\toprule
\textbf{CLD} & \textbf{Model} & \textbf{Min} & \textbf{1st Q} & \textbf{Median} & \textbf{Mean} & \textbf{3rd Q} & \textbf{Max} \\
\midrule
a & Prov-Gigapath & 0.9217 & 0.9586 & 0.9657 & 0.9648 & 0.9715 & 0.9941 \\
a & UNI           & 0.9291 & 0.9583 & 0.9651 & 0.9645 & 0.9709 & 0.9917 \\
b & H-optimus-0   & 0.9273 & 0.9572 & 0.9645 & 0.9637 & 0.9712 & 0.9944 \\
c & UNI2-h        & 0.9250 & 0.9551 & 0.9632 & 0.9625 & 0.9695 & 0.9944 \\
c & Virchow2      & 0.9210 & 0.9549 & 0.9614 & 0.9615 & 0.9691 & 0.9903 \\
d & Hibou-L       & 0.9093 & 0.9536 & 0.9609 & 0.9607 & 0.9688 & 0.9936 \\
e & Virchow       & 0.9224 & 0.9529 & 0.9607 & 0.9600 & 0.9672 & 0.9945 \\
e & H-optimus-1   & 0.9184 & 0.9523 & 0.9599 & 0.9596 & 0.9675 & 0.9875 \\
f & Hibou-B       & 0.9052 & 0.9447 & 0.9530 & 0.9527 & 0.9613 & 0.9821 \\
f & SP85M         & 0.9130 & 0.9438 & 0.9530 & 0.9519 & 0.9601 & 0.9889 \\
g & SP22M         & 0.8993 & 0.9348 & 0.9442 & 0.9436 & 0.9528 & 0.9765 \\
\bottomrule
\label{tab:sclerotic_glomeruli_classification_lr}
\end{tabular}}
\end{subtable}
\hfill
\begin{subtable}{0.49\linewidth}
\centering
\caption{$k$-Nearest Neighbors ($k$NN)}
\resizebox{\linewidth}{!}{%
\begin{tabular}{l l r r r r r r}
\toprule
\textbf{CLD} & \textbf{Model} & \textbf{Min} & \textbf{1st Q} & \textbf{Median} & \textbf{Mean} & \textbf{3rd Q} & \textbf{Max} \\
\midrule
a & Prov-Gigapath & 0.9234 & 0.9536 & 0.9610 & 0.9602 & 0.9679 & 0.9883 \\
b & UNI           & 0.9221 & 0.9445 & 0.9533 & 0.9529 & 0.9605 & 0.9860 \\
c & H-optimus-0   & 0.9113 & 0.9436 & 0.9525 & 0.9513 & 0.9592 & 0.9877 \\
d & Virchow       & 0.8933 & 0.9274 & 0.9363 & 0.9361 & 0.9453 & 0.9717 \\
d & H-optimus-1   & 0.8810 & 0.9265 & 0.9364 & 0.9360 & 0.9459 & 0.9739 \\
e & SP85M         & 0.8797 & 0.9154 & 0.9259 & 0.9248 & 0.9353 & 0.9663 \\
f & Virchow2      & 0.8693 & 0.9117 & 0.9214 & 0.9211 & 0.9312 & 0.9632 \\
g & SP22M         & 0.8657 & 0.9096 & 0.9196 & 0.9192 & 0.9291 & 0.9687 \\
h & UNI2-h        & 0.8613 & 0.8991 & 0.9098 & 0.9096 & 0.9206 & 0.9522 \\
i & Hibou-B       & 0.8528 & 0.8917 & 0.9017 & 0.9019 & 0.9131 & 0.9480 \\
j & Hibou-L       & 0.8463 & 0.8870 & 0.8990 & 0.8984 & 0.9103 & 0.9515 \\
\bottomrule
\label{tab:sclerotic_glomeruli_classification_knn}
\end{tabular}}
\end{subtable}
\caption{ \textbf{Performance comparison of HFMs for global glomerulosclerosis classification.} Tables~\ref{tab:sclerotic_glomeruli_classification_lr} and~\ref{tab:sclerotic_glomeruli_classification_knn} rank the performance of 11 HFMs for global glomerulosclerosis classification using logistic regression and $k$-nearest neighbors classifiers, respectively. Performance is reported as the Matthews Correlation Coefficient (MCC) statistics (minimum to maximum) across 1000 bootstraps obtained from repeated stratified group five-fold cross-validation runs. Statistical ranking using Compact Letter Display (CLD): The 'CLD' column groups models based on their statistical similarity. Models sharing a letter (e.g., 'a') are not significantly different from one another ($p > .05$). Models with different letters (e.g., 'a' vs. 'b') have significantly different performance ($p < .05$, Wilcoxon signed rank test).}
\label{tab:sclerotic_glomeruli_classification}
\end{table}

In the globally sclerotic glomerular detection task based on PAS-stained images, most HFMs demonstrate strong discriminative power, achieving high median MCC values above 0.90 using both linear and $k$NN probes (Table~\ref{tab:sclerotic_glomeruli_classification}). Prov-Gigapath and UNI formed the top-performing group (CLD group 'a', $p = 0.63$, Wilcoxon signed rank test), with Prov-Gigapath achieving a median MCC of 0.9657 (Inter Quartile Range (IQR) = 0.9586–0.9715), a performance statistically indistinguishable from UNI (median MCC, 0.9651, IQR = 0.9583-0.9709) under linear probe evaluation. However, under $k$NN probe evaluation, Prov-Gigapath achieved the best performance (median MCC of 0.9610, IQR = 0.9536–0.9602).  Despite being pre-trained on H\&E-stained cancer datasets, HFMs successfully captured PAS-based morphological changes such as capillary loop collapse. This reflects cross-stain generalization, suggesting that HFMs encode stain-invariant morphometric features critical for chronic glomerular injury assessment. Tight IQR across encoders further confirms the stability and reproducibility of the learned representations.

\begin{table}[H]
\centering

\begin{subtable}{0.49\linewidth}
\centering
\caption{Logistic Regression (LR)}
\resizebox{\linewidth}{!}{%
\begin{tabular}{l l r r r r r r}
\toprule
\textbf{CLD} & \textbf{Model} & \textbf{Min} & \textbf{1st Q} & \textbf{Median} & \textbf{Mean} & \textbf{3rd Q} & \textbf{Max} \\
\midrule
a & H-optimus-1   & 0.1492 & 0.3241 & 0.3831 & 0.3790 & 0.4346 & 0.6208 \\
b & H-optimus-0   & 0.0640 & 0.2660 & 0.3265 & 0.3259 & 0.3858 & 0.5631 \\
b & Prov-Gigapath & 0.0745 & 0.2654 & 0.3243 & 0.3253 & 0.3859 & 0.5918 \\
c & UNI           & 0.0216 & 0.2570 & 0.3155 & 0.3138 & 0.3686 & 0.5393 \\
d & Virchow2      & -0.0575 & 0.2248 & 0.2823 & 0.2786 & 0.3369 & 0.5645 \\
d & Virchow       & -0.0130 & 0.2177 & 0.2784 & 0.2755 & 0.3324 & 0.5145 \\
e & UNI2-h        & -0.0150 & 0.2089 & 0.2696 & 0.2672 & 0.3303 & 0.5263 \\
f & SP85M         & -0.0801 & 0.1424 & 0.1973 & 0.2013 & 0.2610 & 0.4594 \\
g & Hibou-L       & -0.1660 & 0.0993 & 0.1578 & 0.1579 & 0.2177 & 0.4275 \\
h & Hibou-B       & -0.1431 & 0.0886 & 0.1488 & 0.1482 & 0.2080 & 0.4432 \\
i & SP22M         & -0.1349 & 0.0750 & 0.1382 & 0.1360 & 0.1952 & 0.3756 \\
\bottomrule
\label{tab:gbm_spike_classification_lr}
\end{tabular}}
\end{subtable}
\hfill
\begin{subtable}{0.49\linewidth}
\centering
\caption{$k$-Nearest Neighbors ($k$NN)}
\resizebox{\linewidth}{!}{%
\begin{tabular}{l l r r r r r r}
\toprule
\textbf{CLD} & \textbf{Model} & \textbf{Min} & \textbf{1st Q} & \textbf{Median} & \textbf{Mean} & \textbf{3rd Q} & \textbf{Max} \\
\midrule

a & H-optimus-0   & 0.0853 & 0.3316 & 0.3897 & 0.3877 & 0.4461 & 0.6432 \\
b & Hibou-L       & 0.0397 & 0.2924 & 0.3514 & 0.3494 & 0.4105 & 0.6096 \\
b & UNI           & 0.0431 & 0.2919 & 0.3486 & 0.3482 & 0.4070 & 0.6408 \\
c & Prov-Gigapath & 0.0008 & 0.2705 & 0.3329 & 0.3272 & 0.3835 & 0.5965 \\
d & Hibou-B       & 0.0063 & 0.2631 & 0.3190 & 0.3206 & 0.3808 & 0.6062 \\
e & UNI2-h        & 0.0794 & 0.2546 & 0.3124 & 0.3121 & 0.3734 & 0.5492 \\
f & H-optimus-1   & -0.0442 & 0.1982 & 0.2572 & 0.2580 & 0.3178 & 0.5271 \\
f & SP85M         & -0.0679 & 0.2029 & 0.2571 & 0.2573 & 0.3140 & 0.5287 \\
g & SP22M         & 0.0023 & 0.1889 & 0.2443 & 0.2467 & 0.3003 & 0.5166 \\
h & Virchow2      & -0.0652 & 0.1717 & 0.2313 & 0.2318 & 0.2902 & 0.4858 \\
i & Virchow       & -0.0759 & 0.1542 & 0.2144 & 0.2149 & 0.2766 & 0.5146 \\

\bottomrule
\label{tab:gbm_spike_classification_knn}
\end{tabular}}
\end{subtable}
\caption{\textbf{Performance comparison of HFMs for glomerular basement membrane spike classification.}Tables~\ref{tab:gbm_spike_classification_lr} and~\ref{tab:gbm_spike_classification_knn} rank the performance of 11 HFMs for glomerular basement membrane spike classification using logistic regression and $k$-nearest neighbors classifiers, respectively. Performance is reported as the Matthews Correlation Coefficient (MCC) statistics (minimum to maximum) across 1000 bootstraps obtained from repeated stratified group five-fold cross-validation runs. Statistical ranking using Compact Letter Display (CLD): The 'CLD' column groups models based on their statistical similarity. Models sharing a letter (e.g., 'a') are not significantly different from one another ($p > .05$). Models with different letters (e.g., 'a' vs. 'b') have significantly different performance ($p < .05$, Wilcoxon signed rank test).}
\label{tab:gbm_spike_classification}
\end{table}

In the GBM spike detection task based on PASM-stained images, HFMs exhibited moderate discriminative capability with median MCC values ranging between 0.1382 and 0.3897 across HFMs and probes. H-optimus-1 achieved the highest linear-probe performance (median MCC = 0.3831, IQR = 0.3241-0.4346), followed by H-optimus-0 and Prov-Gigapath (CLD group 'b', $p=0.76$, Wilcoxon signed rank test). Under $k$NN probing, H-optimus-0 outperformed others (median MCC = 0.3897, IQR = 0.3316-0.4461),  followed by Hibou-L and UNI (CLD group 'b', $p=0.87$, Wilcoxon signed rank test) (Table~\ref{tab:gbm_spike_classification}). Despite being trained exclusively on H\&E datasets, these models showed limited but measurable transfer to the PASM stain domain, implying that their feature spaces encode contour-based information relevant to basement membrane irregularities. Nevertheless, the relatively low to moderate overall performance underscores the intrinsic difficulty of detecting GBM alterations that are subtle even for pathologists and often require electron microscopy confirmation.

\begin{table}[H]
\centering

\begin{subtable}{0.49\linewidth}
\centering
\caption{Logistic Regression (LR)}
\resizebox{\linewidth}{!}{%
\begin{tabular}{l l r r r r r r}
\toprule
\textbf{CLD} & \textbf{Model} & \textbf{Min} & \textbf{1st Q} & \textbf{Median} & \textbf{Mean} & \textbf{3rd Q} & \textbf{Max} \\
\midrule
a & UNI2-h        & -0.1544 & 0.0241 & 0.0671 & 0.0689 & 0.1114 & 0.2747 \\
b & Virchow       & -0.2112 & 0.0102 & 0.0520 & 0.0535 & 0.0985 & 0.2629 \\
b & Virchow2      & -0.1417 & 0.0064 & 0.0503 & 0.0494 & 0.0940 & 0.2529 \\
b & Prov-Gigapath & -0.2412 & 0.0010 & 0.0461 & 0.0468 & 0.0899 & 0.2808 \\
c & Hibou-B       & -0.1716 & -0.0185 & 0.0273 & 0.0287 & 0.0727 & 0.2239 \\
d & Hibou-L       & -0.1684 & -0.0250 & 0.0173 & 0.0165 & 0.0587 & 0.2169 \\
d & H-optimus-1   & -0.2054 & -0.0280 & 0.0165 & 0.0158 & 0.0590 & 0.2601 \\
e & SP85M         & -0.1826 & -0.0340 & 0.0084 & 0.0065 & 0.0494 & 0.2087 \\
f & SP22M         & -0.2016 & -0.0421 & 0.0004 & -0.0010 & 0.0420 & 0.1970 \\
f & UNI           & -0.2841 & -0.0516 & -0.0065 & -0.0048 & 0.0402 & 0.1902 \\
g & H-optimus-0   & -0.2239 & -0.0515 & -0.0097 & -0.0104 & 0.0336 & 0.1843 \\
\bottomrule
\label{tab:tubule_classification_lr}
\end{tabular}}
\end{subtable}
\hfill
\begin{subtable}{0.49\linewidth}
\centering
\caption{$k$-Nearest Neighbors ($k$NN)}
\resizebox{\linewidth}{!}{%
\begin{tabular}{l l r r r r r r}
\toprule
\textbf{CLD} & \textbf{Model} & \textbf{Min} & \textbf{1st Q} & \textbf{Median} & \textbf{Mean} & \textbf{3rd Q} & \textbf{Max} \\
\midrule

a & UNI2-h        & -0.1597 & 0.0161 & 0.0613 & 0.0601 & 0.1031 & 0.2786 \\
b & SP22M         & -0.1523 & 0.0101 & 0.0514 & 0.0520 & 0.0984 & 0.2506 \\
c & Hibou-B       & -0.1596 & 0.0064 & 0.0490 & 0.0488 & 0.0960 & 0.2346 \\
c & UNI           & -0.1607 & 0.0034 & 0.0484 & 0.0481 & 0.0916 & 0.2450 \\
c & Prov-Gigapath & -0.1573 & 0.0053 & 0.0495 & 0.0471 & 0.0902 & 0.2384 \\
c & Virchow       & -0.1501 & -0.0014 & 0.0434 & 0.0438 & 0.0858 & 0.2418 \\
d & H-optimus-0   & -0.1882 & -0.0045 & 0.0400 & 0.0411 & 0.0834 & 0.2499 \\
e & Virchow2      & -0.1952 & -0.0099 & 0.0333 & 0.0325 & 0.0753 & 0.2066 \\
e & H-optimus-1   & -0.1900 & -0.0170 & 0.0258 & 0.0261 & 0.0683 & 0.2247 \\
f & Hibou-L       & -0.1498 & -0.0165 & 0.0236 & 0.0246 & 0.0671 & 0.2332 \\
g & SP85M         & -0.1847 & -0.0323 & 0.0157 & 0.0153 & 0.0580 & 0.2195 \\

\bottomrule
\label{tab:tubule_classification_kNN}
\end{tabular}}
\end{subtable}
\caption{\textbf{Performance comparison of HFMs for tubular classification.} Tables~\ref{tab:tubule_classification_lr} and~\ref{tab:tubule_classification_kNN} rank the performance of 11 HFMs for tubular classification using logistic regression and $k$-nearest neighbors classifiers, respectively. Performance is reported as the Matthews Correlation Coefficient (MCC) statistics (minimum to maximum) across 1000 bootstraps obtained from repeated stratified group five-fold cross-validation runs. Statistical Ranking using Compact Letter Display (CLD): The 'CLD' column groups models based on their statistical similarity. Models sharing a letter (e.g., 'a') are not significantly different from one another ($p > .05$). Models with different letters (e.g., 'a' vs. 'b') have significantly different performance ($p < .05$, Wilcoxon signed rank test).}
\label{tab:tubule_classification}

\end{table}

In the tubule classification task based on H\&E-stained images, HFMs showed near-random performance, with mean MCCs fluctuating around zero for both linear and $k$NN probes, with median MCC ranging from $−0.0097$ to 0.0671 across HFMs and probes. UNI2-h yielded the highest performance under both linear (median MCC  = 0.0671, IQR = 0.0241-0.1114) and $k$NN probing (median MCC  $=$ 0.0613, IQR $=$ 0.0161-0.1031) (Table~\ref{tab:tubule_classification}). Even though these HFMs were trained on massive H\&E datasets, they were unable to distinguish between reference and abnormal tubules labeled from molecular information. This finding highlights that current vision HFMs could not capture the cell state-driven changes in tubular morphology that are present in the molecular information, rather than gross structural change.

\begin{table}[H]
\centering
\begin{subtable}{0.49\linewidth}
\centering
\caption{Logistic Regression (LR)}
\resizebox{\linewidth}{!}{%
\begin{tabular}{l l r r r r r r}
\toprule
\textbf{CLD} & \textbf{Model} & \textbf{Min} & \textbf{1st Q} & \textbf{Median} & \textbf{Mean} & \textbf{3rd Q} & \textbf{Max} \\
\midrule
a & SP22M         & 0.4048 & 0.4597 & 0.4735 & 0.4736 & 0.4870 & 0.5370 \\
b & Hibou-B       & 0.3753 & 0.4324 & 0.4472 & 0.4469 & 0.4610 & 0.5222 \\
b & UNI           & 0.3732 & 0.4299 & 0.4468 & 0.4467 & 0.4626 & 0.5136 \\
c & SP85M         & 0.3721 & 0.4213 & 0.4357 & 0.4365 & 0.4518 & 0.5087 \\
d & Hibou-L       & 0.3385 & 0.4025 & 0.4156 & 0.4166 & 0.4318 & 0.4838 \\
d & H-optimus-1   & 0.3496 & 0.3984 & 0.4156 & 0.4154 & 0.4313 & 0.4756 \\
e & Virchow2      & 0.3318 & 0.3870 & 0.4025 & 0.4028 & 0.4186 & 0.4734 \\
f & Virchow       & 0.3237 & 0.3814 & 0.3979 & 0.3974 & 0.4132 & 0.4710 \\
g & H-optimus-0   & 0.3104 & 0.3762 & 0.3912 & 0.3917 & 0.4066 & 0.4766 \\
h & UNI2-h        & 0.3169 & 0.3698 & 0.3841 & 0.3845 & 0.3999 & 0.4595 \\
i & Prov-Gigapath & 0.3089 & 0.3622 & 0.3805 & 0.3788 & 0.3949 & 0.4474 \\
\bottomrule
\label{tab:inflammation_classification_lr}
\end{tabular}}
\end{subtable}
\hfill
\begin{subtable}{0.49\linewidth}
\centering
\caption{$k$-Nearest Neighbors ($k$NN)}
\resizebox{\linewidth}{!}{%
\begin{tabular}{l l r r r r r r}
\toprule
\textbf{CLD} & \textbf{Model} & \textbf{Min} & \textbf{1st Q} & \textbf{Median} & \textbf{Mean} & \textbf{3rd Q} & \textbf{Max} \\
\midrule

a & Prov-Gigapath & 0.3908 & 0.4594 & 0.4733 & 0.4735 & 0.4883 & 0.5435 \\
b & H-optimus-0   & 0.4134 & 0.4565 & 0.4719 & 0.4721 & 0.4885 & 0.5373 \\
c & UNI           & 0.3964 & 0.4488 & 0.4625 & 0.4628 & 0.4765 & 0.5394 \\
d & Hibou-B       & 0.3879 & 0.4424 & 0.4574 & 0.4575 & 0.4727 & 0.5346 \\
e & SP22M         & 0.3813 & 0.4357 & 0.4497 & 0.4496 & 0.4651 & 0.5244 \\
e & SP85M         & 0.3876 & 0.4355 & 0.4500 & 0.4495 & 0.4643 & 0.5274 \\
e & Virchow       & 0.3703 & 0.4335 & 0.4494 & 0.4492 & 0.4643 & 0.5350 \\
f & UNI2-h        & 0.3808 & 0.4287 & 0.4439 & 0.4440 & 0.4584 & 0.5157 \\
g & H-optimus-1   & 0.3702 & 0.4249 & 0.4389 & 0.4394 & 0.4554 & 0.5082 \\
h & Hibou-L       & 0.3747 & 0.4225 & 0.4381 & 0.4377 & 0.4527 & 0.4994 \\
i & Virchow2      & 0.3634 & 0.4166 & 0.4308 & 0.4313 & 0.4466 & 0.5021 \\

\bottomrule
\label{tab:inflammation_classification_knn}
\end{tabular}}
\end{subtable}
\caption{\textbf{Performance comparison of HFMs for inflammation classification.} Tables~\ref{tab:inflammation_classification_lr} and~\ref{tab:inflammation_classification_knn} rank the performance of 11 HFMs for inflammation classification using logistic regression and $k$-nearest neighbors classifiers, respectively. Performance is reported as the Matthews Correlation Coefficient (MCC) statistics (minimum to maximum) across 1000 bootstraps obtained from repeated stratified group five-fold cross-validation runs. Statistical Ranking using Compact Letter Display (CLD): The 'CLD' column groups models based on their statistical similarity. Models sharing a letter (e.g., 'a') are not significantly different from one another ($p > .05$). Models with different letters (e.g., 'a' vs. 'b') have significantly different performance ($p < .05$, Wilcoxon signed rank test).}
\label{tab:inflammation_classification}

\end{table}

\begin{table}[H]
\centering
\begin{subtable}{0.49\linewidth}
\centering
\caption{Logistic Regression (LR)}
\resizebox{\linewidth}{!}{%
\begin{tabular}{l l r r r r r r}
\toprule
\textbf{CLD} & \textbf{Model} & \textbf{Min} & \textbf{1st Q} & \textbf{Median} & \textbf{Mean} & \textbf{3rd Q} & \textbf{Max} \\
\midrule
a & Virchow       & -0.0746 & 0.3229 & 0.3913 & 0.3956 & 0.4677 & 0.7476 \\
b & Hibou-L       &  0.0060 & 0.2995 & 0.3752 & 0.3786 & 0.4562 & 0.7960 \\
c & H-optimus-0   &  0.0246 & 0.2904 & 0.3659 & 0.3700 & 0.4473 & 0.7890 \\
c & Virchow2      &  0.0033 & 0.2848 & 0.3612 & 0.3642 & 0.4501 & 0.8132 \\
d & H-optimus-1   & -0.0467 & 0.2699 & 0.3565 & 0.3551 & 0.4370 & 0.7348 \\
d & Prov-Gigapath &  0.0399 & 0.2689 & 0.3460 & 0.3483 & 0.4248 & 0.6708 \\
d & UNI2-h        & -0.0186 & 0.2573 & 0.3402 & 0.3437 & 0.4219 & 0.7336 \\
e & Hibou-B       & -0.0157 & 0.2327 & 0.3142 & 0.3132 & 0.3893 & 0.6594 \\
e & SP22M         & -0.0630 & 0.2319 & 0.3083 & 0.3098 & 0.3857 & 0.6639 \\
e & SP85M         & -0.0487 & 0.2301 & 0.3080 & 0.3056 & 0.3776 & 0.6620 \\
f & UNI           & -0.1170 & 0.1865 & 0.2740 & 0.2702 & 0.3491 & 0.6778 \\
\bottomrule
\label{tab:artery_stenosis_classification_lr}
\end{tabular}}
\end{subtable}
\hfill
\begin{subtable}{0.49\linewidth}
\centering
\caption{$k$-Nearest Neighbors ($k$NN)}
\resizebox{\linewidth}{!}{%
\begin{tabular}{l l r r r r r r}
\toprule
\textbf{CLD} & \textbf{Model} & \textbf{Min} & \textbf{1st Q} & \textbf{Median} & \textbf{Mean} & \textbf{3rd Q} & \textbf{Max} \\
\midrule

a & Prov-Gigapath & -0.1507 & 0.1305 & 0.2349 & 0.2226 & 0.3131 & 0.6097 \\
b & SP22M              & -0.2009 & 0.0819 & 0.1904 & 0.1797 & 0.2757 & 0.5821 \\
c & SP85M              & -0.1725 & 0.0537 & 0.1602 & 0.1498 & 0.2448 & 0.4758 \\
c & H-optimus-1 & -0.1500 & 0.0000 & 0.1480 & 0.1386 & 0.2315 & 0.4970 \\
d & Virchow            & -0.1520 & 0.0000 & 0.1298 & 0.1248 & 0.2187 & 0.5811 \\
e & H-optimus-0 & -0.1903 & 0.0000 & 0.1062 & 0.1128 & 0.2136 & 0.5490 \\
f & Hibou-B       & -0.2486 & -0.0365 & 0.0733 & 0.0770 & 0.1920 & 0.4824 \\
g & UNI                & -0.2451 & -0.0386 & 0.0195 & 0.0513 & 0.1509 & 0.5031 \\
g & UNI2-h        &  0.0000 & 0.0000 & 0.0000 & 0.0464 & 0.0526 & 0.4336 \\
g & Virchow2           & -0.1417 & 0.0000 & 0.0000 & 0.0452 & 0.0696 & 0.4111 \\
g & Hibou-L       & -0.1702 & 0.0000 & 0.0000 & 0.0409 & 0.0907 & 0.4686 \\

\bottomrule
\label{tab:artery_stenosis_classification_knn}
\end{tabular}}
\end{subtable}
\caption{\textbf{Performance comparison of HFMs for arterial stenosis classification.} Tables~\ref{tab:artery_stenosis_classification_lr} and~\ref{tab:artery_stenosis_classification_knn} rank the performance of 11 HFMs for arterial stenosis classification using logistic regression and $k$-nearest neighbors classifiers, respectively. Performance is reported as the Matthews Correlation Coefficient (MCC) statistics (minimum to maximum) across 1000 bootstraps obtained from repeated stratified group five-fold cross-validation runs. Statistical Ranking using Compact Letter Display (CLD): The 'CLD' column groups models based on their statistical similarity. Models sharing a letter (e.g., 'a') are not significantly different from one another ($p > .05$). Models with different letters (e.g., 'a' vs. 'b') have significantly different performance ($p < .05$, Wilcoxon signed rank test).}
\label{tab:artery_stenosis_classification}
\end{table}

In the inflammation classification task based on H\&E-stained images, HFMs demonstrate moderate discriminative ability, with median MCCs ranging from 0.3805 to 0.4735 across HFMs and probes. SP22M achieved the highest linear-probe performance (median MCC = 0.4735, IQR = 0.4597-0.4870), followed by Hibou-B and UNI (CLD group 'b', $p=0.94$, Wilcoxon signed rank test). Under $k$NN probing, Prov-Gigapath outperformed others (median MCC = 0.4733, IQR = 0.4594-0.4883), followed by H-optimus-0 (median MCC = 0.4719, IQR = 0.4565-0.4885) (Table~\ref{tab:inflammation_classification}). Since the majority of HFMs are being pretrained on cancer-oriented H\&E datasets, where inflammation is high, these HFMs effectively capture features involving nuclear crowding and immune cell clustering, which suggest inflammation, implying that their internal representations generalize to immune-driven injury patterns. However, inflammation is a complex biological process defined by immune cell heterogeneity and cytokine signaling, which are not directly visible on histology. The moderate yet stable performance indicates that HFMs encode consistent tissue-level correlates of inflammation.

In the artery stenosis classification task based on PAS-stained images, HFMs demonstrate moderate performance, with median MCCs ranging from 0.0000 to 0.3913 across HFMs and probes. Virchow reached the highest score (median MCC = 0.3956, IQR = 0.3229-0.4677), followed closely by Hibou-L (median MCC = 0.3752, IQR = 0.2995-0.4562) under linear probing evaluation (Table~\ref{tab:artery_stenosis_classification}). Under $k$NN probe, Prov-Gigapath (median MCC = 0.2349, IQR = 0.1305-0.3131) outperformed all other HFMs. HFMs have moderate discriminative power in distinguishing mild, moderate, and severe vascular stenosis despite the color-domain shift from the H\&E training sets. Performance dispersion across HFMs reflects the inherent difficulty of classifying intermediate stages of vascular sclerosis, whose exhibited changes are continuous rather than clearly separated morphological variations.

\begin{table}[H]
\centering
\begin{tabular}{l l r r r r r r}
\toprule
\textbf{CLD} & \textbf{Model} & \textbf{Min} & \textbf{1st Quart.} & \textbf{Median} & \textbf{Mean} & \textbf{3rd Quart.} & \textbf{Max} \\
\midrule
a & UNI2-h        & 0.3533 & 0.3612 & 0.3634 & 0.3635 & 0.3657 & 0.3737 \\
b & H-optimus-1 & 0.3286 & 0.3364 & 0.3388 & 0.3387 & 0.3411 & 0.3487 \\
c & H-optimus-0 & 0.3035 & 0.3129 & 0.3150 & 0.3150 & 0.3171 & 0.3245 \\
d & UNI                & 0.2910 & 0.3017 & 0.3042 & 0.3041 & 0.3063 & 0.3139 \\
e & Virchow2           & 0.2797 & 0.2886 & 0.2910 & 0.2910 & 0.2934 & 0.3025 \\
f & Virchow            & 0.2760 & 0.2836 & 0.2858 & 0.2858 & 0.2881 & 0.2986 \\
g & Hibou-L       & 0.2564 & 0.2701 & 0.2723 & 0.2723 & 0.2745 & 0.2824 \\
h & SP22M              & 0.2513 & 0.2610 & 0.2630 & 0.2631 & 0.2653 & 0.2735 \\
i & Prov-Gigapath & 0.2500 & 0.2574 & 0.2597 & 0.2596 & 0.2617 & 0.2706 \\
j & SP85M              & 0.2456 & 0.2539 & 0.2560 & 0.2559 & 0.2580 & 0.2668 \\
k & Hibou-B       & 0.2395 & 0.2483 & 0.2503 & 0.2504 & 0.2525 & 0.2613 \\
\bottomrule
\end{tabular}
\caption{\textbf{Performance comparison of HFMs for cell type estimation.} Table~\ref{tab:cell_type_estimation} ranks the performance of 11 HFMs for cell type estimation using ridge regression. Performance is reported as the Pearson Correlation Coefficient (PCC) statistics (minimum to maximum) across 1000 bootstraps obtained from repeated group five-fold cross-validation runs. Ranking of model performance for cell type estimation. Statistical Ranking using Compact Letter Display (CLD): The 'CLD' column groups models based on their statistical similarity. Models sharing a letter (e.g., 'a') are not significantly different from one another ($p > .05$). Models with different letters (e.g., 'a' vs. 'b') have significantly different performance ($p < .05$, Wilcoxon signed rank test).}
\label{tab:cell_type_estimation}

\end{table}

\subsubsection*{Regression task}
In the cell-type regression task based on H\&E-stained images, performed using ridge regression, we see median Pearson's correlation coefficients (PCC) ranging from 0.2503 to 0.3634 (Table~\ref{tab:cell_type_estimation}). PCC evaluates the correlation between predicted and ground-truth cellular proportions derived from Visium ST. UNI2-h obtained the best performance (median PCC = 0.3634, IQR = 0.3612-0.3657), followed by H-optimus-1 (median PCC = 0.3388, IQR = 0.3364-0.3411). While these moderate correlations demonstrate that HFMs encode biologically relevant features reflecting local cellular composition, the results also reveal that morphological embeddings alone can only partially approximate molecular cell-state diversity. 

\begin{table}[h!]
  \centering
  \small
  \renewcommand{\arraystretch}{1.2}
  \resizebox{\textwidth}{!}{%
    \begin{tabular}{l*{11}{c}}
      \toprule
      Augmentation 
        & H-optimus-0 & H-optimus-1 & Hibou-B & Hibou-L & Prov-Gigapath & SP22M & SP85M 
        & UNI & UNI2-h & Virchow & Virchow2 \\
      \midrule
      color  
         & 0.9343  & 0.9359  & 0.8997 & 0.8735  & 0.9538 & 0.8669 & 0.8798 & 0.9480 & 0.9397 & 0.9255 & \textbf{0.9555} \\
      deform 
       & \textbf{1.0000} & \textbf{1.0000} & \textbf{1.0000} & \textbf{1.0000} & \textbf{1.0000} & \textbf{1.0000} & \textbf{1.0000} & \textbf{1.0000} & \textbf{1.0000} & 0.9717 & 0.9995 \\
      geo    
        & 0.9979 & 0.9954 & 0.9891 & 0.8827 & \textbf{1.0000} & 0.9987 & 0.9991 & 0.9987 & 0.9879 & 0.9663 & 0.9575 \\
      noise  
        & 0.8124 & 0.6715 & 0.2457 & 0.4324 & \textbf{0.8656} & 0.8074 & 0.6307 & 0.8112 & 0.7758 & 0.6216 & 0.6199 \\
      \bottomrule
    \end{tabular}%
  }
  \caption{Top-1 classification accuracy for HFMs under four augmentation types (color, deform, geometric, noise), evaluating representation robustness.}
  \label{tab:top1_accuracy_only}
\end{table}

\subsubsection*{Copy detection task}
To evaluate the stability of HFM representations under common visual perturbations, we measured top-1 classification accuracy across four augmentation types: color, geometric, deformation, and noise (Table~\ref{tab:top1_accuracy_only}). All models exhibited near-perfect robustness to deformation, color, and geometric transformations, with several HFMs achieving accuracies close to 1.0. This indicates that HFMs learn highly invariant features resilient to these augmentations. Performance under noise transformations showed the greatest variability, with top-1 accuracies ranging from 0.2457 to 0.8656, highlighting that noise levels remain a challenging perturbation for certain architectures. Deformation-based perturbations were handled exceptionally well ($\geq$0.99 for most models), reflecting that HFMs encode shape-consistent representations robust to local warping. Overall, these findings confirm that large-scale HFMs are robust to deformation, color, and geometric variations but differ in noise stability.

\subsection*{Slide-level tasks}
\bigskip
\subsubsection*{Classification tasks}
Slide-level classification performance was assessed using an ABMIL framework. For slide-level specific tasks, all evaluations were conducted using repeated nested stratified K-fold cross validation (as shown in Figure~\ref{fig:slide_level_benchmark}), consisting of five outer folds and three random seeds (15 total train-test evaluations per task). This design produced up to three independent test-time predictions per WSI, which were aggregated to obtain final patient-level HFM performance estimates.

\begin{table}[h!]
\centering
\begin{tabular}{l l r r r r r r}
\hline
\textbf{CLD} & \textbf{Model} & \textbf{Min} & \textbf{1st Quart.} & \textbf{Median} & \textbf{Mean} & \textbf{3rd Quart.} & \textbf{Max} \\
\hline
a & H-optimus-0     & 0.7266 & 0.9270 & 1.0000 & 0.9685 & 1.0000 & 1.0000 \\
b & Virchow         & 0.6574 & 0.9220 & 1.0000 & 0.9538 & 1.0000 & 1.0000 \\
c & Virchow2        & 0.5883 & 0.9117 & 0.9282 & 0.9381 & 1.0000 & 1.0000 \\
d & SP85M           & 0.5416 & 0.9035 & 0.9250 & 0.9269 & 1.0000 & 1.0000 \\
d & UNI2-h          & 0.6064 & 0.9035 & 0.9250 & 0.9268 & 1.0000 & 1.0000 \\
d & UNI             & 0.6064 & 0.8748 & 0.9250 & 0.9240 & 1.0000 & 1.0000 \\
d & SP22M           & 0.6128 & 0.8748 & 0.9250 & 0.9232 & 1.0000 & 1.0000 \\
e & H-optimus-1     & 0.5883 & 0.8516 & 0.9220 & 0.9076 & 1.0000 & 1.0000 \\
f & Prov-Gigapath   & 0.5883 & 0.8460 & 0.9177 & 0.8973 & 1.0000 & 1.0000 \\
g & Hibou-B         & 0.3380 & 0.8333 & 0.9117 & 0.8821 & 0.9270 & 1.0000 \\
h & Hibou-L         & 0.3171 & 0.8071 & 0.8748 & 0.8675 & 0.9270 & 1.0000 \\
\hline
\end{tabular}
\caption{\textbf{Performance comparison of HFMs for healthy reference samples vs. DN classification.} Table~\ref{tab:DN_vs_ref} ranks the performance of 11 HFMs for healthy reference samples vs. DN classification using attention based multiple instance learning (ABMIL) classifier. Performance is reported as the Matthews Correlation Coefficient (MCC) statistics (minimum to maximum) across 1000 bootstraps obtained from repeated nested stratified five-fold cross-validation runs. Statistical ranking using Compact Letter Display (CLD): The 'CLD' column groups models based on their statistical similarity. Models sharing a letter (e.g., 'a') are not significantly different from one another ($p > .05$). Models with different letters (e.g., 'a' vs. 'b') have significantly different performance ($p < .05$, Wilcoxon signed rank test).}
\label{tab:DN_vs_ref}

\end{table}

In the slide-level task of differentiating the DN WSIs from healthy reference WSIs, performed using PAS-stained WSIs, HFMs demonstrate strong discriminative power, with median MCCs exceeding 0.8748 for all HFMs (Table~\ref{tab:DN_vs_ref}). H-optimus-0 (median MCC = 1.0000, IQR = 0.9270-1.0000) outperformed all other HFMs, followed by Virchow (median MCC = 1.0000, IQR = 0.9220-1.0000). This near-perfect separation underscores the capacity of H\&E-trained HFMs to generalize robustly to PAS-stained renal tissue. The results indicate that HFMs encode high-level morphological abstractions that go beyond the stain variations, enabling effective cross-domain diagnostic discrimination in kidney pathology.

\begin{table}[h!]
\centering
\begin{tabular}{l l r r r r r r}
\hline
\textbf{CLD} & \textbf{Model} & \textbf{Min} & \textbf{1st Quart.} & \textbf{Median} & \textbf{Mean} & \textbf{3rd Quart.} & \textbf{Max} \\
\hline
a & H-optimus-1     & -0.2025 & 0.0000 & 0.0000 & 0.0488 & 0.0000 & 1.0000 \\
a & UNI2-h          & -0.2988 & -0.0913 & 0.0000 & 0.0484 & 0.0000 & 1.0000 \\
b & Virchow2        & -0.2988 & -0.0913 & 0.0000 & 0.0148 & 0.0000 & 1.0000 \\
b & Hibou-L         &  0.0000 & 0.0000 & 0.0000 & 0.0000 & 0.0000 & 0.0000 \\
b & Hibou-B         &  0.0000 & 0.0000 & 0.0000 & 0.0000 & 0.0000 & 0.0000 \\
c & UNI             & -0.4767 & -0.1333 & -0.0625 & -0.0099 & 0.0000 & 1.0000 \\
c & H-optimus-0     & -0.3873 & -0.1157 & 0.0000 & -0.0308 & 0.0000 & 1.0000 \\
c & Virchow         & -0.3055 & -0.1157 & 0.0000 & -0.0368 & 0.0000 & 1.0000 \\
c & SP22M           & -0.3419 & -0.1157 & 0.0000 & -0.0512 & 0.0000 & 0.0000 \\
d & SP85M           & -0.4364 & -0.1333 & -0.0913 & -0.0627 & 0.0000 & 1.0000 \\
e & Prov-Gigapath   & -0.3443 & -0.1157 & -0.0625 & -0.0667 & 0.0000 & 0.0000 \\

\hline
\end{tabular}
\caption{\textbf{Performance comparison of HFMs for MN treatment response classification.} Table~\ref{tab:MN_response} ranks the performance of 11 HFMs for MN treatment response classification using attention based multiple instance learning (ABMIL) classifier. Performance is reported as the Matthews Correlation Coefficient (MCC) statistics (minimum to maximum) across 1000 bootstraps obtained from repeated nested stratified five-fold cross-validation runs. Statistical Ranking using Compact Letter Display (CLD): The 'CLD' column groups models based on their statistical similarity. Models sharing a letter (e.g., 'a') are not significantly different from one another ($p > .05$). Models with different letters (e.g., 'a' vs. 'b') have significantly different performance ($p < .05$, Wilcoxon signed rank test).}
\label{tab:MN_response}
\end{table}

\begin{table}[h!]
\centering
\begin{tabular}{l l r r r r r r}
\hline
\textbf{CLD} & \textbf{Model} & \textbf{Min} & \textbf{1st Quart.} & \textbf{Median} & \textbf{Mean} & \textbf{3rd Quart.} & \textbf{Max} \\
\hline
a & Hibou-B         & -0.5449 & -0.0339 &  0.1038 &  0.1039 &  0.2381 & 0.7542 \\
b & Virchow2        & -0.5449 & -0.1340 &  0.0000 &  0.0169 &  0.1581 & 0.6325 \\
b & H-optimus-1     & -0.6172 & -0.1372 &  0.0000 &  0.0012 &  0.1372 & 0.7688 \\
b & Virchow         & -0.7006 & -0.1494 & -0.0109 & -0.0060 &  0.1340 & 0.6720 \\
c & Hibou-L         & -0.5709 & -0.1682 & -0.0369 & -0.0393 &  0.0895 & 0.5916 \\
d & UNI             & -0.6231 & -0.2186 & -0.0714 & -0.0687 &  0.0778 & 0.6706 \\
d & Prov-Gigapath   & -0.6237 & -0.2168 & -0.0714 & -0.0697 &  0.0714 & 0.5367 \\
e & UNI2-h          & -0.6667 & -0.2392 & -0.0976 & -0.0949 &  0.0610 & 0.5733 \\
e & SP85M           & -0.6905 & -0.2308 & -0.0976 & -0.0979 &  0.0369 & 0.5007 \\
e & H-optimus-0     & -0.7542 & -0.2381 & -0.1038 & -0.1047 &  0.0369 & 0.4629 \\
f & SP22M           & -0.6367 & -0.2958 & -0.1543 & -0.1511 & -0.0120 & 0.3858 \\

\hline
\end{tabular}
\caption{\textbf{Performance comparison of HFMs for eGFR decline classification in post-renal transplantation.} Table~\ref{tab:eGFR_decline} ranks the performance of 11 HFMs for eGFR decline classification in post-renal transplant cases using an attention-based multiple instance learning (ABMIL) classifier. Performance is reported as the Matthews Correlation Coefficient (MCC) statistics (minimum to maximum) across 1000 bootstraps obtained from repeated nested stratified five-fold cross-validation runs. Statistical ranking using Compact Letter Display (CLD): The 'CLD' column groups models based on their statistical similarity. Models sharing a letter (e.g., 'a') are not significantly different from one another ($p > .05$). Models with different letters (e.g., 'a' vs. 'b') have significantly different performance ($p < .05$, Wilcoxon signed rank test).}
\label{tab:eGFR_decline}
\end{table}

\begin{table}[h!]
\centering
\begin{tabular}{l l r r r r r r}
\hline
\textbf{CLD} & \textbf{Model} & \textbf{Min} & \textbf{1st Quart.} & \textbf{Median} & \textbf{Mean} & \textbf{3rd Quart.} & \textbf{Max} \\
\hline

a & Hibou-L         & -0.4802 & -0.1604 & -0.0808 & -0.0925 & -0.0133 & 0.2477 \\
b & Hibou-B         & -0.6246 & -0.2143 & -0.1234 & -0.1320 & -0.0372 & 0.2657 \\
c & Virchow2        & -1.0401 & -0.3271 & -0.1756 & -0.1888 & -0.0375 & 0.3247 \\
d & UNI2-h          & -0.9761 & -0.3342 & -0.1883 & -0.2131 & -0.0713 & 0.2603 \\
e & SP22M           & -1.0089 & -0.3750 & -0.2259 & -0.2453 & -0.0991 & 0.2664 \\
f & SP85M           & -0.9875 & -0.4216 & -0.2682 & -0.2827 & -0.1330 & 0.3867 \\
g & Virchow         & -1.1700 & -0.4501 & -0.2919 & -0.3069 & -0.1538 & 0.2748 \\
g & UNI             & -1.2127 & -0.4780 & -0.3030 & -0.3154 & -0.1377 & 0.3232 \\
h & H-optimus-1     & -1.4375 & -0.5383 & -0.3437 & -0.3651 & -0.1586 & 0.3125 \\
i & Prov-Gigapath   & -1.4614 & -0.5769 & -0.4015 & -0.4229 & -0.2485 & 0.1577 \\
j & H-optimus-0     & -1.4034 & -0.6905 & -0.4793 & -0.5059 & -0.3005 & 0.3026 \\

\hline
\end{tabular}
\caption{\textbf{Performance comparison of HFMs for one-year eGFR prediction in post-renal transplantation.} Table~\ref{tab:eGFR_prediction} ranks the performance of 11 HFMs for eGFR prediction using an attention-based multiple instance learning (ABMIL) regressor. Performance is reported as the $R^2$ statistics (minimum to maximum) across 1000 bootstraps obtained from repeated nested five-fold cross-validation runs. Statistical ranking using Compact Letter Display (CLD): The 'CLD' column groups models based on their statistical similarity. Models sharing a letter (e.g., 'a') are not significantly different from one another ($p > .05$). Models with different letters (e.g., 'a' vs. 'b') have significantly different performance ($p < .05$, Wilcoxon signed rank test).}
\label{tab:eGFR_prediction}

\end{table}

In the slide-level task of predicting treatment response in MN patients using PAS-stained WSIs, HFMs exhibited near-random performance across all HFMs with median MCCs ranging from $-$0.0625 to 0.0000 (Table ~\ref{tab:MN_response}). These results suggest that global morphological embeddings, even from large-scale HFMs, poorly capture the subtle histomorphologic features that correlate with treatment response, which are likely driven by immunologic processes invisible to standard histology. The general consistency across encoders demonstrates representational stability across HFMs.

In the slide-level prediction of eGFR decline between one and three years post-renal transplant, performed using PAS-stained WSIs, HFMs showed limited discriminative power, with median MCCs ranging from $-$0.1543 to 0.1038 (Table~\ref{tab:eGFR_decline}). These findings could suggest that histologic morphology alone, even when modeled by large-scale HFMs, is insufficient to capture the complex biological processes of graft functional decline, which involve microvascular injury and immunological changes. Nevertheless, consistent results across HFMs suggest shared sensitivity to global tissue patterns cannot capture features responsible for eGFR decline, hinting at weak morphological correlates of functional loss.
\bigskip
\subsubsection*{Regression tasks}
For slide-level prediction of one-year post–renal transplant eGFR, we employed the same ABMIL framework used for slide-level classification, modifying only the final layer by replacing the classification head with a regression head. The post-renal transplant one-year eGFR prediction task from donor kidney biopsies using PAS-stained biopsies yielded weak to no correlations. Table~\ref{tab:eGFR_prediction} shows that HFMs achieve a median $R^2$ score between $-$0.4793 and $-$0.0808. These results indicate that HFMs embeddings could not capture macroscopic correlates of renal function. 

\subsection*{Summary of Results}
Overall, tile-level evaluation results indicate that on tasks governed by clear morphological contrasts, such as glomerulosclerosis and inflammation classification, HFMs achieved moderate to high performance, indicating that HFMs capture robust structural and textural features underlying chronic kidney pathology. Evaluation tasks involving subtle morphological changes like GBM spike detection and artery stenosis classification,  HFMs performance deteriorated, achieving low to moderate performance.  In contrast, biologically complex or molecularly defined tasks, such as tubular injury classification and cell-type proportion regression, showed weak to no performance, underscoring the current limitation of vision representations in modeling molecular information. Slide-level evaluations via ABMIL further revealed that global embeddings capture diagnostic morphology (e.g., DN vs. reference) but struggle to predict functional changes such as eGFR decline in renal transplant recipients, one-year eGFR prediction post-renal transplant, or treatment-related outcomes such as MN prognosis, where complex molecular processes dominate over visible structure. The augmentation-robustness analysis confirmed that most HFMs encode stable, color- and texture-invariant representations, though noise perturbations remain a sensitive augmentation. Collectively, these findings highlight that current HFMs have achieved stain-agnostic generalization across certain renal tasks, with clear morphological changes. However, their applicability in settings that involve extracting molecular information from histology and prognosis remains a challenge.

\section{Discussion}
The HFMs have been a major advancement in the field of digital pathology, and several HFMs are being developed and made open source. Although prior benchmarking studies have focused on evaluating HFMs in cancer-specific datasets, kidney-focused evaluations remain underexplored. In this study, we systematically evaluated vision HFMs (pretrained on > 100k WSIs) to maximize exposure to diverse morphological features and assessed their generalization performance across both within and cross-stain for kidney-specific histology tasks covering CKDs. From a biological standpoint, this analysis explores whether HFMs pretrained on tumor-centric cancer data comprising RCC and UC cancer datasets may capture features that are relevant to CKD. From a representation-learning standpoint, the evaluation tells us if the DINO-based augmentations are enough to induce stain invariance or for explicit color domain alignment (stain transfer GANs) for true generalization. In this study, we evaluated 11 publicly available HFMs across 11 kidney-specific downstream tasks across various staining protocols (PAS, H\&E, PASM, IHC), scales (tile and slide-level), task categories (classification, regression, and copy detection), and outcome types (detection, diagnosis, and prognosis) to provide a comprehensive understanding of HFMs transferability to kidney-specific histology tasks.

Using the frozen HFMs embeddings and lightweight task-specific heads, we evaluated HFMs on seven tile-level and four slide-level tasks. Tile-level classification was evaluated using linear and $k$NN probing, tile-level regression using ridge regression, and slide-level classification/regression using the ABMIL framework on frozen embeddings obtained from HFMs. We developed a robust benchmarking pipeline for estimating the generalized predictive performance of these HFMs on various downstream tasks. Tile-level classification and regression were performed using repeated stratified group five-fold cross-validation and repeated group five-fold cross-validation, respectively. For slide-level tasks, we adopted a repeated nested stratified five-fold cross-validation framework for classification and repeated nested five-fold cross-validation for regression to better estimate generalization performance. These evaluation frameworks help in reducing variance by sampling across random seeds, preserving class distribution by stratified splitting, and preventing optimistic bias by finding the optimal set of hyperparameters within the training folds. To assess statistical significance, we performed post-hoc statistical analysis, including the Friedman test, followed by the pairwise Wilcoxon signed-rank test with Holm-Bonferroni correction, and visualization with compact letter display. We see stratified performance of HFMs across various renal evaluation tasks, giving insights into biologically shared features between the CKD and RCC and UC in the training datasets used for HFMs pretraining and underscoring the representational capacity of embeddings obtained from HFMs.

For evaluation tasks involving coarse and well-defined visible meso-scale features on kidney CKD WSIs involving glomerulosclerosis, inflammation, and DN, HFMs achieved moderate to high performance. These findings are consistent with the biological possibility that HFMs pretrained primarily on large cancer-oriented datasets can learn meso-scale visual morphological features (glomerular tuft collapse, immune cell clustering, GBM thickening, etc.) that are shared in both malignant and non-malignant kidney pathologies. These renal morphological features are generic enough to transfer to CKD phenotypes, suggesting that this cancer-derived pretraining captures these morphological features in WSIs of CKD patients. From a representation learning standpoint, these results could suggest that representations obtained from SSL pretraining remain robust to a certain degree of stain variation while still preserving key structural features.

In contrast, tasks involving subtle morphological features, such as GBM spike formation and artery stenosis classification, HFMs show deterioration in performance relative to performance on meso-scale features. Based on the biological hypothesis, this could mean that fine microstructural features may be weakly presented in the pretraining data. From a representational learning standpoint, these results are compatible with the augmentation-driven explanation, where these tiny structural features may not be preserved under augmentations involving color jitter and gaussian blur. These findings can suggest a need for kidney-specific pretraining, where you can preserve these microstructural features under various augmentations rather than relying on general-purpose augmentations.

The performance of these HFMs further deteriorated for biologically complex tasks such as tubular injury assessment, cell-type estimation, and prognostic slide-level tasks. From the biological transfer hypothesis, the limitation could be that even if the pretraining data comprises kidney morphology, these targets may depend on mechanisms that are not strongly expressed in static visual patterns in WSI, or the required context may not be captured in WSI alone. Performance deterioration could also suggest that morphological changes did not encode the transcriptomic-derived information that is not visually grounded in tubular injury assessment and cell-type estimation tasks. From a representation learning standpoint, the stain invariance alone is insufficient if the limiting factor is not appearance shift but rather signal availability (limited cohort size) and modeling complexity (weakly supervised learning (ABMIL)) for prognostic slide-level tasks. The lack of generalization of slide-level ABMIL training to predict functional decline or prognosis could suggest that the embeddings capture static structural features and not the dynamic molecular and longitudinal processes that can guide disease progression and treatment response in kidney histology. 

The copy detection task results provide an additional generic evaluation for the representation robustness of frozen HFM embeddings under various DINO-based augmentations. The results suggest invariance of frozen HFM embeddings to color, deformation, and geometric perturbations. However, sensitivity to noise perturbations suggests that robustness may depend on noise perturbation magnitude. Though the copy detection task is not directly related to kidney disease, this analysis serves as a stress test of the HFMs embeddings obtained under various real-world conditions with unavoidable artifacts. 

Overall, our findings closely align with prior work by Kurata \textit{et al.}~\cite{kurata2025multiple}, who demonstrated that HFMs perform well on diagnostic kidney tasks (healthy control, acute interstitial nephritis, and diabetic kidney disease) using H\&E-stained WSIs. We extended this study across a broad range of HFMs, staining variations (H\&E, PAS, PASM, and IHC), scales (tile and slide-level), task categories (classification, regression, and copy detection), and outcome types (detection, diagnosis, and prognosis). Additionally, the proposed evaluation framework ensures reproducibility and systematic methodology through repeated nested cross-validation. The evaluation pipeline is made available as an open-source Python package publicly available at \url{https://pypi.org/project/kidney-hfm-eval/} and archived on Zenodo with a DOI~\cite{harishwar_reddy_kasireddy_2026_18882125}.

In the slide-level evaluations performed in this study, the dataset size remained limited, mainly due to the difficulty of assembling multi-center and well-annotated cohorts, particularly for follow-up-dependent prognostic outcomes. Our evaluation framework was restricted to vision HFMs pretrained with large training cohorts (>100k WSIs) to ensure exposure to a broad spectrum of histological features, especially RCC and UC. However, expanding future evaluations to include vision-language models~\cite{lu2024visual, huang2023visual, ding2025multimodal}, multi-modal HFMs~\cite{liu2025spemo} that integrate complementary signals such as clinical text and omics profiles can help in providing richer contextual and biological information that is not fully captured by morphology alone, potentially enabling the models to learn more discriminative and biologically meaningful representations of WSIs. In parallel, developing domain-specialized, kidney-focused, or organ-specific HFMs may better capture tissue-specific structures and disease patterns that are underrepresented in general FMs, which might reveal different generalization behavior and performance across diverse kidney-related tasks. Future work should explore a broader range of modalities and kidney-specific pretraining strategies. Extending this framework to chronic cardiovascular diseases and other organ systems may further test the limits of current HFMs and also help guide the training of better HFMs for improved downstream task performance. Moreover, integrating these HFMs with explainable pathomic features can enhance their explainability and potential utility in clinical settings~\cite{kapse2024si,kasireddy2025explainable}. 

In summary, we conducted a comprehensive benchmarking to date of vision HFMs on kidney-specific histopathology tasks and found that the HFMs performance deteriorated as the complexity of the task increased, independent of stain type. From a clinical translational perspective, our results indicate that the current version of HFMs are not yet ready to be deployed in pharmaceutical and clinical decision-making settings that require reliable prediction of treatment response, disease progression, biomarker discovery, or therapy outcome prediction in nephrology. Our results show that HFMs provide a good starting point for coarse diagnostic screening tasks in kidney histopathology. However, our results also suggest that HFMs underperform in detecting fine-grained microstructural features, complex biological phenotypes, and prognostic inference. Addressing these limitations may require the development of organ-specific, multi-stain, and multi-modal FMs that integrate molecular, clinical, and longitudinal information.

\bibliography{sample}
\nolinenumbers
\appendix
\renewcommand{\figurename}{Supplementary Figure}
\renewcommand{\tablename}{Supplementary Table}

\setcounter{figure}{0}
\setcounter{table}{0}
\section{Supplementary}
\subsection{Ground truth generation for tubule data}
\label{sup:sec:tubule_gt} 

\begin{figure}[H]
\centering
    \centering
    \includegraphics[width=\linewidth]{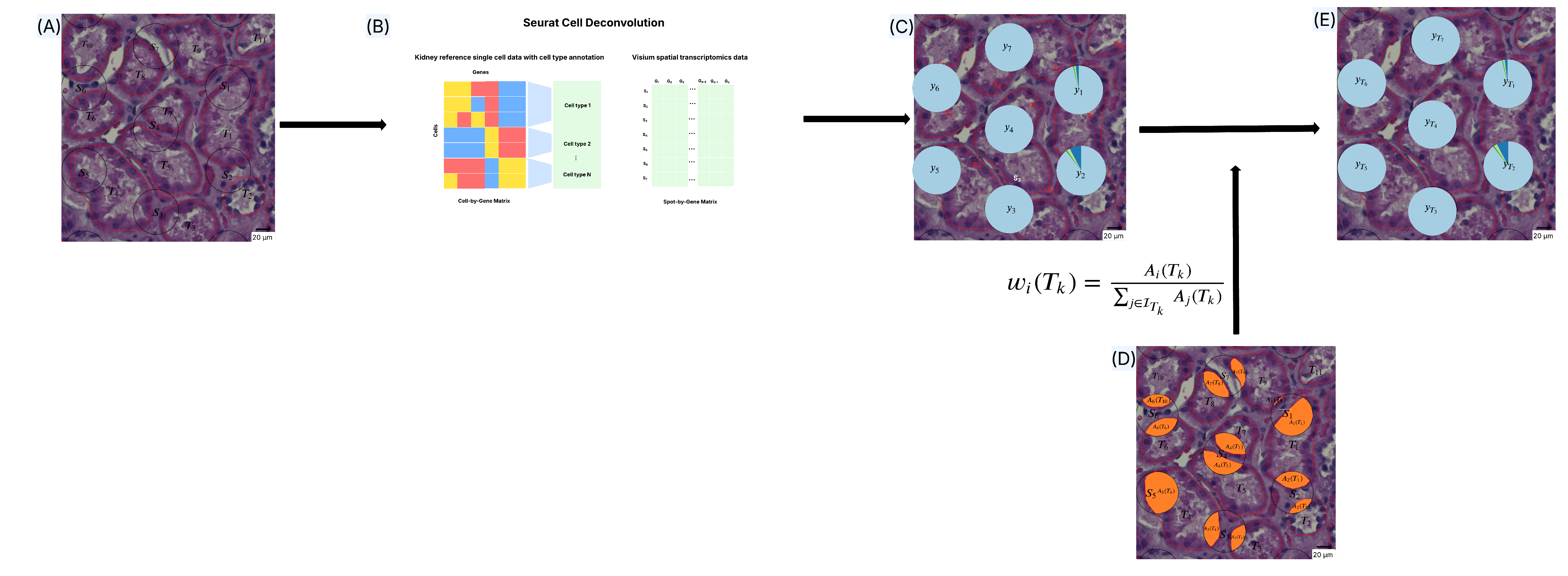}
\caption{
Pipeline for generating ground truth labels for tubules using spatial transcriptomics and histology. 
\textbf{(A)} H\&E-stained histology image patch with segmented tubules ($T_i$) and Visium spatial transcriptomics spot locations ($S_j$). 
\textbf{(B)} Seurat-based cell deconvolution workflow: kidney reference single-cell RNA-seq data (cell-by-gene matrix with cell-type annotations) is integrated with Visium spot-level gene expression (spot-by-gene matrix) to estimate cell-type proportions per spatial spot. 
\textbf{(C)} Deconvolved Visium spots overlaid on the histology image, each showing estimated proportions of multiple cell types. 
\textbf{(D)} Assignment of each spatial spot to its nearest tubule segment  transferring spot-specific cell-type proportions to the corresponding tubule. 
\textbf{(E)} Final tubule-level ground truth labels, obtained by taking weighted ($w_i$) combination of $y_i$ where each tubule ($T_i$) is associated with its aggregated cell-type proportion vector ($\mathbf{y}_{T_k}$) for downstream task.
}
    \label{sup:fig:Tubule_GT_generation}
\end{figure}

In this section, we discuss in detail the ground truth generation process for the tubular classification task. Let the histology image occupy a spatial domain $\Omega \subset \mathbb{R}^2$. Each Visium ST sample is comprised of $N$ spatial spots, where each spot $i$ has a geometric support $S_i \subset \Omega$ and an associated cell-type proportion vector $\mathbf{y}_i \in \mathbb{R}^{16}$ comprising 16 main cell types, obtained using Seurat's cell deconvolution algorithm~\cite{stuart2019comprehensive}. Renal tubules segmented from histology were indexed by $k = 1,\dots,K$, where $K$ is the total number of tubules, each represented by a mask $T_k \subset \Omega$. For each tubule $T_k$, we identified all Visium spots intersecting the tubule and quantified their contribution using the overlap area $A_i(T_k) = |S_i \cap T_k|$. The set of overlapping spots is denoted by $\mathcal{I}_{T_k} = \{\, i : A_i(T_k) > 0 \,\}$. Each spot was assigned a weight, $w_i(T_k) = \frac{A_i(T_k)}{\sum_{j \in \mathcal{I}_{T_k}} A_{j}(T_k)}$ and the tubule-level cell-type composition was obtained as the weighted average of the spot-level proportion vectors, $\mathbf{y}_{T_k} =  \sum_{i \in \mathcal{I}_{T_k}} w_{i}(T_k) \mathbf{y}_i$. To assign a tubule-level label, we restricted the analysis to epithelial cell types and identified the epithelial class with the highest aggregated proportion in $\mathbf{y}_{T_k}$. We then compared the aggregated proportions of reference-state and altered-state cells within this majority epithelial class.

\subsection{Augmentations used for copy detection task}
\label{sup:sec:aug_copy_detection_task} 
To evaluate the robustness of HFM embeddings in copy-detection scenarios, we apply four different augmentations that reflect common sources of variability in real-world WSI acquisition and digitization. Augmentations are implemented using OpenCV and custom routines, applied independently per patch.

\begin{enumerate}
  \item \textbf{Geometric Transformations:}  
    Histopathology slides are rarely perfectly aligned, subtle rotations, translations, and scaling occur due to tissue placement or scanning conditions. To simulate slight misalignments from tissue placement or scanning, we use random 2D similarity transformations with \texttt{shift\_range}=10\%,\texttt{scale\_range}=10\%, \texttt{rotate\_range}=30$^\circ$
    implemented via \texttt{cv2.getRotationMatrix2D} and \texttt{cv2.warpAffine}. These modifications evaluate whether embeddings remain stable under realistic geometric perturbations.

  \item \textbf{Color Transformations (Brightness / Contrast / Gamma):}  
    Staining intensity and imaging parameters can vary widely across labs and scanners. To replicate variability in staining intensity and scanner settings, we applied following, \texttt{brightness\_limit}=0.30, \texttt{contrast\_limit}=0.30,
    \texttt{gamma\_limit}=(0.80,\,1.20), verifying that embeddings handle a wide range of color and illumination changes.
  
  \item \textbf{Noise and JPEG Artifacts:}  
    Different scanners introduce varying noise patterns, and many slides are stored or transmitted in compressed formats. To mirror scanner noise and compression, we inject Gaussian noise via our custom \texttt{var\_limit} = [1,\,25] testing embedding tolerance to both sensor noise and compression-induced distortions.
  
  \item \textbf{Elastic Deformations:}  
    During tissue mounting or handling, microscopic distortions can occur. To mimic subtle tissue distortions during mounting or handling, we apply     \texttt{sigma}=12, \texttt{alpha}=20), introducing mild, localized elastic deformations and ensuring embeddings remain robust under small non‐rigid perturbations.
\end{enumerate}

These augmentations target the dominant sources of appearance variability encountered in clinical scanning workflows, thereby evaluating whether foundation models generate embeddings that are resilient to specific transformations.

\begin{table}[h!]
\centering
\small
\begin{tabular}{lll}
\hline
\textbf{Category} & \textbf{Hyperparameter} & \textbf{Value(s)} \\
\hline
\multirow{3}{*}{Model / Architecture} 
& Instance embedding dimension ($M$) & $\{256,\;512,\;1024\}$ \\ \cline{2-3}
& Attention hidden dimension ($L$) & $\{32,\;128,\;256\}$ \\ \cline{2-3}
& Dropout & $0.6$ (fixed) \\ \hline

\multirow{5}{*}{Optimization} 
& Optimizer & AdamW \\ \cline{2-3}
& Learning rate ($\eta$) & $\{5e-5,\;1e-4,\;2e-4\}$ \\ \cline{2-3}
& Weight decay & $1e-2$ (fixed) \\ \cline{2-3}
& Betas & $(0.95,\;0.99)$ \\ \cline{2-3}
& Epsilon & $1e-4$ \\ \hline

\multirow{3}{*}{Training / Selection} 
& Loss function & Cross-entropy \\ \cline{2-3}
& Epochs & up to $50$ \\ \cline{2-3}
& Early stopping patience & $10$ epochs (based on validation MCC) \\ \hline

\multirow{2}{*}{Cross-validation} 
& Outer CV & 5-fold, repeated across 3 seeds \\ \cline{2-3}
& Inner CV (tuning) & 4-fold StratifiedKFold \\ \hline

\end{tabular}
\caption{Hyperparameters used for repeated nested stratified ABMIL training for slide-level evaluation.}
\label{sup:tab:mil_hyperparams}

\end{table}

\begin{figure*}[!htbp]
\centering

\begin{subfigure}[t]{0.25\linewidth}
    \centering
    \includegraphics[width=\linewidth]{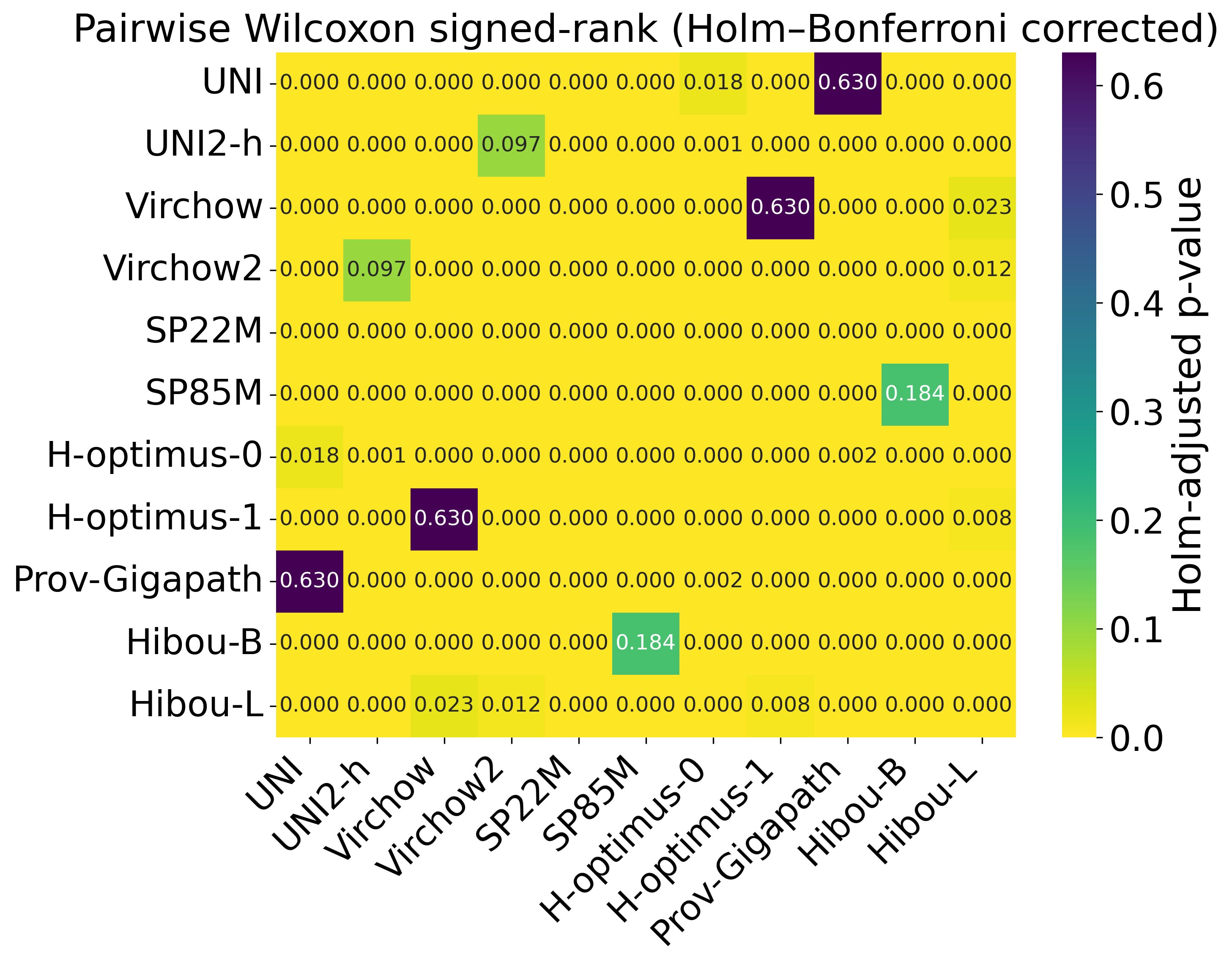}
    \caption{}
    \label{fig:lr_sclerotic}
\end{subfigure}
\begin{subfigure}[t]{0.25\linewidth}
    \centering
    \includegraphics[width=\linewidth]{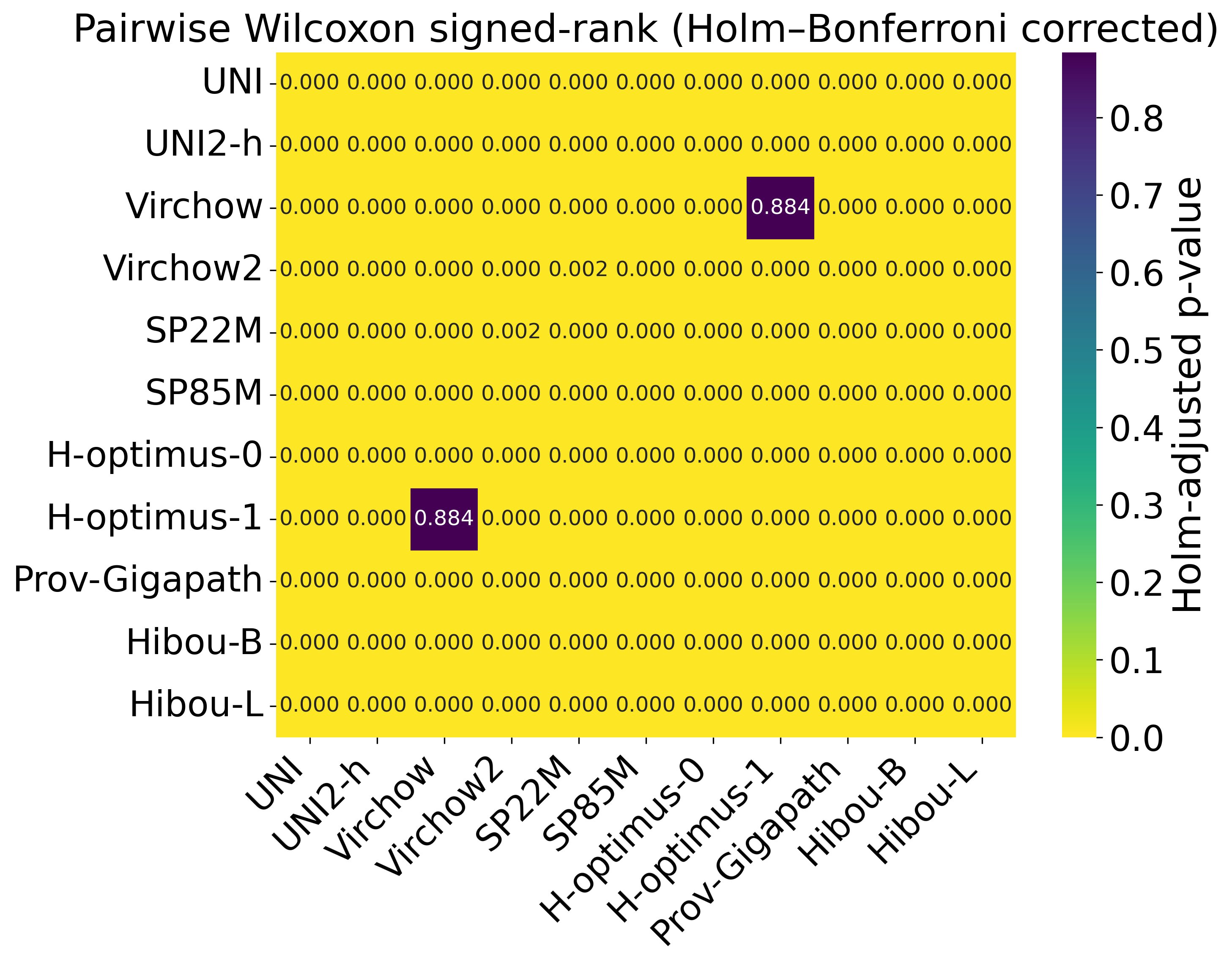}
    \caption{}
    \label{fig:kNN_sclerotic}
\end{subfigure}
\begin{subfigure}[t]{0.25\linewidth}
    \centering
    \includegraphics[width=\linewidth]{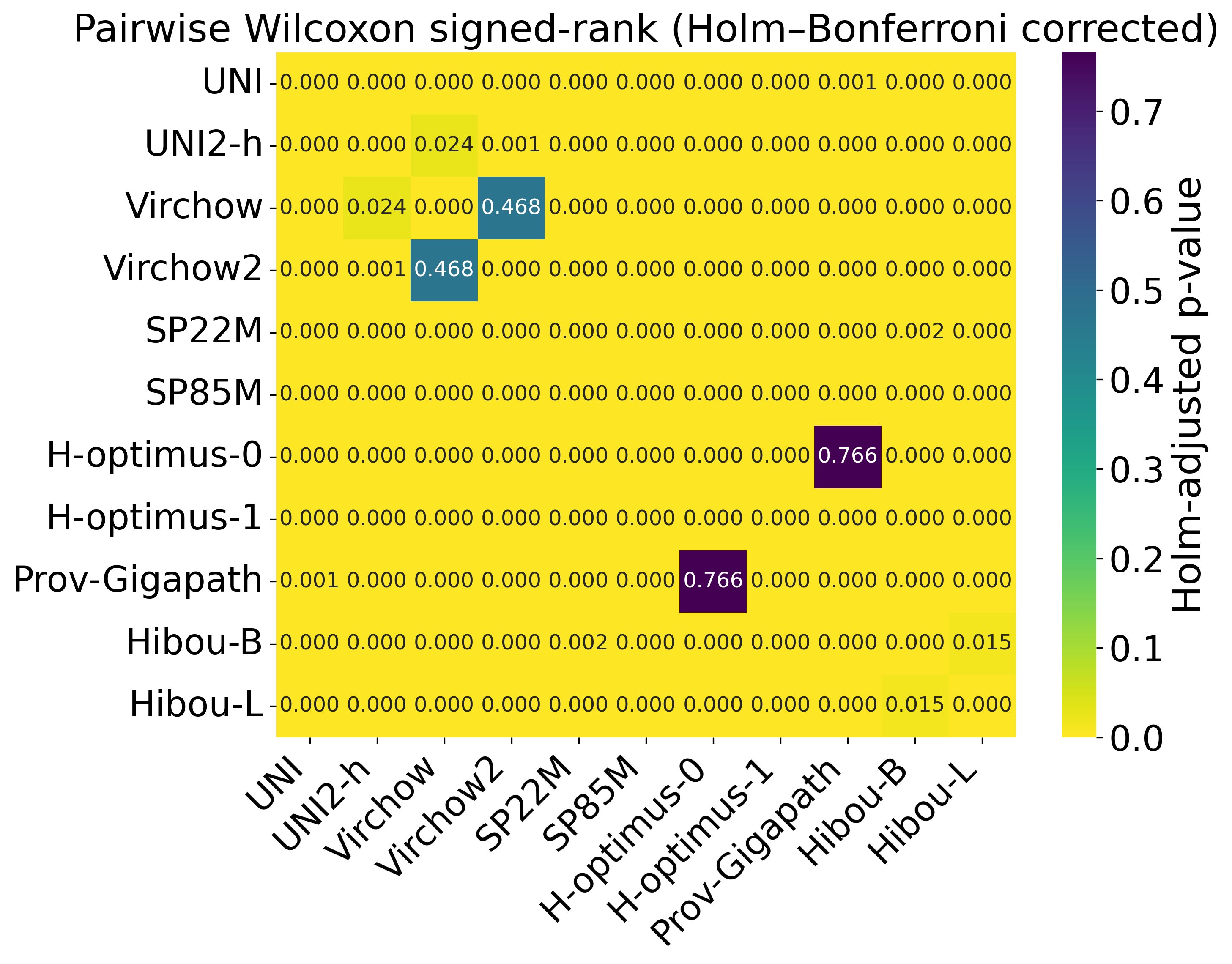}
    \caption{}
    \label{fig:lr_spike}
\end{subfigure}

\par\medskip

\begin{subfigure}[t]{0.25\linewidth}
    \centering
    \includegraphics[width=\linewidth]{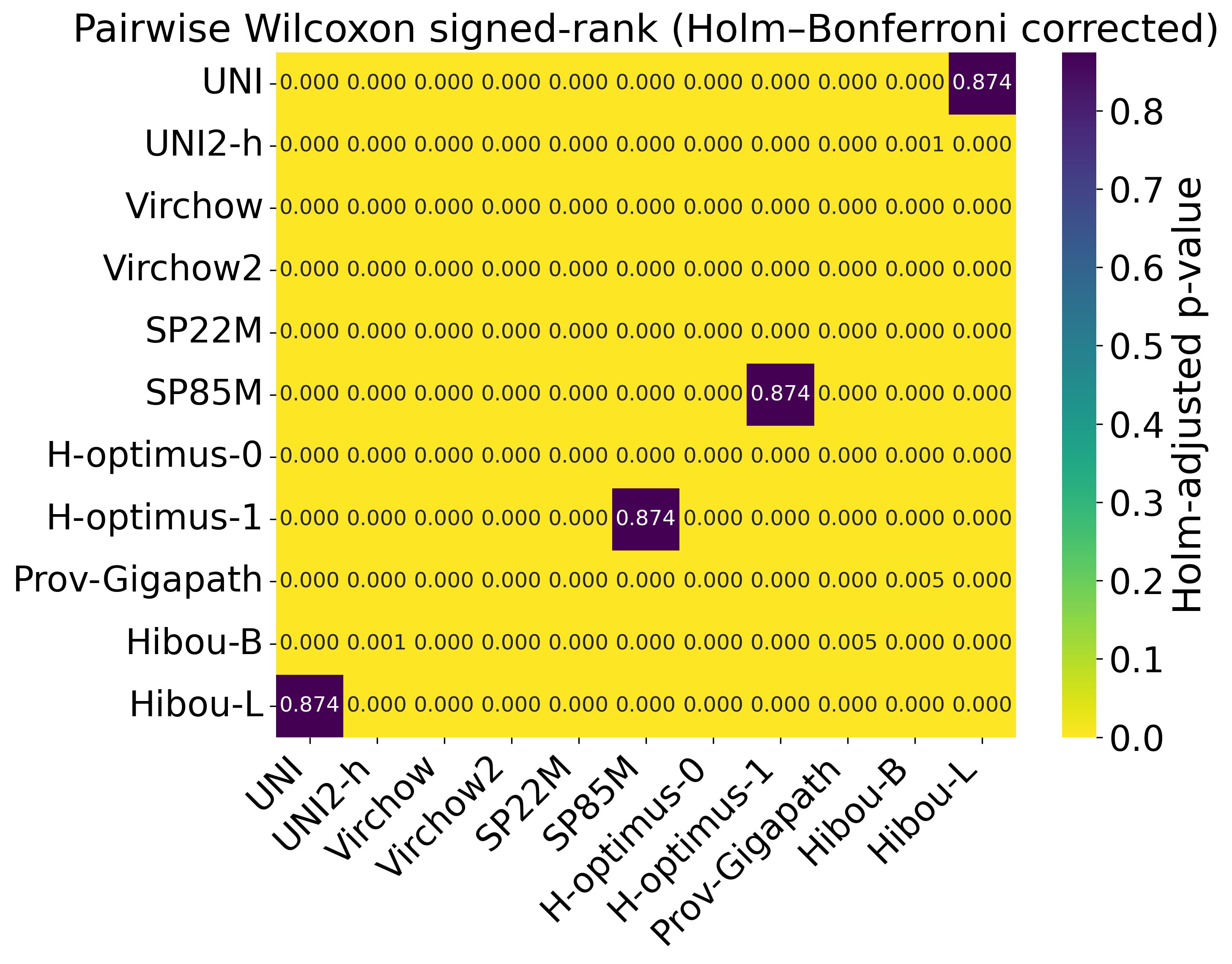}
    \caption{}
    \label{fig:kNN_spike}
\end{subfigure}
\begin{subfigure}[t]{0.25\linewidth}
    \centering
    \includegraphics[width=\linewidth]{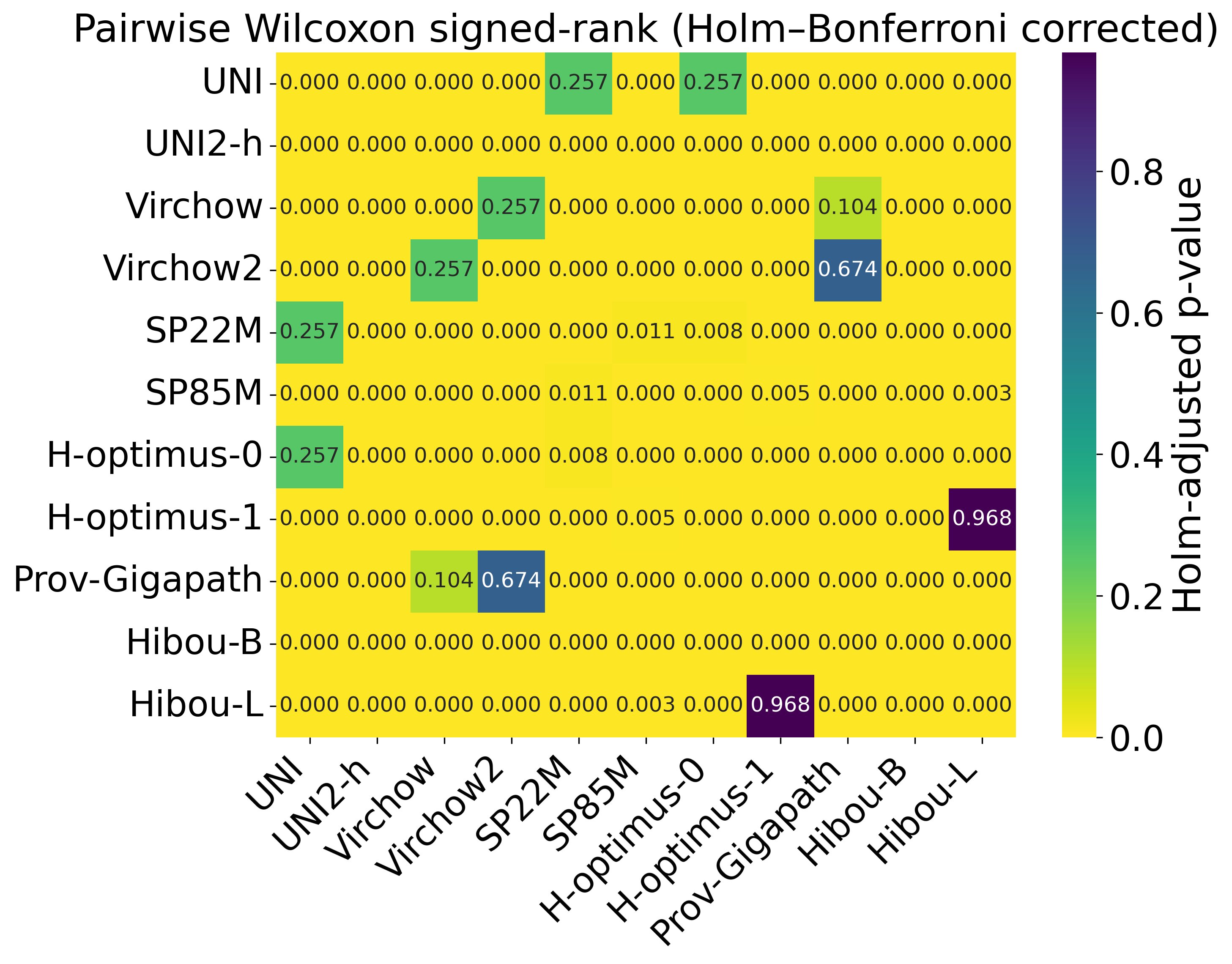}
    \caption{}
    \label{fig:lr_tubule}
\end{subfigure}
\begin{subfigure}[t]{0.25\linewidth}
    \centering
    \includegraphics[width=\linewidth]{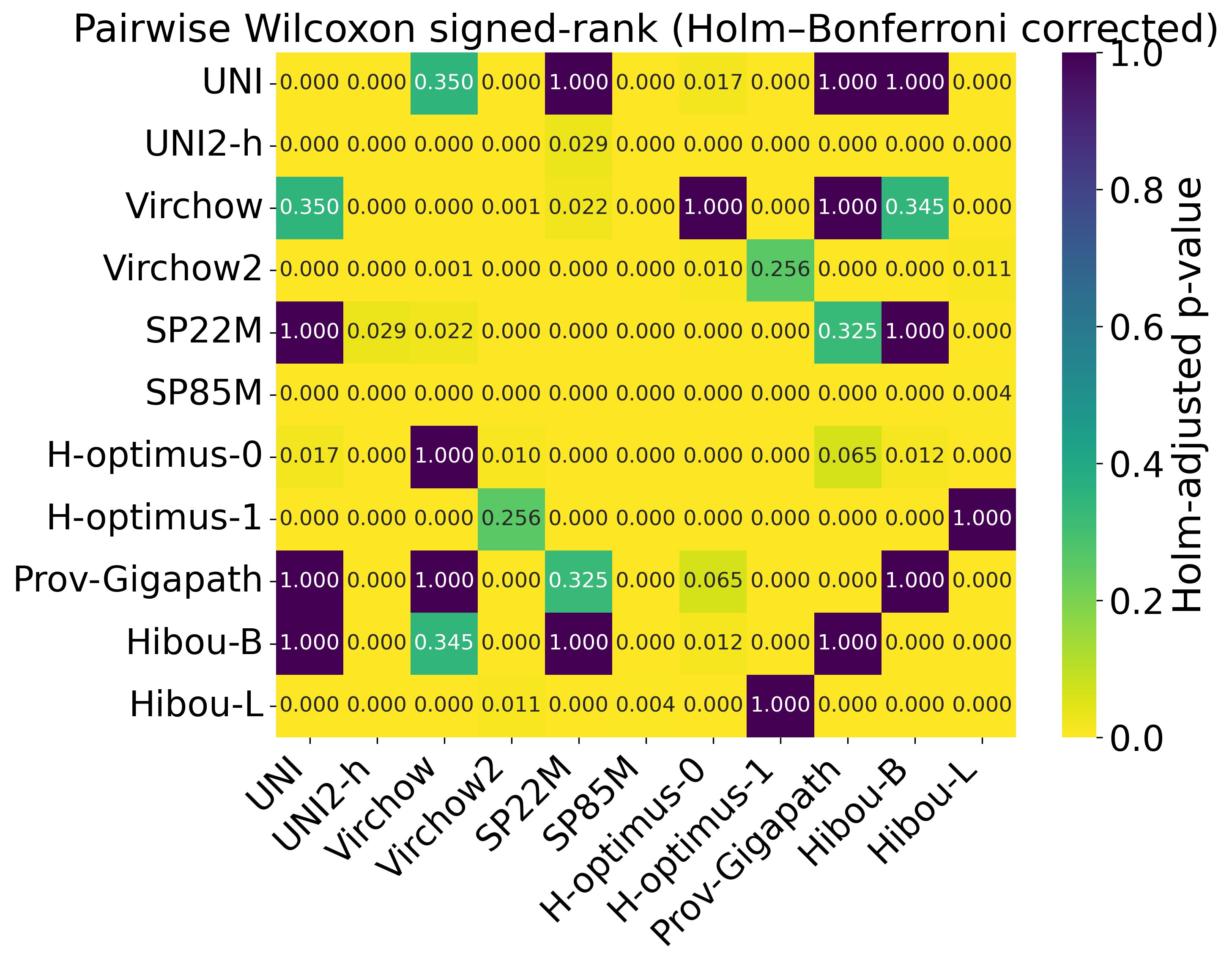}
    \caption{}
    \label{fig:kNN_tubule}
\end{subfigure}

\par\medskip

\begin{subfigure}[t]{0.25\linewidth}
    \centering
    \includegraphics[width=\linewidth]{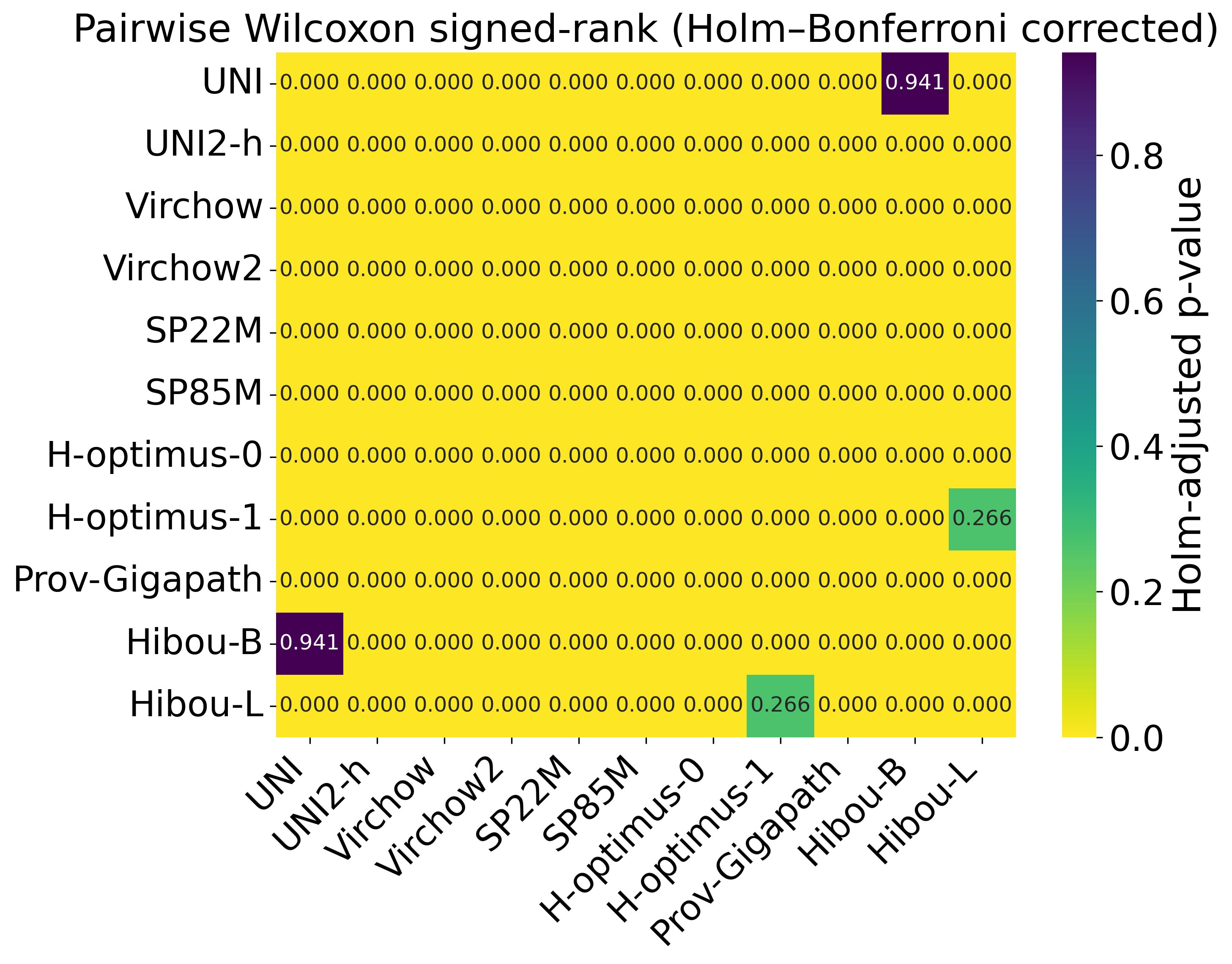}
    \caption{}
    \label{fig:lr_inflammation}
\end{subfigure}
\begin{subfigure}[t]{0.25\linewidth}
    \centering
    \includegraphics[width=\linewidth]{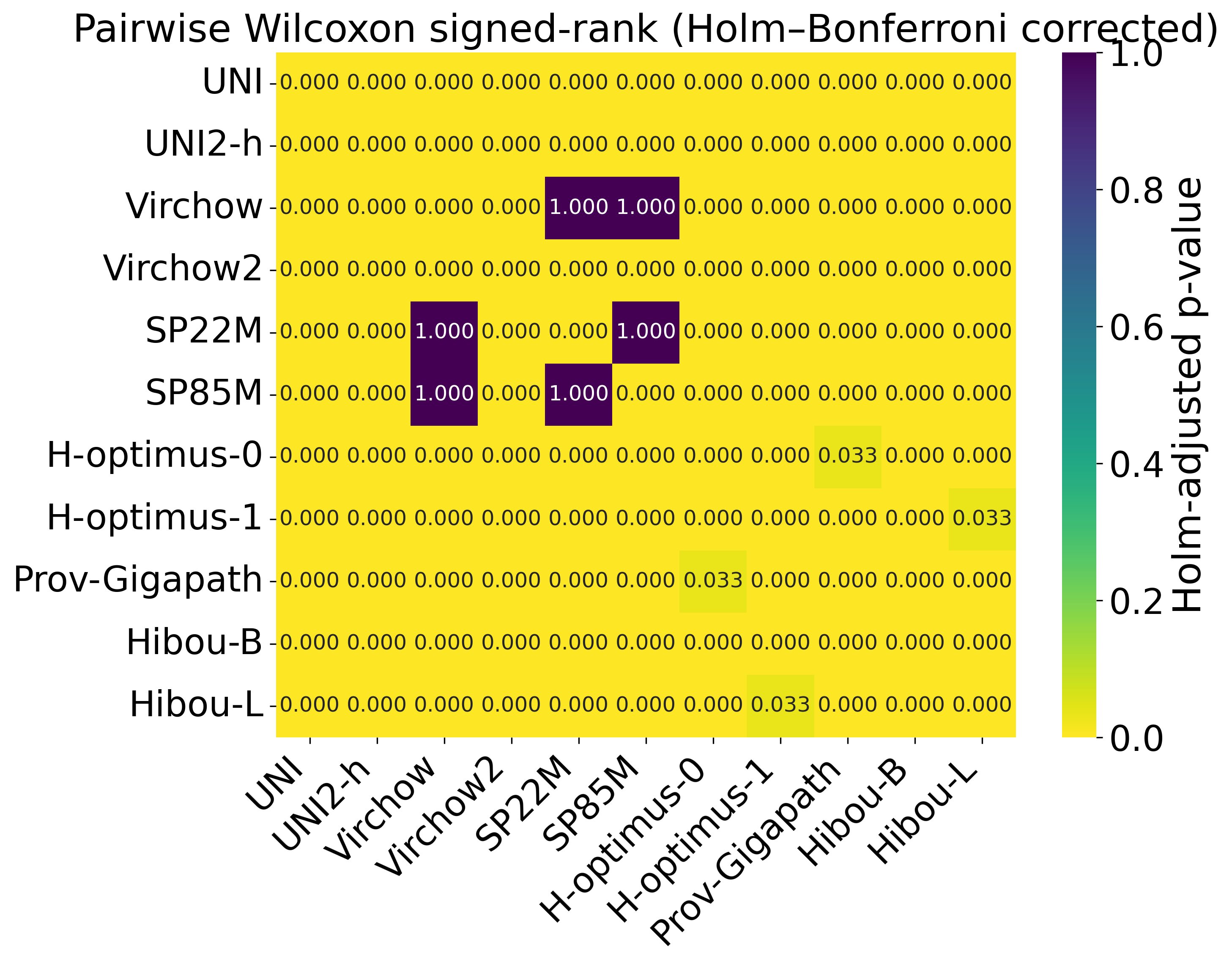}
    \caption{}
    \label{fig:kNN_inflammation}
\end{subfigure}
\begin{subfigure}[t]{0.25\linewidth}
    \centering
    \includegraphics[width=\linewidth]{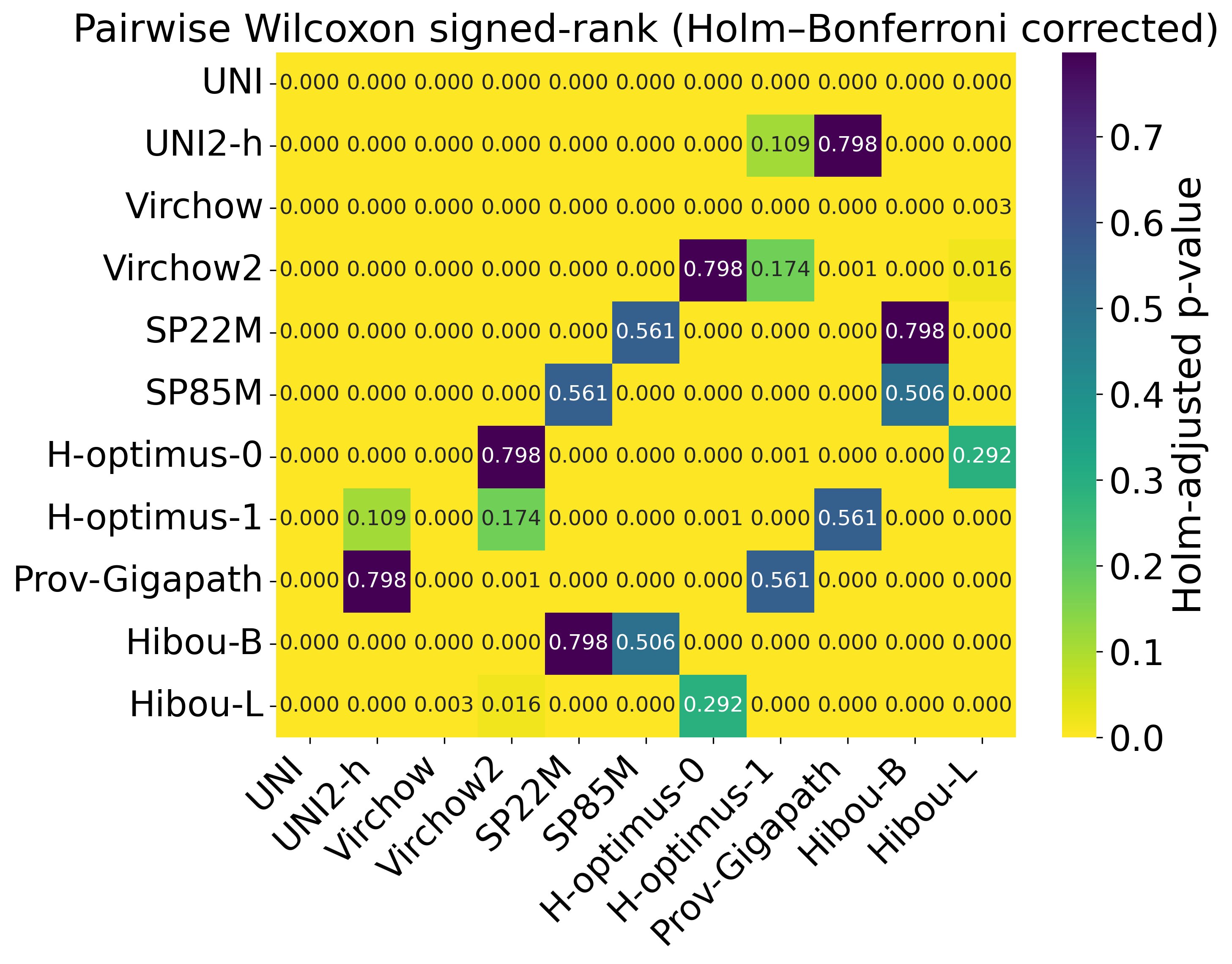}
    \caption{}
    \label{fig:lr_artery}
\end{subfigure}

\par\medskip

\begin{subfigure}[t]{0.25\linewidth}
    \centering
    \includegraphics[width=\linewidth]{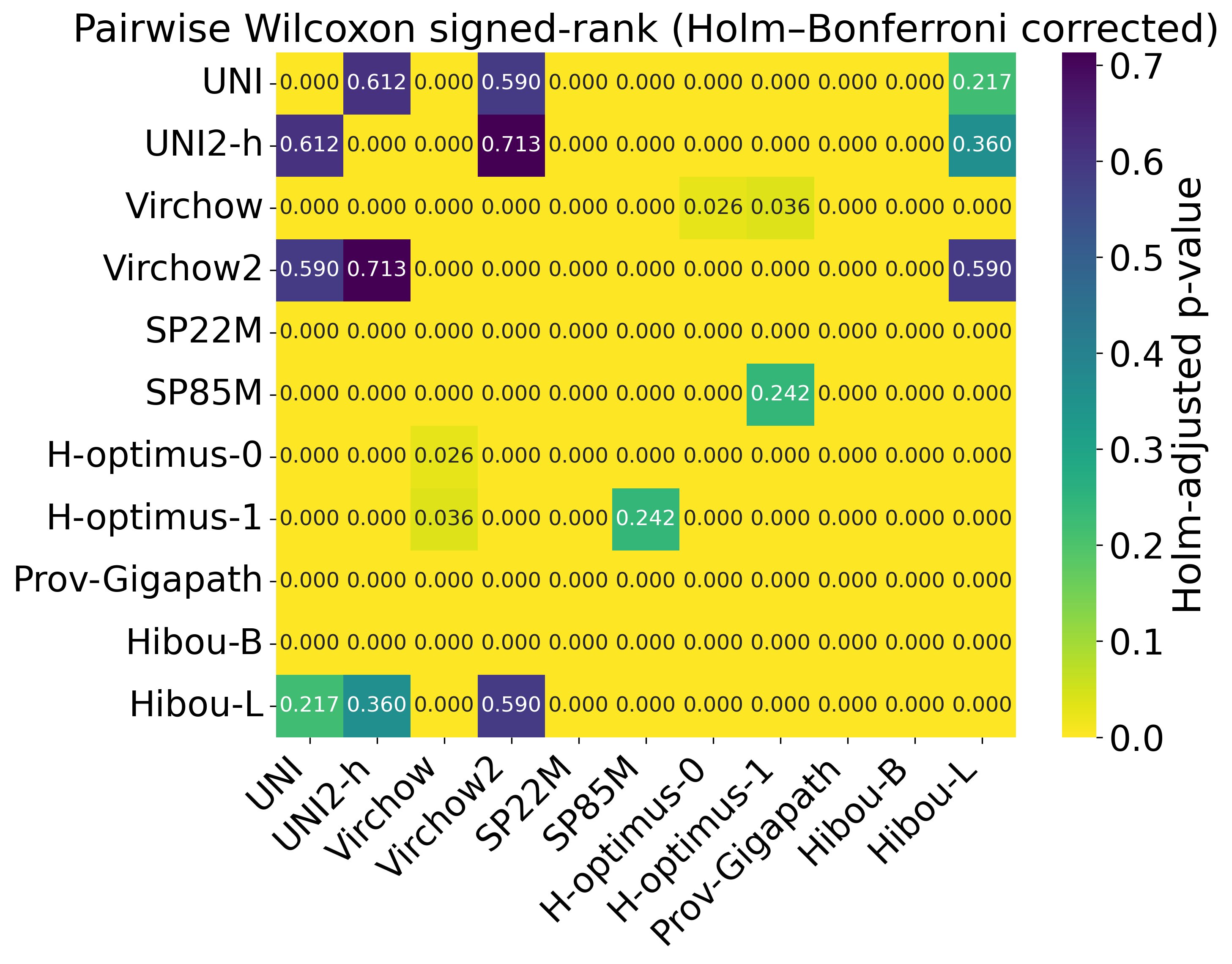}
    \caption{}
    \label{fig:kNN_artery}
\end{subfigure}
\begin{subfigure}[t]{0.25\linewidth}
    \centering
    \includegraphics[width=\linewidth]{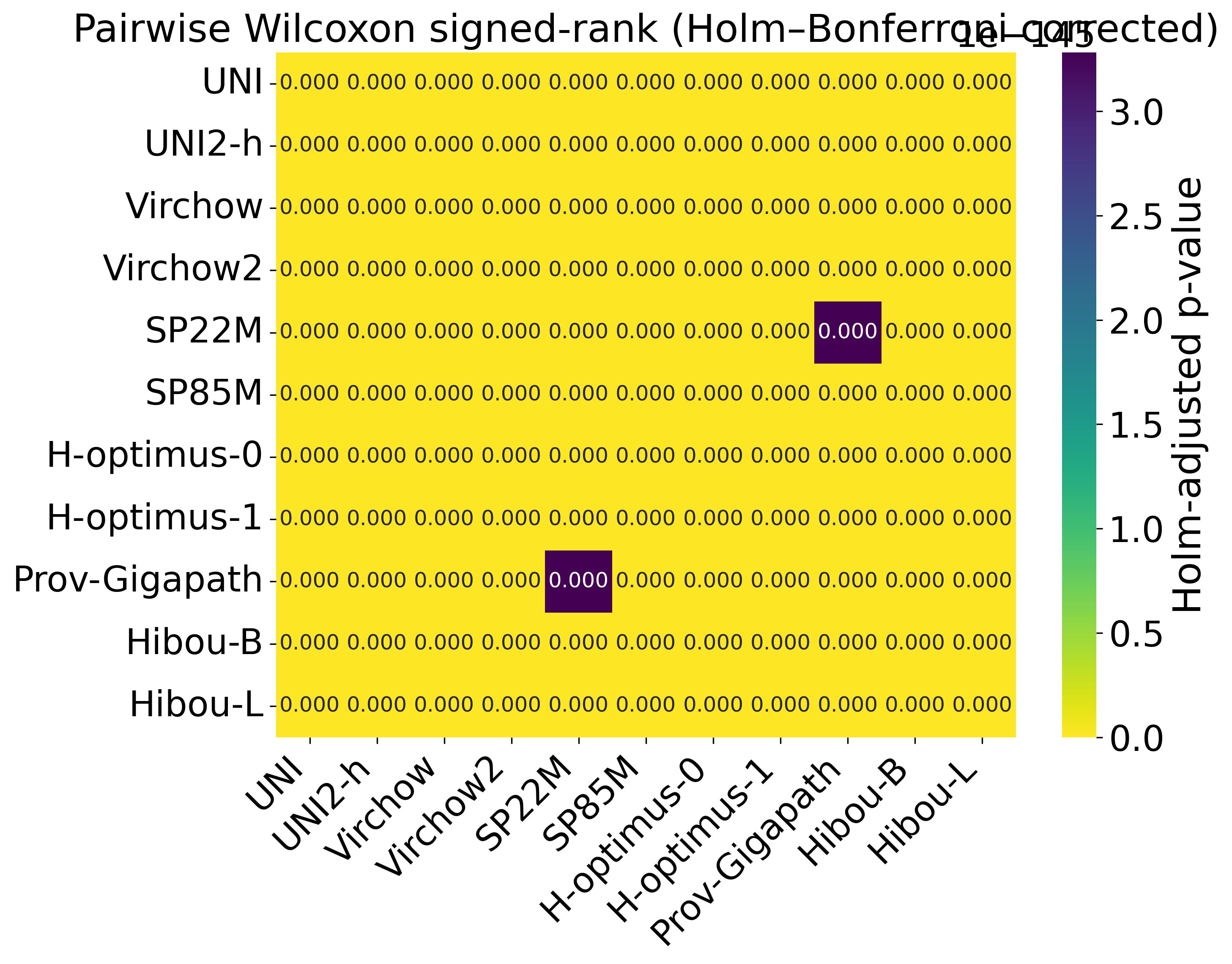}
    \caption{}
    \label{fig:kNN_artery}
\end{subfigure}
\begin{subfigure}[t]{0.25\linewidth}
    \centering
    \includegraphics[width=\linewidth]{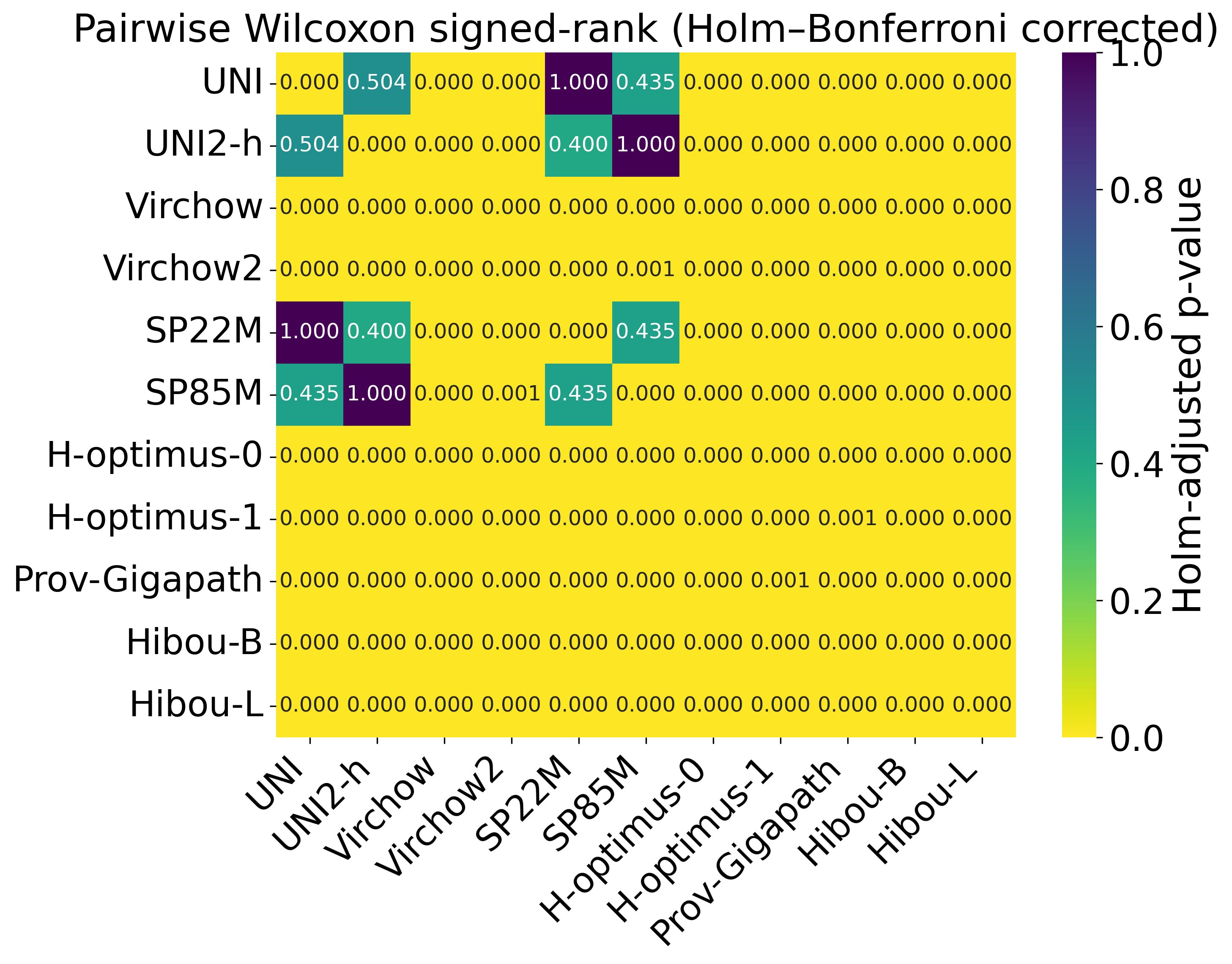}
    \caption{}
    \label{fig:MIL_DN}
\end{subfigure}

\par\medskip

\begin{subfigure}[t]{0.25\linewidth}
    \centering
    \includegraphics[width=\linewidth]{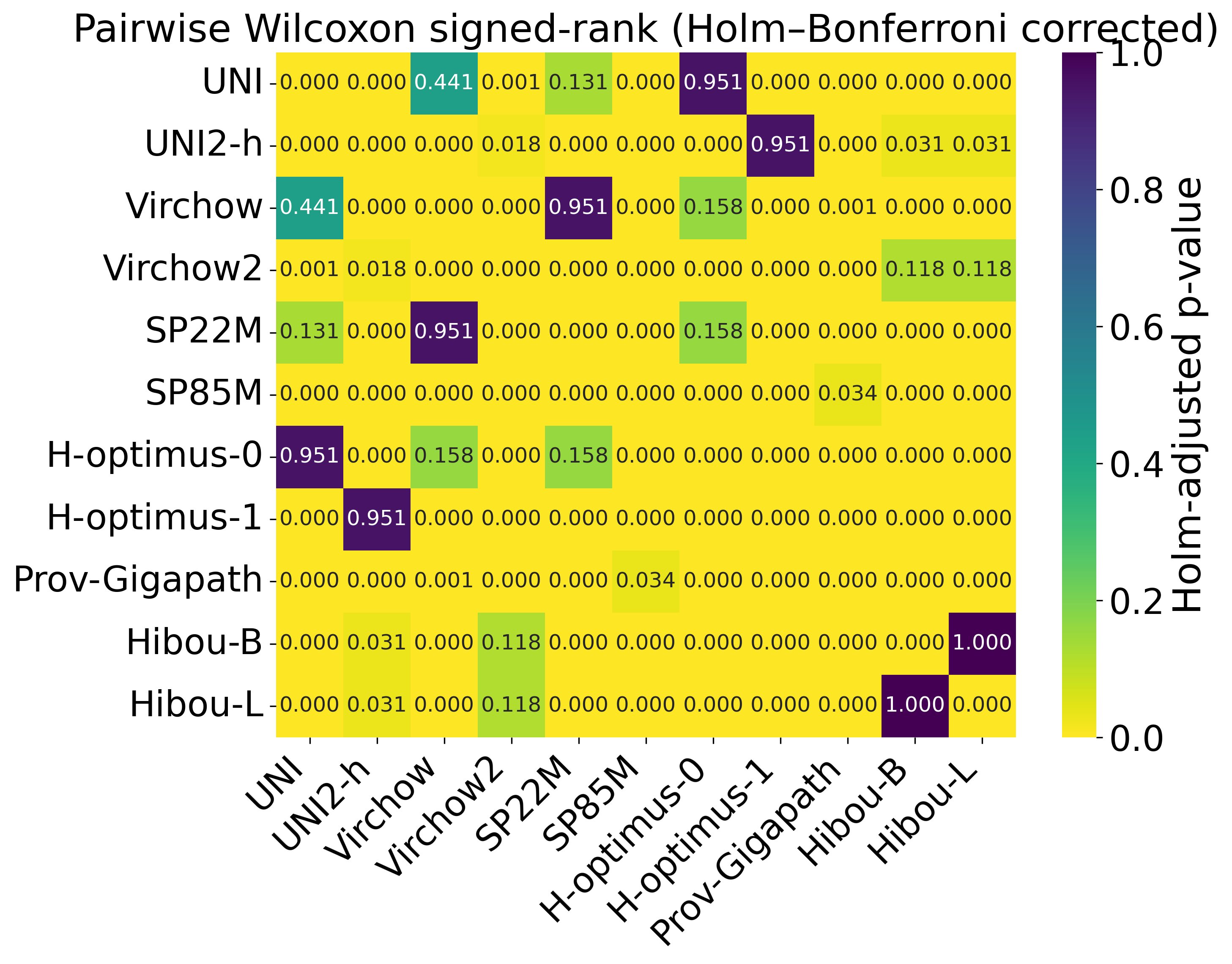}
    \caption{}
    \label{fig:MIL_MN}
\end{subfigure}
\begin{subfigure}[t]{0.25\linewidth}
    \centering
    \includegraphics[width=\linewidth]{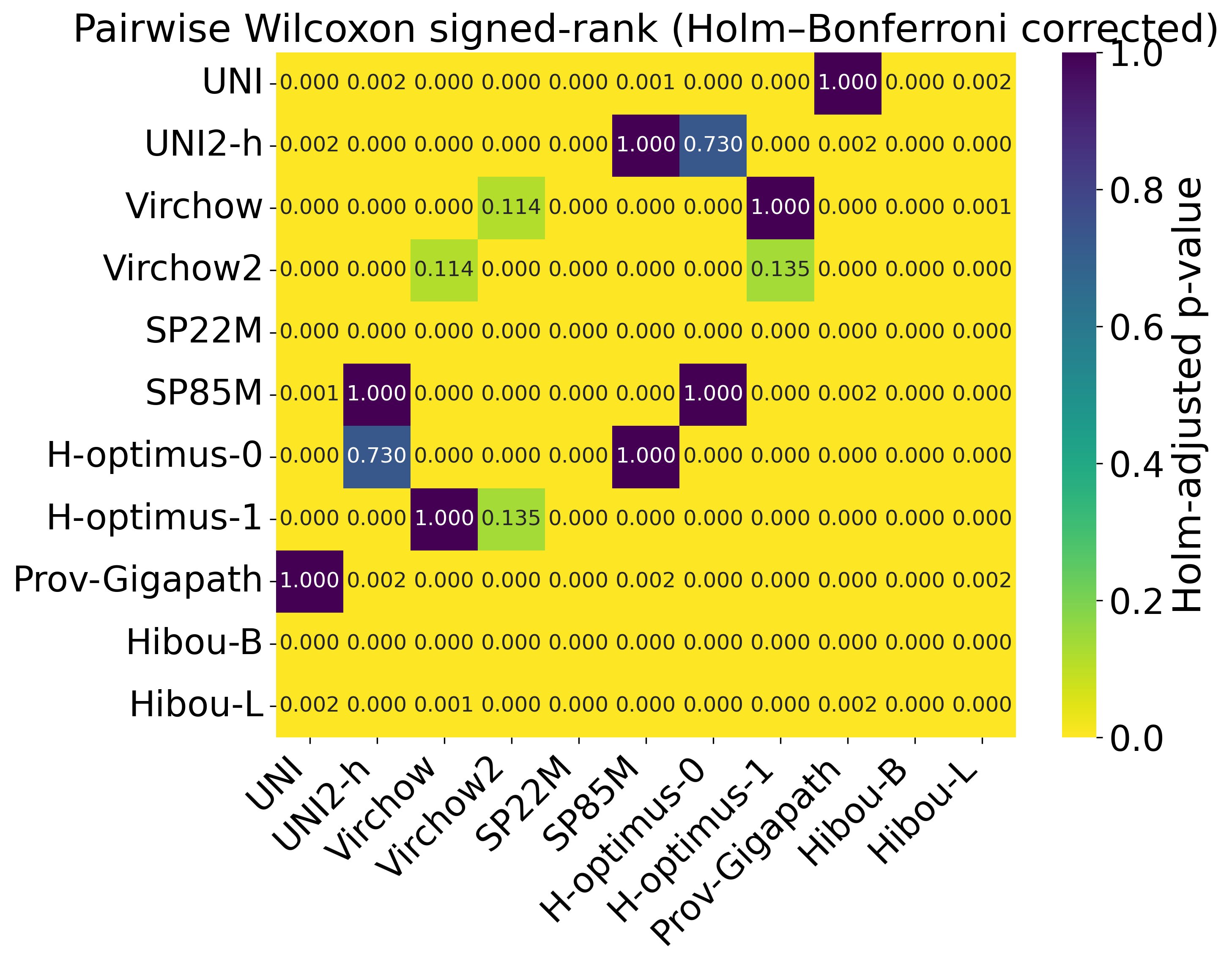}
    \caption{}
    \label{fig:MIL_Tx}
\end{subfigure}
\begin{subfigure}[t]{0.25\linewidth}
    \centering
    \includegraphics[width=\linewidth]{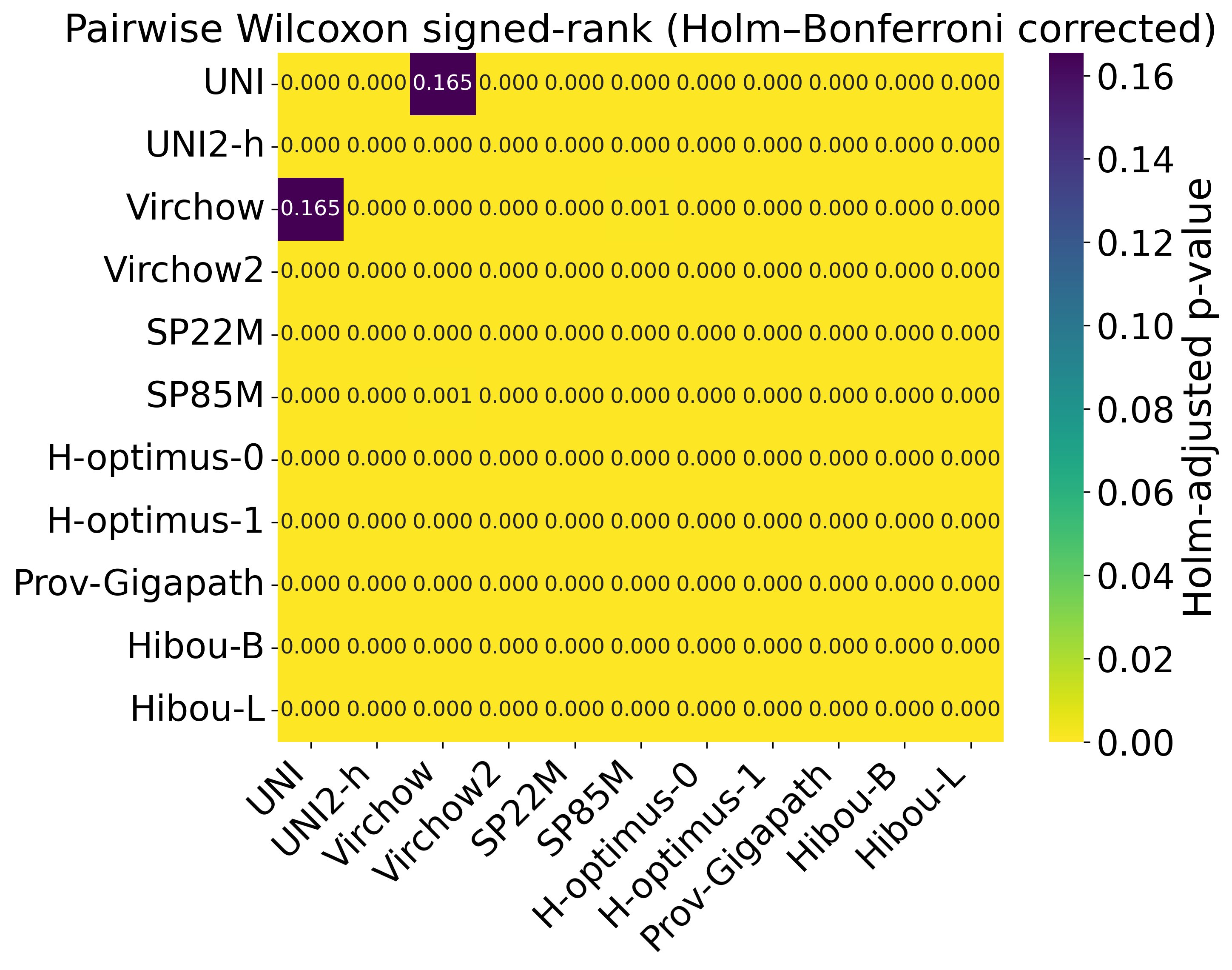}
    \caption{}
    \label{fig:MIL_DN}
\end{subfigure}

\caption{Heatmap plots showing the \textit{p}-values obtained from comparing the Matthews correlation coefficient (MCC) values across HFMs for multiple downstream tasks. Each subplot represents a specific task and probing method: 
(a)–(b) globally vs.\ non-globally sclerotic glomeruli classification using logistic regression (LR) and $k$NN, 
(c)–(d) GBM spike vs.\ no-GBM spike classification (LR, $k$NN), 
(e)–(f) normal vs.\ abnormal tubule classification (LR, $k$NN), 
(g)–(h) inflammation vs.\ non-inflammation classification (LR, $k$NN), 
(i)–(j) arteriolar stenosis multi-class classification (LR, $k$NN), 
(k) cell type estimation (ridge regression)
(i) diabetic nephropathy vs.\ control classification using MIL. 
(m) membranous nephropathy treatment-response prediction using multiple-instance learning (MIL), 
(n) renal transplant eGFR decline classification using MIL,
(o) renal transplant eGFR prediction using MIL, and 
Wider violins indicate greater performance variability, and higher median lines indicate more consistent model performance across cross-validation replicates.}
\label{sup:fig:heatmaps_HB_corrected}
\end{figure*}
\subsection*{Institutional Review Board statement}
This study complies with all relevant ethical regulations and was
approved by the Institutional Review Board (IRB) at the University of Florida, including IRB202300413 (“Developing an AI-Ready Infrastructure for Computational Renal Pathology”), IRB202501118 (“Artificial Intelligence-Based \& Ethically-Focused Multi-Modal Data Integration Framework for Screening, Diagnosing, and Caring for CKD Patients”), IRB202501224 (“Computational Image Analysis of Renal Transplant Biopsies to Predict Graft Outcome”), IRB202400360 (“Computational Imaging of Renal Structures for Diagnosing Diabetic Nephropathy 2.0 - First Renewal”), IRB202500133 (“Computational Renal Pathology Suite (ComPRePS): AI-Ready Data for Developing Computational AI Tools to Assess Frozen Section Histology in Kidney Transplant - A Partnership with Digpath Inc.”), IRB202201842 (“Computational Imaging of Renal Structures for Diagnosing Diabetic Nephropathy”), IRB202201030 (“A Cloud-Based Distributed Tool for Computational Renal Pathology”), and CED000000717 (“Kidney Precision Medicine Project (KPMP): A Computational Renal Pathology Suite for KPMP”), as well as associated Computational Image Analysis Platform (CIMAP) and Human BioMolecular Atlas Program (HuBMAP) initiatives where applicable. All human tissue samples analyzed were obtained under IRB-approved protocols from contributing institutions. Samples were fully de-identified before access and analysis. Informed consent was obtained by the contributing institutions, and no compensation was provided to participants.

\section*{Acknowledgements}
We thank Fatemeh Afsari, Dhiraj Srivastava, and Ahmed Naglah for facilitating access to the datasets, Christina Beharry for help with proofreading the paper, and Samantha Hoffman for her help with coordinating manuscript preparation. P.S.'s work is supported by NIH funding from NIDDK—R21 DK128668, R01 DK114485, R01 DK129541, OD OT2 OD033753, OD OT2 038014, and from NIAMS—R01 AR080668. 

\section*{Author contributions statement}
H.R.K. and P.S. conceived the experiments and design of the study.  H.R.K. conducted experiment (s) and wrote the original draft. P.S.L.R., J.L.F., A.G., A.S.P., J.B.H., and P.S. edited the paper. L.R., A.Z.R., K.Y.J., S.J., T.M.E., M.A.W., W.L.C., J.B.H., and M.T.E. contributed to data collection and acquisition used for the experiments. H.R.K., P.S.L.R., and P.S. analysed the experiments. All authors contributed to the revision of the manuscript. 

\end{document}